%% file: main.tex
\pdfoutput=1
\documentclass[11pt]{article}
\usepackage[preprint]{acl}
\usepackage{times}
\usepackage{latexsym}
\usepackage[T1]{fontenc}
\usepackage[utf8]{inputenc}
\usepackage{microtype}
\usepackage{inconsolata}
\usepackage{graphicx}
\usepackage{amsmath,amssymb}
\usepackage{booktabs}
\usepackage{tabularx}
\usepackage{adjustbox}
\usepackage{multirow}
\usepackage[table]{xcolor}
\newsavebox{\ftbox}\newlength{\ftw}%
\newcommand{\ftL}{\hskip\dimexpr3pt-\tabcolsep\relax}\newcommand{\ftR}{\hskip\dimexpr3pt-\tabcolsep\relax}%
\newcommand{\filltabular}[4]{\begingroup\setlength{\tabcolsep}{0pt}\sbox{\ftbox}{\begin{tabular}{#3}#4\end{tabular}}\setlength{\ftw}{#1}\addtolength{\ftw}{-\wd\ftbox}\setlength{\tabcolsep}{\dimexpr\ftw/(2*#2-2)\relax}\ifdim\tabcolsep<0.5pt\setlength{\tabcolsep}{0.5pt}\fi\begin{tabular}{#3}#4\end{tabular}\endgroup}
\definecolor{oursrow}{RGB}{255,248,214}
\definecolor{buyshade}{RGB}{226,240,228}
\definecolor{passshade}{RGB}{236,236,236}
\usepackage{wrapfig}
\usepackage{enumitem}
\usepackage{afterpage}\usepackage{placeins}\usepackage{stfloats}%
\usepackage{algorithm}
\usepackage{algpseudocode}
\usepackage{subcaption}
\usepackage{tikz}
\renewcommand{\topfraction}{0.85}\renewcommand{\dbltopfraction}{0.85}\renewcommand{\floatpagefraction}{0.75}

\usetikzlibrary{positioning,arrows.meta,shapes,fit,calc}

\definecolor{divablue}{RGB}{31,119,180}
\definecolor{eigorange}{RGB}{230,120,30}
\definecolor{llmgray}{RGB}{120,120,120}
\newcommand{\diva}{DIVA}
\newcommand{\NumWorlds}{one sampled world}

\title{Not All Information Is Worth Asking For:\\Decision-Informed Information Seeking with LLMs}

\author{Seunghan Lee \quad Jun Seo \quad Junhyeok Kang \quad Jaehoon Lee \quad Sangjun Han \quad Sungdong Yoo \\
\textbf{Minjae Kim \quad Tae Yoon Lim \quad Dongwan Kang \quad Hwanil Choi \quad Soonyoung Lee \quad Wonbin Ahn} \\
LG AI Research}

\input{numbers}
\begin{document}
\maketitle
\input{sections/abstract}
\input{sections/intro}
\input{sections/related}
\input{sections/method}
\input{sections/experiments}
\input{sections/analysis}
\input{sections/conclusion}

\bibliography{refs}

\clearpage
\appendix
\raggedbottom%
\setcounter{dbltopnumber}{3}\renewcommand{\dbltopfraction}{0.95}\renewcommand{\dblfloatpagefraction}{0.85}
\setcounter{topnumber}{4}\renewcommand{\topfraction}{0.95}\setcounter{totalnumber}{6}\renewcommand{\floatpagefraction}{0.75}
\setcounter{bottomnumber}{2}\renewcommand{\bottomfraction}{0.6}%
\makeatletter\renewenvironment{table*}[1][]{\@dblfloat{table}[tbp]}{\end@dblfloat}\renewenvironment{figure*}[1][]{\@dblfloat{figure}[tbp]}{\end@dblfloat}\renewenvironment{table}[1][]{\@float{table}[tbp]}{\end@float}\renewenvironment{figure}[1][]{\@float{figure}[tbp]}{\end@float}\makeatother
\input{sections/appendix}

\end{document}

%% file: numbers.tex
\newcommand{\RedBin}{75}
\newcommand{\RedBinNoReason}{78}
\newcommand{\IncohCorrMin}{$-$0.01}
\newcommand{\IncohCorrMax}{0.44}
\newcommand{\IncohShrink}{3.1}
\newcommand{\IncohTP}{6.3}
\newcommand{\NumIncoh}{20}
\newcommand{\KthreeSecN}{80}
\newcommand{\KthreeSecA}{68}
\newcommand{\KthreeSecB}{79}
\newcommand{\KFCells}{80}
\newcommand{\KFWorse}{40}
\newcommand{\RecDivaUPlacebo}{43}
\newcommand{\RecDivaPlacebo}{24}
\newcommand{\KfourA}{8}
\newcommand{\KfourCells}{144}
\newcommand{\KfourC}{119}
\newcommand{\GoldSameMin}{95}
\newcommand{\GoldSameMax}{99}

%% file: sections/abstract.tex
\begin{abstract}
Large language models (LLMs) are increasingly used to support decisions under incomplete information, where deciding \textit{what} to look up and \textit{when} to stop matters as much as the final answer.
However, LLMs fall into an \textit{information trap}: They pay for checks, i.e., costly pieces of evidence, that cannot change their decision or are not worth their cost, and they keep asking when the evidence is already sufficient.
To this end, we propose \textbf{D}ecision-\textbf{I}nformed \textbf{V}alue-of-information \textbf{A}cquisition (DIVA), a simple yet effective procedure for LLM information seeking based on the \textit{decision value} of evidence.
DIVA 1)~lets the LLM specify a small decision model, i.e., how likely a bad outcome is and how each check would shift this probability, through short numeric questions and 2)~values each candidate check by how much its imagined outcomes improve the decision net of its cost, stopping when no check is worth its cost.
Its variant, DIVA-U, further accounts for the LLM's uncertain estimates of how strongly each check shifts this probability.
DIVA is training-free and applies to any LLM.
Across four domains and 20 LLMs, DIVA-U has lower regret, i.e., value lost relative to the best possible decision, than every LLM baseline on real-world data, mainly by stopping once evidence suffices, and DIVA cuts the regret of the best LLM baseline by \RedBin\% in synthetic scenarios where every check's reliability is given.
Code: \mbox{\href{https://github.com/seunghan96/diva}{\nolinkurl{github.com/seunghan96/diva}}}.
\end{abstract}

%% file: sections/intro.tex
\section{Introduction}
\label{sec:intro}
\input{tables/diff}

\begin{figure}[t]
\centering
\includegraphics[width=\columnwidth]{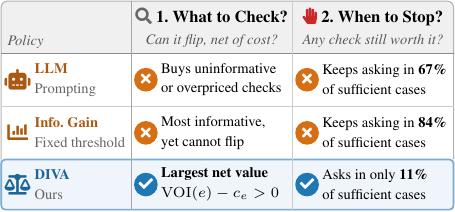}
\caption{\textbf{Desiderata of information seeking.} A good decision maker acquires only checks that can change its decision net of their cost (\textit{what}) and stops once no check is worth its cost (\textit{when}). Prompted LLMs and information gain fail both, whereas \diva{} meets both.}
\label{fig:desiderata}
\end{figure}

Large language models (LLMs) are increasingly used as assistants for decisions under uncertainty, including investment~\citep{li2025investorbench,jin2026finsight}, credit~\citep{feng2023empowering}, and clinical decisions~\citep{li2024mediq}.
In these settings, the evidence at hand is rarely complete, and a good decision maker must decide 1)~\textit{which} additional information to acquire and 2)~\textit{when} to stop acquiring and act, as illustrated in Figure~\ref{fig:desiderata}.
For example, before buying a stock, an analyst may check the company's forward guidance (its sales outlook) or its gross margin, but each check costs time and money, and only some can change the decision.

Recent works equip LLMs to ask clarifying questions~\citep{zhang2025clarify}, to identify missing information~\citep{li2025questbench}, and to select questions by expected information gain (EIG), i.e., how much they are expected to reduce uncertainty~\citep{choudhury2026bedllm}.
However, these approaches treat information as valuable \textit{per se} (Table~\ref{tab:diff}), overlooking a basic principle of decision theory: A piece of information is worth acquiring only if \textit{some of its possible outcomes would change the decision}, and only if \textit{this expected improvement exceeds its cost}~\citep{howard1966information}.

\begin{figure}[t]
\centering
\includegraphics[width=\columnwidth]{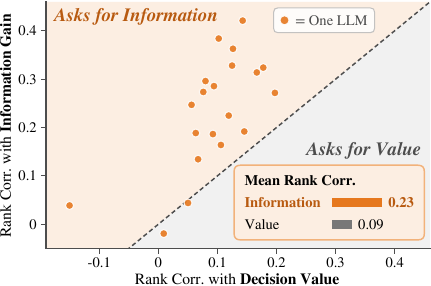}
\caption{\textbf{LLMs ask for information, not value.} Rank correlation between 1)~how likely each LLM is to pick a check and 2)~the check's exact \textit{information gain} or net \textit{decision value}. Each dot is one LLM.}
\label{fig:infovalue}
\end{figure}

A highly informative check can be worthless when none of its outcomes would change the decision, and conversely, a check that reveals little can be highly valuable when it can prevent a costly mistake.
People are known to violate this principle, a tendency called \textit{information bias}~\citep{baron1988heuristics}, and we find that LLMs do so more broadly, paying for checks of little or no decision value.

In this paper, we show that LLMs fall into what we call the \textit{information trap}, i.e., paying for evidence whose decision value does not cover its cost.
As shown in Figure~\ref{fig:infovalue}, the checks that LLMs choose follow their \textit{information gain} rather than their \textit{decision value}, and on real credit and patient records, most LLM baselines are several times \textit{worse} than never acquiring any evidence (Section~\ref{sec:main}).
Most of their \textit{regret}, i.e., the value lost relative to the best possible decision, comes from this trap, and they keep acquiring in most cases, even where the evidence is already sufficient (Appendix~\ref{app:stop}).

To this end, we propose \textbf{D}ecision-\textbf{I}nformed \textbf{V}alue-of-information \textbf{A}cquisition (\diva{}), a simple yet effective, training-free procedure for LLM information seeking based on the \textit{decision value} of evidence.
Specifically, \diva{} 1)~lets the LLM specify a small decision model, i.e., how likely a bad outcome is and how each check would shift this probability, through short numeric questions, 2)~imagines the outcomes of every check and computes the updated beliefs by Bayes' rule, and 3)~values each check by how much it is expected to improve the decision net of its cost, stopping when no check is worth its cost.
Furthermore, we introduce \mbox{\diva{}-U}, which values each check by averaging its decision value over several plausible \textit{strengths} of that check, i.e., how strongly it shifts this probability, as the LLM's answers can be wrong when the scenario does not state how informative the checks are.
As the LLM specifies the decision model once per case, \diva{} needs no further LLM calls during acquisition, and one configuration, selected on development data, serves every domain and LLM.

The main contributions are summarized as:
\setlist[itemize]{leftmargin=0.3cm,itemsep=1pt,parsep=0pt,topsep=3pt}
\begin{itemize}
\item We identify the \textit{information trap} in LLM information seeking, where LLMs pay for checks whose decision value does not cover their cost and keep acquiring even when evidence suffices, often doing worse than never acquiring any evidence.
\item We propose \textbf{\diva{}}, a training-free procedure in which the LLM only specifies a small decision model and the decision value of every check is computed from it, and \textbf{\mbox{\diva{}-U}} for when check strengths must come from the LLM's knowledge.
\item We build 12 decision environments across four domains, eight \textit{knowledge-required} ones from real records and four \textit{controlled} synthetic ones, and show with 20 LLMs that \diva{}-U has the lowest regret in all environments, with analyses of the information trap, robustness to costs and prompts, and cross-domain transfer.
\end{itemize}

%% file: tables/diff.tex
\begin{table*}[t]
\centering
\small
\setlength{\tabcolsep}{4.4pt}
\begin{tabular}{l|ccc|c}
\toprule
 & \multicolumn{2}{c}{\textit{[1] Which information to acquire?}} & \textit{[2] When to stop?} & \multirow{2}{*}[-0.6ex]{Training-free} \\
\cmidrule(lr){2-3}\cmidrule(lr){4-4}
 & Decision value & Elicited model & Stopping rule & \\
\midrule
\multicolumn{1}{l|}{\textit{Information seeking with LLMs}} & & & \multicolumn{1}{c|}{} & \\
FinRiskAtlas \citep{zhong2026finriskatlas} & -- & -- & Ask-or-proceed & \checkmark \\
QuestBench \citep{li2025questbench} & -- & -- & -- & \checkmark \\
MediQ \citep{li2024mediq} & -- & -- & Abstention & \checkmark \\
\midrule
\multicolumn{1}{l|}{\textit{Information gain}} & & & \multicolumn{1}{c|}{} & \\
BED-LLM \citep{choudhury2026bedllm} & -- & -- & -- & \checkmark \\
Entropy Search \citep{truong2025gathering} & -- & -- & -- & \checkmark \\
AFA / EDDI \citep{li2021active,ma2019eddi} & -- & -- & \checkmark & -- \\
\midrule
\multicolumn{1}{l|}{\textit{Decision-aware acquisition}} & & & \multicolumn{1}{c|}{} & \\
VoI ask-vs-act \citep{dong2026value} & \checkmark & -- & Ask-or-act & \checkmark \\
CAM-DF \citep{feng2026scores} & -- & -- & Cost-aware & -- \\
\midrule
\multicolumn{1}{l|}{\textit{LLM-elicited models}} & & & \multicolumn{1}{c|}{} & \\
DeLLMa \citep{liu2025dellma} & -- & \checkmark & -- & \checkmark \\
BIRD \citep{feng2025bird} & -- & \checkmark & -- & -- \\
\midrule
\rowcolor{oursrow} \textbf{\diva{} (ours)} & \checkmark & \checkmark & \textbf{Value-based} & \checkmark \\
\bottomrule
\end{tabular}
\caption{\textbf{Comparison with the closest work.} Unlike prior work, \diva{} (1)~values each check by its \textit{decision value} under (2)~an \textit{LLM-elicited model}, i.e., a decision model whose probabilities are obtained by asking the LLM short numeric questions, (3)~stops by a \textit{value-based rule}, and (4)~works \textit{without training} with any LLM.}
\label{tab:diff}
\end{table*}

%% file: sections/related.tex
\section{Related Works}
\label{sec:related}

\textbf{Information seeking with LLMs.}
Learning to ask has a long history in NLP, from ranking clarification questions \citep{rao2018learning,aliannejadi2019asking} to selective clarification with LLMs \citep{kuhn2022clam,zhang2025clarify}.
Benchmarks test whether LLMs ask for or identify missing information in clinical, underspecified, and game settings \citep{li2024mediq,zhou2025passive,li2025questbench} and in finance \citep{zhong2026finriskatlas,dai2026realfin}, and agents retrieve financial information \citep{jin2026finsight}.
While these works test \textit{whether} a model asks, we evaluate \textit{how much} each check is worth for the decision and when to stop.

\textbf{Information gain and experimental design.}
Bayesian experimental design selects queries by expected information gain \citep{lindley1956measure,rainforth2024modern}, which recent LLM methods adopt for question asking \citep{mazzaccara2024learning,choudhury2026bedllm,hartmann2026amortising}.
Entropy search over the optimal action \citep{truong2025gathering}, value-of-information prompting \citep{dong2026value}, and cost-aware stopping \citep{feng2026scores} move toward decision awareness, and sequential diagnosis and active feature acquisition (AFA) select tests by decision value or learned policies, typically from labeled data \citep{gorry1968experience,heckerman1992toward,krause2009optimal,kachuee2019opportunistic}.
Seeking information that cannot change the decision is known in cognitive science as information bias \citep{baron1988heuristics,nelson2005finding}.
Unlike EIG, \diva{} values a check by its effect on the decision.

\textbf{LLMs as specifiers of probabilistic models.}
LLMs can state priors and likelihoods, i.e., base rates and how evidence shifts these rates, for a solver that computes with them, as in decision-theoretic and Bayesian pipelines \citep{liu2025dellma,feng2025bird}, prior elicitation \citep{capstick2025autoelicit,requeima2024llm}, feature selection \citep{jeong2025llmselect}, and program-aided reasoning \citep{gao2023pal,chen2023program}, although elicited probabilities are often imperfect \citep{xiong2024can,paruchuri2024what}.
We instead use the elicited model to \textit{imagine} unseen check outcomes and decide what to ask and when to stop.

\textbf{LLMs for high-stakes decisions.}
LLMs are evaluated as assistants for financial \citep{xie2024finben,li2025investorbench}, credit \citep{feng2023empowering}, clinical \citep{singhal2023large}, and legal \citep{guha2023legalbench} decisions, and recent studies examine how investor roles shift LLM judgments \citep{asaad2026analyst}.
While these works answer from the evidence at hand, we study which evidence to acquire before deciding, using \textit{point-in-time} filings, i.e., data available at the decision date, which guard against \textit{look-ahead bias}, i.e., using future information \citep{glasserman2023assessing,lee2026palm}.

Table~\ref{tab:diff} compares \diva{} with the closest work on what to acquire, when to stop, and whether training is needed, and Appendix~\ref{app:related} discusses further work.

%% file: sections/method.tex
\section{Methodology}
\label{sec:method}

We now describe \diva{}, a simple yet effective procedure that lets an LLM decide \textit{what} to ask and \textit{when} to stop by reasoning about how each possible answer would change its decision.
The overall framework of DIVA is illustrated in Figure~\ref{fig:overview}.
\input{figs/overview_float}
\subsection{Problem Setup}
\label{sec:setup}
\textbf{Decision problem.}
An \textit{episode}, i.e., one decision case, consists of a binary state $Y\in\{0,1\}$ that is unknown when deciding. Here, $Y{=}1$ denotes the \textit{adverse} case (e.g., the stock underperforms, the borrower defaults), the model chooses one of two actions $a\in\{\textsc{act},\textsc{pass}\}$ (e.g., \textsc{buy}/\textsc{pass}, \textsc{approve}/\textsc{decline}), and a payoff table stated to the model (Table~\ref{tab:payoff}) gives the gain $g$ of acting in the benign case and the loss $l$ in the adverse case.

\begin{table}[t]
\centering
\small
\setlength{\tabcolsep}{13.3pt}
\begin{tabular}{l|cc}
\toprule
Action & $Y{=}0$ (Benign) & $Y{=}1$ (Adverse) \\
\midrule
\textsc{act} & \cellcolor{buyshade}$+g$ & \cellcolor{buyshade}$-l$ \\
\textsc{pass} & \cellcolor{passshade}$0$ & \cellcolor{passshade}$0$ \\
\bottomrule
\end{tabular}
\caption{\textbf{Payoff table.} Utility $U(a,Y)$ of each action under each state. Acting is optimal when the belief is below $b^\dagger=g/(g+l)$. Table~\ref{tab:notation} lists all notation.}
\label{tab:payoff}
\end{table}
The model observes evidence $E$ (the base rate and any known findings) and a set of candidate checks $\mathcal{C}$, where each check $e\in\mathcal{C}$ reveals one of a finite set of outcomes $\mathcal{V}_e$ at a cost $c_e$.
The \textit{reliability} of a check is how likely each of its outcomes is under each state, $P(e{=}v|Y{=}y)$ (e.g., a test that flags 80\% of adverse and 5\% of benign cases). It is either stated in the scenario or must come from the model's knowledge.
At every step, the model either acquires a remaining check or stops and acts.

\textbf{Value of information (VOI).}
For a belief $b=P(Y{=}1|E)$, the value of acting optimally is $V^*(b)=\max\{(1-b)g-bl,\,0\}$, i.e., acting if it is expected to pay off and passing otherwise.
The decision value of a check is the expected improvement of this quantity after observing its outcome:
\begin{equation}
\mathrm{VOI}(e|E)=\sum_{v\in\mathcal{V}_e} P(v|E)\,V^*(b_{e,v})-V^*(b),
\label{eq:voi}
\end{equation}
where $b_{e,v}=P(Y{=}1|E,e{=}v)$ is the belief after observing outcome $v$ of check $e$,
and acquiring $e$ is worthwhile only if $\mathrm{VOI}(e|E)>c_e$ \citep{howard1966information,raiffa1961applied}.
In contrast, the EIG used by Bayesian experimental design \citep{rainforth2024modern} and by recent LLM question-asking methods \citep{choudhury2026bedllm} replaces $V^*$ with the negative entropy of $Y$, which \textit{ignores the payoffs and the cost} of each check.
Note that $\mathrm{VOI}(e|E)=0$ whenever no outcome of $e$ can change the decision, i.e., move the belief across $b^\dagger$, however informative $e$ is.

\textbf{Information trap.}
We say that a policy falls into the \textit{information trap} when it pays for a check whose decision value does not cover its cost, i.e., $\mathrm{VOI}(e|E)<c_e$.
We distinguish three forms by comparing each acquisition with the \textit{DP-optimal} decision, i.e., the best decision when all future checks and stops are planned exactly by dynamic programming (DP): The check is 1)~\textit{uninformative} (EIG at most 0.05 bits), 2)~\textit{decision-irrelevant}, i.e., informative but unable to change the decision ($\mathrm{VOI}=0$ with EIG above 0.05 bits), or 3)~\textit{overpriced} ($0<\mathrm{VOI}<c_e$).
The remaining errors are acquiring a valuable but non-optimal check (\textit{suboptimal}) and stopping when acquiring is optimal (\textit{premature stop}). Errors are decisions that differ from the DP-optimal one, classified by the decision value and EIG of the check (Appendix~\ref{app:stop}).

\begin{figure}[t]
\centering
\includegraphics[width=\columnwidth]{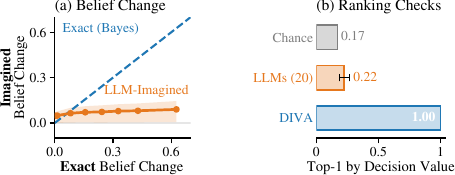}
\caption{\textbf{LLMs do not imagine evidence coherently.} (a) How much each LLM says the belief would change after an imagined check outcome vs.\ the exact Bayesian change, binned and pooled over \NumIncoh{} LLMs. (b) How often the top check by imagined decision value is the best one when acquiring is optimal.}
\label{fig:incoh}
\end{figure}

\subsection{Why Asking the LLM Directly Fails}
\label{sec:why}
\textbf{Failure of direct imagination.}
A natural way to obtain Eq.~\eqref{eq:voi} from an LLM is to ask it for its current belief and for its belief after each \textit{imagined} outcome of each check.
We find that such raw imagined updates are \textit{incoherent} (Figure~\ref{fig:incoh}a): Across \NumIncoh{} LLMs, the correlation of elicited and exact Bayesian belief changes is at most \IncohCorrMax{}, and the elicited beliefs violate the law of total probability, i.e., their outcome-weighted average differs from the current belief, by \IncohTP{} percentage points on average.
As VOI is a difference of maxima over such beliefs, these errors are amplified, and the top check under raw imagined beliefs matches the best one barely more often than chance (Figure~\ref{fig:incoh}b).

\textbf{Separating knowledge from computation.}
DIVA therefore separates \textit{what the LLM knows} from \textit{how the decision is computed}.
The LLM only answers numeric questions that specify the decision model, and Bayes' rule evaluates each imagined outcome, making every update coherent.

\subsection{DIVA: Imagine, Value, Stop}
\textbf{1) Model specification.}
The LLM is asked short, independent numeric questions about two quantities: (i)~the \textit{base rate} $\hat\pi$ of the adverse case and (ii)~the \textit{outcome likelihoods}, i.e., estimates of the reliabilities, of every outcome $v$ of each known or candidate check $e$,
\begin{equation}
\hat\ell_e(v|y)\approx P(e{=}v|Y{=}y),\quad y\in\{0,1\}.
\end{equation}
If the scenario text already gives these numbers, the LLM simply copies them.
Otherwise, it answers from its own knowledge and is asked instead for the \textit{outcome-conditioned risk} $\hat r_e(v)\approx P(Y{=}1|e{=}v)$, i.e., the probability of the adverse case given each outcome, and Bayes' rule converts these risks into likelihoods (Appendix~\ref{app:know}).
Each answer is a single percentage decoded without sampling, and specifying the decision model takes a few LLM calls per check, once per episode.

\textbf{2) Coherent imagination.}
Given the decision model from the LLM (i.e., the base rate $\hat\pi$ and the likelihoods $\hat\ell_e(v|y)$), the current belief $\hat b$ and every imagined belief $\hat b_{e,v}$ follow from Bayes' rule:\predisplaypenalty=0 %
\begin{equation}
\begin{gathered}
\hat b=\frac{\hat\pi\prod_{k\in\mathcal{K}}\hat\ell_k(v_k|1)}{\sum_{y}\hat\pi_y\prod_{k\in\mathcal{K}}\hat\ell_k(v_k|y)},\\
\hat b_{e,v}=\frac{\hat b\,\hat\ell_e(v|1)}{\hat b\,\hat\ell_e(v|1)+(1-\hat b)\,\hat\ell_e(v|0)},
\end{gathered}
\label{eq:imagine}
\end{equation}
\predisplaypenalty=10000
where $\mathcal{K}$ is the set of known findings with outcomes $v_k$, $\hat\pi_1=\hat\pi$ and $\hat\pi_0=1-\hat\pi$, and the predictive probability of each imagined outcome is $\hat P(v|E)=\hat b\,\hat\ell_e(v|1)+(1-\hat b)\,\hat\ell_e(v|0)$.
By construction, the imagined beliefs are \textit{coherent} with the current belief, i.e., $\sum_v \hat P(v|E)\,\hat b_{e,v}=\hat b$.

\textbf{3) Decision valuation.}
DIVA scores each check by its \textit{net decision value}, using the stated payoffs:
\begin{equation}
\begin{aligned}
\widehat{\Delta}(e)&=\sum_{v\in\mathcal{V}_e}\hat P(v|E)\,V^*(\hat b_{e,v})-V^*(\hat b)-c_e\\
&=\widehat{\mathrm{VOI}}(e|E)-c_e.
\end{aligned}
\label{eq:net}
\end{equation}

\textbf{4) Value-based stopping.}
DIVA acquires $e^\star=\arg\max_e\widehat\Delta(e)$ if $\widehat\Delta(e^\star)>0$ and otherwise stops and takes $\arg\max_a \mathbb{E}_{\hat b}[U(a,Y)]$.
After an acquisition, the observed outcome is added to $\mathcal{K}$ and Steps 2--4 are repeated without further LLM calls, as the decision model does not change.

\textbf{Non-myopic planning.}
Steps 3--4 are \textit{myopic}, i.e., they value only the next acquisition. We also consider a non-myopic variant, \textit{DIVA-DP}, which replaces Eq.~\eqref{eq:net} with exact dynamic programming over the remaining checks under the specified model, so that a cheap check whose value arises only in combination with later checks is valued correctly. It needs no extra parameter.

\textbf{Design choices.}
When the scenario states the reliabilities, the LLM copies them into a correct decision model, and it is optimal to stop acquiring once no check has a positive net value.
Otherwise, we select the question form, the \textit{shrinkage}, and the stopping rule on development data (Appendix~\ref{app:know}).
Shrinkage discounts the LLM's answers by pulling the two likelihoods of each check together, weakening its strength, as \textit{noise can overstate the strengths of useless checks and make them look valuable}.

\subsection{DIVA-U: Uncertainty-Aware Valuation}
\label{sec:divau}
Formally, the \textit{strength} of check $e$ is the log-odds gap $\mathrm{logit}\,\hat\ell_e(1|1)-\mathrm{logit}\,\hat\ell_e(1|0)$ derived from the LLM's answers, which measures \textit{how far apart its two outcomes push the belief}.
Without stated reliabilities, the implied strengths are uncertain, yet \diva{} treats them as exact and values a check at zero whenever no outcome crosses the threshold, even if its true strength would make it valuable.
DIVA-U therefore \textit{treats the strength of each check as uncertain} and values it by the expected net decision value under this uncertainty.
Specifically, it shifts the strength of each check by $\sigma z$ with $z\sim\mathcal{N}(0,1)$, moving the log-odds of $\hat\ell_e(1|1)$ and $\hat\ell_e(1|0)$ apart or together. The value is:
\begin{equation}
\widehat\Delta_{\mathrm{U}}(e)=\mathbb{E}_{z\sim\mathcal{N}(0,1)}\big[\widehat{\mathrm{VOI}}_{z}(e)\big]-c_e,
\label{eq:netu}
\end{equation}
where $\widehat{\mathrm{VOI}}_{z}(e)$ is Eq.~\eqref{eq:voi} under the shifted decision model, and the expectation is a weighted sum over three values of $z$ (\textit{Gauss--Hermite quadrature}).

As the decision value cannot fall below zero, averaging can only raise the value of checks that \diva{} values at zero, so \mbox{\diva{}-U} acquires more often than \diva{}. This average is deliberately optimistic and is not the exact decision value under the uncertainty (Appendix~\ref{app:f8}).
We use $\sigma=0$ when reliabilities are stated and select $\sigma=1$ on development data otherwise (Appendix~\ref{app:know}).

%% file: figs/overview_float.tex
\begin{figure*}[t]
\centering
\includegraphics[width=\textwidth]{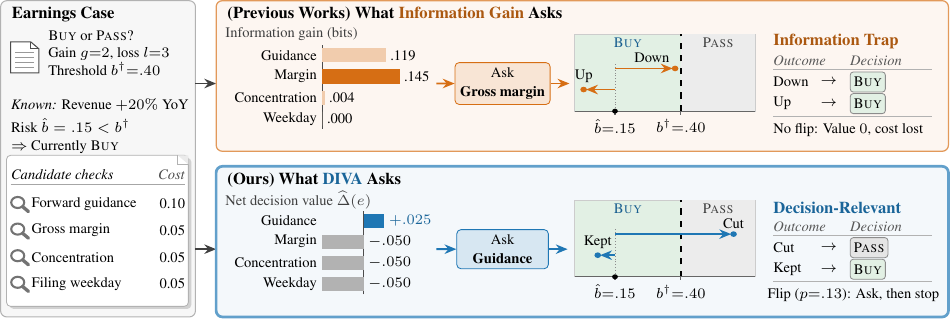}
\caption{\textbf{Overall framework of \diva{}.} In an earnings case, \textbf{\textcolor{eigorange}{information gain}} asks for the gross margin, whose outcomes cannot move the \textit{belief} $\hat b$ across the decision threshold $b^\dagger$, so the cost is paid without changing the decision. In contrast, \textbf{\textcolor{divablue}{\diva{}}} imagines every outcome of every check and values each by how much it can change the decision net of its cost. It asks for the forward guidance, whose cut would flip \textsc{buy} to \textsc{pass}, and then stops.}
\label{fig:overview}
\end{figure*}

%% file: sections/experiments.tex
\input{tables/main_know}
\section{Experiments}
\label{sec:experiments}

\subsection{Experimental Settings}
\label{sec:settings}

\textbf{Two types of environments.}
We group the environments by whether the reliabilities of the checks are stated.
\textit{(i)~Knowledge-required environments}: Each scenario states the payoffs, a base rate, and costly candidate checks, but not their reliabilities: \textit{SEC filings} of S\&P 500 companies (2025--2026) and two public \textit{credit} \citep{yeh2009comparisons,hofmann1994statlog} and two \textit{clinical} \citep{cdc2017diabetes,janosi1989heart} datasets (the \textit{tabular records}), with ground-truth decision values from a \textit{reference model}, i.e., a classifier fitted on training records (Appendix~\ref{app:real}).
\textit{(ii)~Controlled environments}: Four synthetic diagnostic domains (finance, credit, clinical, and legal) state every reliability and have exact decision values (Appendix~\ref{app:envs}).

\textbf{LLMs.}
We evaluate 20 open-weight LLMs from eight families (0.6B to 72B) served with vLLM \citep{kwon2023efficient}: Qwen3 \citep{yang2025qwen3}, Qwen2.5 \citep{qwen2024qwen25}, Llama-3.1/3.3 \citep{grattafiori2024llama}, Gemma-2/3 \citep{gemma2024gemma,gemma2025gemma}, Mistral \citep{jiang2023mistral}, Phi-4 \citep{abdin2024phi4}, and the reasoning models gpt-oss \citep{openai2025gptoss} and DeepSeek-R1-Distill \citep{deepseekai2025deepseekr1}.

\textbf{Baseline methods.}
We compare five categories of baselines.
\textit{Prompting}: Direct, chain-of-thought (CoT) \citep{wei2022chain}, self-consistency (SC) \citep{wang2023selfconsistency}, and ReAct \citep{yao2023react}.
\textit{Decision-aware prompting}: Ask-P, which first decides whether to ask \citep{zhong2026finriskatlas}, VoI-P, which verbally values each check, and VoI-Ask \citep{dong2026value}.
\textit{Information gain}: BED-LLM \citep{choudhury2026bedllm}, which acquires the check with the largest EIG.
\textit{Program-aided}: PAL \citep{gao2023pal}.
\textit{Reference}: Never (no check) and Always ask (every check, in \diva{}'s order).

\textbf{Evaluation metrics.}
We report the \textit{regret}, i.e., the value lost relative to an optimal decision, normalized by the loss $l$ and multiplied by 100 (\textit{lower is better}, with formulas in Appendix~\ref{app:setup}):
\begin{itemize}[leftmargin=0.4cm,itemsep=0pt,topsep=2pt,parsep=0pt]
\item \textit{DP regret}: Regret of the first decision against the \textit{DP optimum}, i.e., the best action under DP.
\item \textit{Myopic regret}: Regret against the \textit{myopic optimum}, i.e., the best action when only the next acquisition is valued (results in Appendix~\ref{app:allmetrics}).
\item \textit{Episode regret}: Regret of a full acquire-then-decide episode against the optimal policy, normalized by $g{+}l$ in the controlled environments.
\end{itemize}

\subsection{Main Results}
\label{sec:main}

\textbf{Knowledge-required environments.}
In Table~\ref{tab:know} (left), on the four tabular records, every Prompting baseline has at least twice the regret of not acquiring at all, demonstrating the information trap (Figure~\ref{fig:trap}).
In contrast, \diva{}-U has the \textit{lowest regret in all eight environments and beats every LLM baseline} for at least 18 of the 20 LLMs on each tabular record.
Notably, the results with the SEC volatility targets (Vol-5d, Vol-20d) show why \diva{}-U is needed: \diva{}, which treats the elicited strengths as exact, is worse than always asking there, whereas \diva{}-U, which accounts for their uncertainty, has the lowest regret.

\textbf{Controlled environments.}
With all reliabilities stated (Table~\ref{tab:know}, right), only VoI-Ask and PAL among the LLM baselines beat never acquiring in all four domains.
\diva{} beats every LLM baseline for all 20 LLMs in every domain.

\begin{figure}[t]
\centering
\includegraphics[width=\columnwidth]{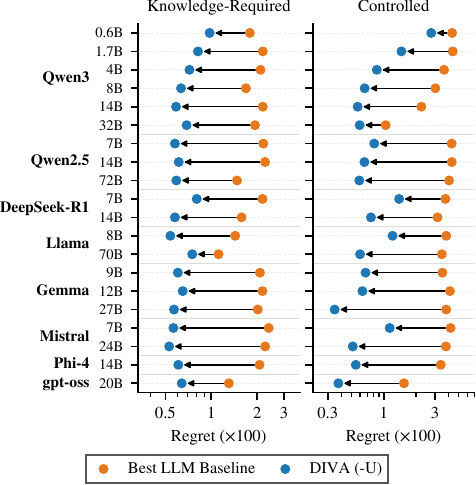}
\caption{\textbf{Effectiveness across 20 LLMs.} Regret of the best LLM baseline of each LLM (orange) and \diva{} (blue, \diva{}-U on knowledge-required environments).}
\label{fig:llms}
\end{figure}

\textbf{Effectiveness across 20 LLMs.}
Figure~\ref{fig:llms} compares \diva{} with the best LLM baseline of each of the 20 LLMs: \diva{} outperforms this baseline for every LLM in both environment types, reducing it by \RedBin\% on average in the controlled environments, so its gains \textit{do not depend on the particular LLM}.

\subsection{End-to-End Episodes}
\label{sec:episodes}
Whereas the main results in Section~\ref{sec:main} score only the \textit{first} decision, we now run each policy \textit{until it stops and acts}, which also scores when to stop.

\input{tables/episodes}

\textbf{Knowledge-required environments.}
Table~\ref{tab:episodes} (left) runs full episodes on the first 150 test scenarios. To compare only what each policy acquires, every policy takes its final action from the belief of the reference model. \diva{} has the lowest average regret, below never acquiring in five of the eight environments, and \diva{}-U the lowest of all methods on each of the four tabular records.
Every LLM baseline has higher average regret than never acquiring, whereas \diva{}-U falls behind it only on SEC filings, where its fixed uncertainty makes it over-acquire (Appendix~\ref{app:episodes_full}).

\textbf{Controlled environments.}
In Table~\ref{tab:episodes} (right), \diva{} has about one sixth of the regret of the best LLM baseline and beats every LLM baseline in every domain for all 20 LLMs, whereas every LLM baseline has higher regret than never acquiring.

%% file: tables/main_know.tex
\begin{table*}[t]
\centering
\small
\setlength{\tabcolsep}{2pt}
\begin{tabular}{l|cccccccc|cccc}
\toprule
\multirow{3}{*}[-1.4ex]{\begin{tabular}[c]{@{}l@{}}Method\\(Avg.\ across 20 LLMs)\end{tabular}} & \multicolumn{8}{c|}{\textbf{Knowledge-required environments}} & \multicolumn{4}{c}{\textbf{Controlled environments}} \\
\cmidrule(lr){2-9}\cmidrule(lr){10-13}
 & \multicolumn{2}{c}{Credit} & \multicolumn{2}{c}{Clinical} & \multicolumn{4}{c|}{SEC filings (2025--2026)} & \multirow{2}{*}[-0.6ex]{Finance} & \multirow{2}{*}[-0.6ex]{Credit} & \multirow{2}{*}[-0.6ex]{Clinical} & \multirow{2}{*}[-0.6ex]{Legal} \\
\cmidrule(lr){2-3}\cmidrule(lr){4-5}\cmidrule(lr){6-9}
 & Taiwan & German & CDC & Heart & Ret-1d & Ret-5d & Vol-5d & \multicolumn{1}{c|}{Vol-20d} &  & & & \\
\midrule
\multicolumn{13}{l}{\textit{Prompting}} \\
Direct & 6.47 & 5.15 & 5.23 & 6.30 & 1.26 & 1.21 & 1.24 & 1.08 & 4.69 & 4.35 & 6.22 & 5.80 \\
CoT~\citep{wei2022chain} & 7.50 & 6.53 & 6.56 & 6.87 & 1.19 & 1.25 & 1.28 & 1.10 & 4.71 & 4.20 & 5.78 & 5.96 \\
SC~\citep{wang2023selfconsistency} & 7.52 & 6.67 & 6.82 & 6.81 & 1.17 & 1.26 & 1.27 & 1.03 & 4.42 & 4.17 & 5.84 & 5.64 \\
ReAct~\citep{yao2023react} & 8.77 & 7.22 & 8.00 & 7.15 & 1.17 & 1.31 & 1.09 & 0.91 & 5.04 & 4.18 & 5.99 & 5.82 \\
\midrule
\multicolumn{13}{l}{\textit{Decision-aware prompting}} \\
Ask-P~\citep{zhong2026finriskatlas} & 6.97 & 6.69 & 6.25 & 7.71 & 1.49 & 1.39 & 1.87 & 1.74 & 5.08 & 4.46 & 6.38 & 6.11 \\
VoI-P & 5.97 & 5.49 & 5.42 & 6.19 & 1.75 & 1.54 & 2.03 & 2.11 & 4.70 & 4.38 & 5.34 & 5.57 \\
VoI-Ask~\citep{dong2026value} & 0.80 & 1.12 & 1.30 & 2.73 & 1.88 & 1.39 & 3.00 & 3.66 & 4.39 & 4.03 & 3.40 & 4.43 \\
\midrule
\multicolumn{13}{l}{\textit{Information gain}} \\
\begin{tabular}[c]{@{}l@{}}BED-LLM\\\citep{choudhury2026bedllm}\end{tabular} & 6.24 & 5.90 & 3.46 & 7.12 & 1.30 & 1.28 & 1.89 & 1.81 & 5.50 & 4.94 & 7.07 & 6.07 \\
\midrule
\multicolumn{13}{l}{\textit{Program-aided}} \\
PAL~\citep{gao2023pal} & 3.56 & 3.05 & 2.53 & 4.31 & 1.88 & 1.49 & 2.54 & 2.84 & 4.35 & 3.97 & 4.34 & 4.90 \\
\midrule
\multicolumn{13}{l}{\textit{Reference}} \\
Never & 0.89 & 1.06 & 0.31 & 2.96 & 2.64 & 1.76 & 3.71 & 4.38 & 4.63 & 4.54 & 4.52 & 4.98 \\
Always ask & 5.17 & 4.48 & 4.78 & 4.08 & \underline{0.72} & \underline{0.77} & \underline{0.90} & \underline{0.77} & \underline{2.38} & \underline{1.99} & \underline{3.10} & \underline{2.87} \\
\midrule
\multicolumn{13}{l}{\textit{Ours}} \\
\rowcolor{oursrow} \diva{} & \underline{0.48} & \underline{0.65} & \underline{0.26} & \underline{1.71} & 1.01 & 0.93 & 2.45 & 2.80 &  &  &  &  \\
\rowcolor{oursrow} \diva{}-U & \textbf{0.41} & \textbf{0.52} & \textbf{0.22} & \textbf{1.12} & \textbf{0.67} & \textbf{0.71} & \textbf{0.83} & \textbf{0.74} & \multirow{-2}{*}{\textbf{0.86}} & \multirow{-2}{*}{\textbf{0.97}} & \multirow{-2}{*}{\textbf{0.73}} & \multirow{-2}{*}{\textbf{0.87}} \\
\midrule
Wins of \diva{}-U & 20/20 & 20/20 & 18/20 & 20/20 & 16/20 & 18/20 & 15/20 & 7/20 & 20/20 & 20/20 & 20/20 & 20/20 \\
\bottomrule
\end{tabular}
\caption{\textbf{DP regret} of the first decision, averaged over 20 LLMs. \diva{}-U equals \diva{} in the controlled environments. Wins: \diva{}-U beats every LLM baseline. Ret / Vol-$k$d: $k$-day return / volatility.}
\label{tab:know}
\end{table*}

%% file: tables/episodes.tex
\begin{table}[t]
\centering
\small
\setlength{\tabcolsep}{5pt}
\begin{tabular}{l|cc}
\toprule
\begin{tabular}[c]{@{}l@{}}Method\\(Avg.\ across 20 LLMs)\end{tabular} & \begin{tabular}[c]{@{}c@{}}Knowledge-\\required\end{tabular} & Controlled \\
\midrule
\multicolumn{3}{l}{\textit{Prompting}} \\
Direct & 18.98 & 28.46 \\
CoT \citep{wei2022chain} & 20.64 & 28.00 \\
SC \citep{wang2023selfconsistency} & 22.87 & 29.84 \\
ReAct \citep{yao2023react} & 19.99 & 23.98 \\
\midrule
\multicolumn{3}{l}{\textit{Decision-aware prompting}} \\
Ask-P \citep{zhong2026finriskatlas} & 18.73 & 28.34 \\
VoI-P & 9.79 & 17.34 \\
VoI-Ask \citep{dong2026value} & 2.40 & 11.16 \\
\midrule
\multicolumn{3}{l}{\textit{Information gain}} \\
\begin{tabular}[c]{@{}l@{}}BED-LLM\\\citep{choudhury2026bedllm}\end{tabular} & 13.15 & 19.90 \\
\midrule
\multicolumn{3}{l}{\textit{Program-aided}} \\
PAL \citep{gao2023pal} & 4.97 & 13.66 \\
\midrule
\multicolumn{3}{l}{\textit{Reference}} \\
Never & \underline{2.25} & \underline{10.93} \\
Always ask & 30.97 & 36.57 \\
\midrule
\multicolumn{3}{l}{\textit{Ours}} \\
\rowcolor{oursrow} \diva{} & \textbf{2.09} &  \\
\rowcolor{oursrow} \diva{}-U & 2.77 & \multirow{-2}{*}{\textbf{1.80}} \\
\bottomrule
\end{tabular}
\caption{\textbf{Episode regret} of full acquire-then-decide episodes, averaged over environments and 20 LLMs.}
\label{tab:episodes}
\end{table}

%% file: sections/analysis.tex
\section{Analysis}
\label{sec:analysis}

\input{tables/ablation}

\textbf{Ablation study.}
In Table~\ref{tab:ablation}, every module added to the direct choice lowers the regret, whereas replacing the decision value of \diva{} with cost-blind EIG on the same decision model erases most of the gain (5.03 vs.\ 0.83), and adding the cost to EIG recovers it.
Giving \diva{} the \textit{ground-truth} likelihoods instead of the LLM's answers lowers its regret from 0.83 to 0.31, indicating that most of its remaining error comes from the LLM's answers.

\begin{figure}[t]
\centering
\includegraphics[width=\columnwidth]{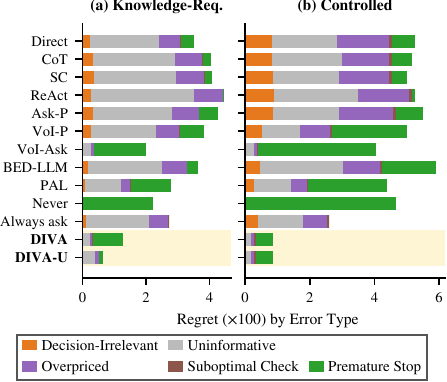}
\caption{\textbf{Anatomy of the information trap.} Regret of the first decision by error type on (a) knowledge-required and (b) controlled environments. \diva{}-U equals \diva{} in (b), where all reliabilities are stated.}
\label{fig:trap}
\end{figure}

\textbf{Anatomy of the information trap.}\label{sec:trap}
Figure~\ref{fig:trap} attributes the regret of every first decision to its error type against the DP-optimal decision. The three forms of the trap cause over 70\% of the regret of six of the nine LLM baselines in both environment types, with uninformative checks as the largest part.
VoI-Ask \citep{dong2026value} avoids the trap but stops prematurely, which causes over 80\% of its regret, whereas \diva{} and its variants have the smallest regret in both (Appendix~\ref{app:stop}).

\input{tables/variants}

\textbf{Myopic and planned acquisition.}
Table~\ref{tab:variants} compares \diva{}, \diva{}-U, and \diva{}-DP with the best LLM baseline, VoI-Ask.
Planning lowers the regret of \diva{} in both environment types, at the first decision and over full episodes, indicating that part of its remaining regret \textit{stems from its myopic valuation}.
Without stated reliabilities, \diva{}-U beats planning at the first decision (0.65 vs.\ 1.07) but falls behind \diva{}-DP and VoI-Ask over full episodes, as it keeps acquiring after the evidence suffices.

\input{tables/generic_main}

\textbf{What the LLM's knowledge contributes.}
Table~\ref{tab:generic} replaces the LLM's answers with one strength shared by every check. Even without any LLM, \diva{}-U has lower regret than the best LLM baseline and never acquiring in every column, indicating that most of the gain stems from cost-aware valuation, mainly of when to stop (Appendix~\ref{app:further}).
When reliabilities are stated, the LLM's strengths lower the regret below one shared strength, whereas otherwise they do not, indicating that the LLM's knowledge of the checks limits the gain.

\textbf{Other analyses.}
\diva{}-U beats DeLLMa \citep{liu2025dellma} and UoT \citep{hu2024uncertainty} for every LLM and environment, and with $\pm$50\% errors in payoffs and costs, it beats every LLM baseline in seven of eight knowledge-required environments.

%% file: tables/ablation.tex
\begin{table}[t]
\centering
\small
\setlength{\tabcolsep}{3.7pt}
\begin{tabular}{l|cc}
\toprule
Variant & \begin{tabular}[c]{@{}c@{}}Regret\\($\times$100)\end{tabular} & $\Delta$ \\
\midrule
Direct & 5.52 & -- \\
+ Decision value on imagined beliefs & 3.49 & $-$2.03 \\
+ Elicited decision model (11-option) & 1.68 & $-$1.81 \\
\rowcolor{oursrow} + Numeric answers (= \diva{}) & \underline{0.83} & $-$0.85 \\
\rowcolor{oursrow} + DP lookahead (= \diva{}-DP) & \textbf{0.48} & $-$0.35 \\
\midrule
\diva{} with cost-blind EIG & 5.03 & +4.20 \\
\midrule
Reference: Ground-truth decision model & 0.31 & $-$0.52 \\
\bottomrule
\end{tabular}
\caption{\textbf{Ablation study.} DP regret on 496 controlled scenarios, averaged over the same 20 LLMs.}
\label{tab:ablation}
\end{table}

%% file: tables/variants.tex
\begin{table}[t]
\centering
\small
\setlength{\tabcolsep}{5pt}
\begin{tabular}{l|cc|cc}
\toprule
\multirow{2}{*}[-0.7ex]{Method} & \multicolumn{2}{c|}{DP regret} & \multicolumn{2}{c}{Episode regret} \\
\cmidrule(lr){2-3}\cmidrule(lr){4-5}
 & Know. & Ctrl. & Know. & Ctrl. \\
\midrule
\begin{tabular}[c]{@{}l@{}}VoI-Ask\\\citep{dong2026value}\end{tabular} & 1.99 & 4.06 & 2.40 & 11.16 \\
\midrule
\rowcolor{oursrow} \diva{} & 1.29 &  & \underline{2.09} &  \\
\rowcolor{oursrow} \diva{}-U & \textbf{0.65} & \multirow{-2}{*}{\underline{0.86}} & 2.77 & \multirow{-2}{*}{\underline{1.80}} \\
\rowcolor{oursrow} \diva{}-DP & \underline{1.07} & \textbf{0.54} & \textbf{1.86} & \textbf{1.58} \\
\bottomrule
\end{tabular}
\caption{\textbf{Myopic and planned acquisition.} Regret averaged over the knowledge-required (Know.) or controlled (Ctrl.) environments and 20 LLMs.}
\label{tab:variants}
\end{table}

%% file: tables/generic_main.tex
\begin{table}[t]
\centering
\small
\setlength{\tabcolsep}{3pt}
\begin{tabular*}{\linewidth}{@{\extracolsep{\fill}}ll|cc|c@{}}
\toprule
\multirow{2}{*}[-0.7ex]{Method} & \multirow{2}{*}[-0.7ex]{\begin{tabular}[c]{@{}l@{}}Check\\strengths\end{tabular}} & \multicolumn{2}{c|}{\textbf{Knowledge-required}} & \multirow{2}{*}[-0.7ex]{\textbf{Controlled}} \\
\cmidrule(lr){3-4}
 & & Records & SEC & \\
\midrule
\multicolumn{2}{@{}l|}{Best LLM baseline} & 1.49 & 1.10 & 4.04 \\
\multicolumn{2}{@{}l|}{Never} & 1.31 & 3.12 & 4.67 \\
\midrule
\multirow{2}{*}{\diva{}} & One shared & 0.72 & 1.23 & 2.07 \\
 & From LLM & 0.77 & 1.80 & \textbf{0.86} \\
\midrule
\multirow{2}{*}{\diva{}-U} & One shared & \textbf{0.52} & \textbf{0.63} & \underline{1.50} \\
 & From LLM & \underline{0.57} & \underline{0.74} & \textbf{0.86} \\
\bottomrule
\end{tabular*}
\caption{\textbf{What the LLM's knowledge contributes.} DP regret of the first decision, averaged over 20 LLMs. Best LLM baseline: Best of Table~\ref{tab:know} per environment.}
\label{tab:generic}
\end{table}

%% file: sections/conclusion.tex
\section{Conclusion}
\label{sec:conclusion}
We introduce \diva{}, which asks an LLM only for a small decision model and computes how much each check can improve the decision net of its cost, and its uncertainty-aware variant \diva{}-U.
Across 20 LLMs, \diva{}-U has lower first-decision regret than every LLM baseline in all twelve environments, mainly by deciding when to stop.
We hope this work highlights decision value, not information, as the goal of evidence-gathering LLM assistants.

\clearpage
\section*{Limitations}
\label{sec:limits}
\looseness=1 First, \diva{} assumes that the decision maker's payoffs are given in the task specification, and our multi-state experiments show that eliciting payoffs from LLMs is error-prone (Appendix~\ref{app:ms}).
Second, \diva{} treats the checks as independent given the state, which can over-count correlated evidence, and its quality is bounded by the LLM's knowledge of how informative each check is, which \diva{}-U mitigates but does not remove (Appendix~\ref{app:knowledge}).
Third, our knowledge-required gains are measured against a reference model fitted on historical records rather than realized outcomes, so they inherit its errors, and gains in realized utility are not reliable (Appendix~\ref{app:realized}). The SEC payoffs also place the threshold near the reference belief, which favors acquiring (Appendix~\ref{app:f8}), and over full episodes, the gains rely on the reference model's belief for the final action.
Fourth, \diva{}-U is designed for the first decision and over-acquires in multi-step episodes, as its uncertainty about the elicited strengths does not shrink with observations. It was also adopted after inspecting interim test results, so its only untouched evidence comes from new test scenarios and later SEC filings (Appendix~\ref{app:prereg}).
Finally, we evaluate open-weight LLMs of up to 72B, with binary or binned checks in English and SEC filings of S\&P 500 firms only, so free-text evidence remains to be evaluated in future work.

\section*{Ethical Considerations}
\label{sec:ethics}
\diva{} is designed as decision support for evidence gathering, not as an autonomous decision maker, and our environments are research benchmarks rather than validated deployment settings.
The credit and clinical records include demographic attributes such as age and sex among the candidate checks, which we use only to study acquisition without evaluating the fairness of the resulting decisions.
The records are de-identified public data from the UCI Machine Learning Repository, and the SEC environments combine public filings with licensed stock prices, involving no human subjects or new personal data.
All LLMs are open-weight models used under their licenses (Appendix~\ref{app:models}).

%% file: sections/appendix.tex
\begin{table*}[t]
\centering
\small
\setlength{\tabcolsep}{5pt}\setlength{\aboverulesep}{0pt}\setlength{\belowrulesep}{0pt}\renewcommand{\arraystretch}{1.25}
\begin{tabular}{l|l|>{\raggedright\arraybackslash}p{3.6cm}lcl}
\toprule
Domain & Environment & Source & Decision & Checks & Test \\
\midrule
\rowcolor{gray!12}\multicolumn{6}{l}{[1] \textit{Knowledge-required environments}} \\
\multirow{4}{*}[1.45pt]{Credit} & Taiwan & Credit-card default\newline\citep{yeh2009comparisons} & Approve / decline & 6 (Binary) & 2,000 \\
 & German & German credit\newline\citep{hofmann1994statlog} & Approve / decline & 6 (Binary) & 1,000 \\
\midrule
\multirow{5}{*}[2.85pt]{Clinical} & CDC & CDC diabetes indicators \citep{cdc2017diabetes} & Skip / refer & 6 (Binary) & 2,000 \\
 & Heart & UCI heart disease\newline\citep{janosi1989heart} & Skip / refer & 6 (Binary) & 1,000 \\
\midrule
\multirow{2}{*}{SEC filings} & Ret-1d, Ret-5d & S\&P 500 SEC filings & Buy / pass & $\le$6 (Terciles) & 3,437 / 3,433 \\
 & Vol-5d, Vol-20d & S\&P 500 SEC filings & Sell volatility / pass & $\le$6 (Terciles) & 3,433 / 3,417 \\
\midrule
\rowcolor{gray!12}\multicolumn{6}{l}{[2] \textit{Controlled environments}} \\
\multirow{2}{*}[-1.1pt]{Synthetic} & Controlled & Four domains & Act / pass & 6--12 (Stated) & 3,000 each \\
 & Multi-state & Four domains & Three actions & 5--13 & 1,000 each \\
\bottomrule
\end{tabular}
\caption{\textbf{Environments}, with SEC filings from 2025--2026 and the controlled domains of equity investment, loan approval, chest-pain triage, and litigation. Checks: Candidate checks per scenario. Test: Number of test scenarios.}
\label{tab:envs}
\end{table*}

\section{Further Related Works}
\label{app:related}

\textbf{Information seeking and retrieval.}
Beyond the works in Section~\ref{sec:related}, LLMs learn to ask clarifying questions through self-improvement \citep{andukuri2024stargate} and interactive elicitation \citep{li2025eliciting}, and game settings test which questions a model asks under a limited budget \citep{grand2026shoot}.
Agentic systems for finance retrieve filings, prices, and news with tools \citep{deng2026alphaquanter,hu2025finsearchcomp,bigeard2025finance}.
These works evaluate whether the retrieved information is correct, not whether acquiring it was worth its cost.

\textbf{Information gain.}
Bayesian experimental design selects experiments by their EIG \citep{chaloner1995bayesian}, variational estimators extend it to complex models \citep{foster2019variational}, and LLM methods select questions by their EIG \citep{hu2024uncertainty,piriyakulkij2023active}.
While these methods value a question by its information content, \diva{} values it by its effect on the decision.

\textbf{Decision-theoretic test selection.}
Early systems for sequential diagnosis chose tests by their expected value \citep{gorry1968experience}, and active feature acquisition learns which feature to acquire with cost-sensitive models, reinforcement learning, or POMDP agents \citep{saartsechansky2009active,shim2018joint,janisch2019classification,contardo2016sequential,rahbar2023efficient,li2025towards}.
Unlike \diva{}, which needs no training, these policies require labeled data from the target domain.
Greedy policies for adaptive submodular objectives, such as equivalence class edge cutting and hyperedge cutting, come close to the lowest cost of identifying the true state or its best decision \citep{golovin2010near,golovin2011adaptive,javdani2014near}, whereas \diva{} maximizes the expected payoff net of costs and stops once no check is worth its cost.
ACTMED simulates patient states with an LLM and selects the test of largest EIG \citep{ruhrberg2025timely}, whereas \diva{} values each check by its decision value.

\textbf{Calibration of elicited probabilities.}
LLMs are partly aware of what they know \citep{kadavath2022language}, and their verbalized confidence can be better calibrated than token probabilities \citep{tian2023just}. For tabular foundation models, adjusting the predictions at test time corrects for a shift in the base rate \citep{lee2026distpfn}.
\diva{}-U instead accounts for the uncertainty in the elicited strengths.

\textbf{LLMs for financial decisions.}
Other benchmarks evaluate question answering over filings and financial reasoning \citep{islam2023financebench,lee2026finstar}, and other studies examine how investor roles and personas shift LLM judgments \citep{ross2026one,jiang2026evaluating}.
Like the works in Section~\ref{sec:related}, they answer from the evidence at hand.

\section{Environments and Data}
\label{app:data}
\begin{table*}[t]%
\centering
\small
\setlength{\tabcolsep}{3pt}\setlength{\aboverulesep}{0pt}\setlength{\belowrulesep}{0pt}\renewcommand{\arraystretch}{1.3}
\renewcommand{\tabularxcolumn}[1]{m{#1}}
\newcommand{\twol}[1]{{\renewcommand{\arraystretch}{1.05}\begin{tabular}[c]{@{}c@{}}#1\end{tabular}}}
\begin{tabularx}{\textwidth}{l|>{\centering\arraybackslash\hsize=0.95\hsize}X>{\centering\arraybackslash\hsize=1.15\hsize}X|>{\centering\arraybackslash\hsize=0.9\hsize}X}
\toprule
\multirow{2}{*}[-1.3ex]{Parameter} & \multicolumn{2}{c|}{\textbf{Knowledge-required}} & \multirow{2}{*}[-1.3ex]{\textbf{Controlled}} \\
\cmidrule(lr){2-3}
 & \twol{Records\\(Taiwan, German, CDC, Heart)} & \twol{SEC filings\\(Return and volatility targets)} & \\
\midrule\midrule
Checks & 6, one uninformative\newline in half of the scenarios & Up to 5 from the filing\newline and 1 uninformative & 6 (10 or 12 in many-item\newline scenarios), 1 uninformative \\
\midrule
Known findings & \multicolumn{3}{c}{0, 1, or 2, with probabilities 0.4, 0.4, and 0.2} \\
\midrule
Payoffs $(g,l)$ & $\{1,2,3\}\times\{2,3,5,8\}$ & $\{1,2,3\}\times\{1,2,3,5,8\}$,\newline with $g/(g{+}l)$ within 0.08\newline of the reference belief & $\{1,2,3\}\times\{2,3,5,8\}$ \\
\midrule
Cost regime $\rho$ & $\{0,0.05,0.1,0.2,0.4\}$ & $\{0,0.005,0.01,0.02,0.04\}$ & $\{0,0.05,0.1,0.2,0.4\}$ \\
\midrule
Cost factor & $\{0.5,1,1,1.5,2\}$ & $\{0.5,1,2\}$ & $\{0.5,1,1,1.5,2\}$ \\
\midrule
Families & -- & -- & Cost variants ($\times0.25$, 1, 4),\newline payoff pairs, many-item \\
\midrule\midrule
VOI $>0$ & 45--74\% & 67--86\% & 83--85\% \\
\midrule
DP acquires & 22--40\% & 55--70\% & 67--71\% \\
\bottomrule
\end{tabularx}
\caption{\textbf{Scenario construction.} Each check costs $\rho\min(g,l)$ times a listed cost factor. Known findings: Outcomes observed before the first decision. VOI $>0$ and DP acquires: Shares of the first 300 test scenarios where some check has positive VOI and where acquiring first is DP-optimal, as ranges over the environments.}
\label{tab:scenarios}
\end{table*}
Table~\ref{tab:envs} summarizes the environments.
The \textit{credit} and \textit{clinical} environments use public tabular records with record attributes as checks, and the \textit{SEC} environments use earnings releases of S\&P 500 companies, where the decision is to buy the stock (Return targets) or to sell volatility, i.e., to bet on small price moves (Volatility targets).
The \textbf{knowledge-required environments} take every decision value from the reference model, and no LLM is involved in any label (Appendix~\ref{app:real}).
The \textbf{controlled scenarios} state every reliability, and their decision values are computed by enumeration (Appendix~\ref{app:envs}).
Besides cost and payoff variants, they include information traps, whose most informative check is not the most valuable, and \textit{already-sufficient} cases, whose known findings settle the decision.
As the stated reliabilities turn the task partly into a calculation, these environments mainly test whether a policy uses the stated numbers correctly.

\subsection{Knowledge-Required Environments}
\label{app:real}
\textbf{Credit and clinical.}
We use the Taiwan credit-card default \citep{yeh2009comparisons} and German credit \citep{hofmann1994statlog} datasets, and the CDC (U.S.\ Centers for Disease Control and Prevention) diabetes indicators \citep{cdc2017diabetes} and UCI heart disease \citep{janosi1989heart,detrano1989international} datasets.
Each feature is binarized with a threshold fixed before looking at outcomes, and uninformative findings are chosen from a pre-declared pool by their lack of association with the outcome on the training records.
For each scenario, the predictions of the reference model are combined into one joint distribution over the checks and the outcome, which keeps every VOI non-negative and the DP solution exact.

\textbf{SEC finance.}
We build firm-quarters of S\&P 500 companies from SEC EDGAR XBRL data available before each earnings release (8-K Item 2.02).
Company names and dates are masked, and test announcements (2025--2026) postdate the training cutoff of most evaluated LLMs.
Outcomes are market-adjusted returns over 1 and 5 trading days and realized volatility over 5 and 20 days after the announcement.
The reference model is a gradient-boosted tree classifier trained on 2015--2022 filings with randomly hidden features to predict from any subset of findings and calibrated on 2023.
Unlike the records, each check's risk is shifted by one log-odds constant such that the imagined beliefs average to the current belief, keeping the first-step VOI exact and non-negative, and the DP treats the checks as independent given the outcome.

\textbf{Scenario construction.}
Table~\ref{tab:scenarios} lists how the scenarios are drawn. SEC checks cost less, as they carry less value of information, and the SEC payoffs place the threshold near the reference belief, which favors acquiring (Appendix~\ref{app:f8}).

\subsection{Controlled Environments}
\label{app:envs}
\textbf{Controlled binary environments.}
Each scenario has a binary hidden state with base rate $\pi\in\{0.10,0.15,\ldots,0.60\}$, with its other parameters and scenario families in Table~\ref{tab:scenarios}.
Each check has $P(\text{finding}|Y{=}1)=s_1$ and $P(\text{finding}|Y{=}0)=s_0$ on a $0.05$ grid, drawn from six kinds: Strong, moderate, weak, one-sided high (The finding is common in both states), one-sided low (Rare in both states), and uninformative ($s_1=s_0$).
All exact quantities (Posterior, VOI, EIG, and the DP value) are enumerated, and each scenario is kept only if an LLM can read its defining numbers from its text.

\textbf{Multi-state environments.}
To test \diva{} beyond a binary state and two actions, these environments have a hidden state of three factors (12--18 combinations) and three actions.
The payoff of each action is a sum of contributions from the factors, and each check reveals one factor with two or three outcomes.
Some factors do not affect the payoff (E.g., the revenue surprise is already priced in over a three-month horizon), making checks on them informative but of zero decision value.

\section{Method Details}
\label{app:method}
\label{app:notation}Table~\ref{tab:notation} lists the notation and terms of the paper.
\input{tables/notation}

\subsection{Checks, Costs, and Elicitation}
\label{app:interface}
\textbf{Candidate checks.}
Every scenario lists a closed menu of candidate checks, each revealing one outcome of the case, e.g., whether the client's most recent bill is two or more months overdue.
Neither \diva{} nor any baseline invents questions, and a stop option ends the acquisition.
Check costs and the payoff table are stated in the scenario text, and \diva{} reads them rather than asking the LLM.

\textbf{Charged actions.}
Only acquiring a check, i.e., revealing a fact about the case, costs points and enters the regret.
As the elicitation questions ask for general knowledge, not for facts about the case, they are not charged, like any baseline's reasoning.

\afterpage{%
\begin{table}[t]
\centering
\small
\setlength{\tabcolsep}{3pt}
\begin{tabular*}{\linewidth}{l|l@{\extracolsep{\fill}}r@{}}
\toprule
Parameter & Value & \\
\midrule
\multicolumn{3}{l}{\textit{Selected on development data}} \\
Elicitation form & Outcome-conditioned risks & \\
Shrinkage $\lambda$ & $0.75$ of $\{1, 0.75, 0.5\}$ & \\
Robust stopping & None of $\{$none, $q\in\{0.5,0.25,0\}\}$ & \\
Uncertainty scale $\sigma$ & $1$ of $\{0, 1, 1.5, 2, 3\}$ & \\
\midrule
\multicolumn{3}{l}{\textit{Fixed}} \\
Quadrature & Three-point Gauss--Hermite & \\
Clipping & $[0.01, 0.99]$ & \\
Failed answer & $0.5$ (Stated base rate for $\hat\pi$) & \\
\bottomrule
\end{tabular*}
\caption{\textbf{Configuration of \diva{} and \diva{}-U.} With stated reliabilities, they use $\lambda=1$ and $\sigma=0$.}
\label{tab:config}
\end{table}\input{tables/spec_example}}%
\textbf{Elicitation protocol.}
In the selected form (Binary checks and outcome-conditioned risks), \diva{} asks one question for the base rate and three per candidate check and known finding, and binned checks ask for each bin's risk.
The questions are:
\begin{itemize}[leftmargin=0.4cm,itemsep=0pt,topsep=2pt]
\item[] \textit{Q1}: ``\textit{Based on the information above and your general knowledge, among cases where the check `[name]' shows this finding ([finding]), what share end up in the case where [adverse case]?}''
\item[] \textit{Q2}: The same question for cases that do NOT show the finding.
\item[] \textit{Q3}: ``\textit{Among [population], what share would show this finding?}''
\end{itemize}
Each question is a separate call that sees the scenario but no other answer, with the answer prefilled with ``\textit{Answer:}'' and limited to six greedily decoded tokens of a single percentage.
All $1+3|\mathcal{C}\cup\mathcal{K}|$ calls are made once before the first acquisition, including those for the known findings, which set the current belief, and each observed outcome then updates the belief by Bayes' rule without asking again.

\subsection{Configuration}
\label{app:know}
\textbf{Elicitation forms.}
Instead of the outcome likelihoods $\hat\ell_e(v|y)$, the LLM can state the outcome-conditioned risk $\hat r_e(v)\approx P(Y{=}1|e{=}v)$, which is easier to state for quantile-binned checks (E.g., the top third of a financial ratio). This risk is converted to likelihoods with the bin frequency $P(e{=}v)$ by Bayes' rule.
For a binary check, the frequency $\hat P(f)$ of the finding is also elicited.
Eliciting the decision model costs $1+2|\mathcal{C}\cup\mathcal{K}|$ short calls for binary checks with likelihoods and $1+3|\mathcal{C}\cup\mathcal{K}|$ with outcome-conditioned risks, once per episode.

\input{tables/models}
\begin{algorithm*}[t]
\small
\caption{\textbf{Pseudocode of \diva{}.}}
\label{alg:diva}
\begin{algorithmic}[1]
\Require LLM $f$; scenario $x$ with payoffs $g,l$, known findings $\mathcal{K}$, and checks $\mathcal{C}$ with costs $c_e$; short numeric questions $q_{\mathrm{base}}$ and $q_e$ for the base rate and each check (Appendix~\ref{app:interface}); shrinkage $\lambda$; uncertainty scale $\sigma$
\Ensure Final action $a\in\{\textsc{act},\textsc{pass}\}$
\Statex \textit{// 1) Model specification: Asked once per episode}
\State $\hat\pi\gets f(x,q_{\mathrm{base}})$; \ $\hat\ell_e(v|y)\gets f(x,q_e)$ for every $e\in\mathcal{K}\cup\mathcal{C}$ \Comment{One short numeric answer per question}
\State Clip every answer and likelihood to $[0.01,0.99]$ and keep a fraction $\lambda$ of each likelihood pair's deviation from its mean
\Statex \textit{// 2)--4) Imagine, value, and stop: No further LLM calls}
\Loop
  \State $\hat b\gets$ Belief from $\hat\pi$ and the observed findings by Bayes' rule
  \State $\widehat\Delta(e)\gets\mathbb{E}_{z}\big[\sum_{v}\hat P(v)\,V^*(\hat b_{e,v})\big]-V^*(\hat b)-c_e$ for every $e\in\mathcal{C}$ \Comment{\diva{}: $z{=}0$; \diva{}-U: $z\sim\mathcal{N}(0,1)$}
  \If{$\mathcal{C}=\emptyset$ \textbf{or} $\max_e\widehat\Delta(e)\le 0$} \State \Return $\arg\max_a\mathbb{E}_{\hat b}[U(a,Y)]$ \Comment{Value-based stopping}
  \EndIf
  \State Acquire $e^\star=\arg\max_e\widehat\Delta(e)$, observe its outcome, and move it from $\mathcal{C}$ to $\mathcal{K}$
\EndLoop
\end{algorithmic}
\end{algorithm*}
\textbf{Configuration selection.}
Without stated reliabilities, the elicited decision model is noisier, and as $V^*$ is convex, noisy likelihoods inflate the estimated decision value and cause over-acquisition.
To limit this, we select from 36 configurations on development splits only: 1)~the \textit{elicitation form} (Outcome likelihoods, outcome-conditioned risks, or their average), 2)~a \textit{shrinkage factor} $\lambda\in\{1,0.75,0.5\}$ that moves each elicited likelihood pair toward its mean, and 3)~no robust stopping or a \textit{robust stopping rule} that requires the net value to be positive under the $q\in\{0.5,0.25,0\}$ quantile of five sampled decision models.
Outcome-conditioned risks with $\lambda=0.75$ and no robust stopping are best over 26 LLM--environment development cells and frozen before testing (Table~\ref{tab:config}).

For binned SEC checks, asking for the risk of each bin separately often collapses the answers to the base rate. A joint elicitation of all bins, a contrastive elicitation, and a hybrid that defers to the direct policy each helped only some targets and were not adopted. \diva{}-U was added after this configuration was frozen (Appendix~\ref{app:prereg}).

Clipping to $[0.01,0.99]$ avoids degenerate likelihood ratios, and a failed answer, under 0.1\% of answers in every environment, falls back to $0.5$ (The stated base rate for the base-rate question).

\textbf{Uncertainty-aware valuation.} Averaging over the uncertainty has two effects: 1)~It can lower the value of strong checks, as VOI saturates at the \textit{value of perfect information}, i.e., the value of learning $Y$ exactly, and 2)~it can value a check of no actual strength, which the cost term limits to cheap checks. We compare it with a multiplicative cost discount and an additive acquisition bonus tuned on development data (\diva{}-$\kappa^\ddagger$ and \diva{}-$\beta^\ddagger$). The same $\sigma=1$, with $\lambda=0.5$, is selected when holding out each data source (Appendix~\ref{app:knowledge}).

\textbf{Perturbation and quadrature.}
\diva{}-U shifts the log-odds of $\hat\ell_e(1|1)$ by $+\sigma z/2$ and those of $\hat\ell_e(1|0)$ by $-\sigma z/2$, which changes the strength of the check by $\sigma z$, and treats this error as $z\sim\mathcal{N}(0,1)$. For a check with ordered bins, the log-odds of the elicited risks of the lowest and highest bins are shifted by $-\sigma z$ and $+\sigma z$ (Linearly in between), and the shifted risks become likelihoods normalized over bins, which keeps the belief unchanged.
As $V^*$ is piecewise linear, the expected net value over $z$ has no closed form, and we use the three-point Gauss--Hermite rule with nodes $z\in\{-\sqrt{3},0,\sqrt{3}\}$ and weights $\{1/6,2/3,1/6\}$, which is exact for polynomials of degree five.
As the three evaluations reuse the same answers, \diva{}-U needs no additional LLM call.

\subsection{Pseudocode}
\label{app:alg}
Algorithm~\ref{alg:diva} details every step of \diva{} and \diva{}-U, and Table~\ref{tab:spec_example} traces these steps on one check, both with and without stated reliabilities.%

\subsection{Models and Hardware}
\label{app:models}
Table~\ref{tab:models} lists the checkpoint, precision, and license of every LLM. We serve them with vLLM on NVIDIA L40 and L40S GPUs (48\,GB each), with tensor parallelism over two GPUs for Gemma-2-27B, Llama-3.3-70B, and Qwen2.5-72B, and greedy decoding. The reference models and the DP optima run on CPUs and involve no LLM.%

\section{Experimental Setup}
\label{app:setup}
\subsection{Evaluation Metrics}
\label{app:metrics}
With $Q(a)$ the expected utility of the first action $a$, i.e., acquiring a check $e$ or stopping, at belief $b$, we define three regrets as follows (Lower is better):
\begin{itemize}[leftmargin=0.4cm,itemsep=0pt,topsep=2pt]
\item \textit{DP regret}: The regret of action $a$,
\[R_{\mathrm{DP}}(a)=\frac{100}{l}\Big(\max_{a'}Q_{\mathrm{DP}}(a')-Q_{\mathrm{DP}}(a)\Big),\]
where $Q_{\mathrm{DP}}(a)$ is the value of taking $a$ and acting optimally afterwards, computed by exact dynamic programming over the remaining checks.
\item \textit{Myopic regret}: The regret $R_{\mathrm{my}}(a)$, defined in the same way with $Q_{\mathrm{my}}(\textsc{stop})=V^*(b)$ and $Q_{\mathrm{my}}(e)=\sum_{v}P(v|E)\,V^*(b_{e,v})-c_e$, i.e., the value of deciding right after one acquisition.
\item \textit{Episode regret}: The regret of a policy $\mu$,
\[R_{\mathrm{ep}}(\mu)=\frac{100}{l}\big(\mathbb{E}[U(\mu^*)]-\mathbb{E}[U(\mu)]\big),\]
where $U(\mu)$ is the utility of $\mu$ net of all acquisition costs and $\mu^*$ is the DP-optimal policy, normalized by the payoff range $g+l$ instead of $l$ on the controlled environments.
\end{itemize}
In the multi-state setting, all regrets are normalized by half the payoff range instead of by $l$.

We also report the over-acquisition rate (Acquiring when stopping is optimal), the information-trap rate (Appendix~\ref{app:behavior}), and clustered bootstrap confidence intervals (CIs) (Appendix~\ref{app:stats}).
For episodes, we report the episode regret with the same random draws for every policy (Appendix~\ref{app:episodes_full}) and the exact episode regret under the reference model on the eight knowledge-required environments.

\textbf{Incoherence probe.}\label{app:incoh}
On the 496 controlled scenarios of the probe, each LLM reports its current belief $\tilde b$ and its belief $\tilde b_{e,v}$ after each imagined finding $v$ of each check $e$.
\begin{table*}[t]
\centering
\small
\setlength{\tabcolsep}{4pt}
\begin{tabularx}{\textwidth}{l|l|X}
\toprule
Category & Method & How the first action is chosen \\
\midrule
\multirow{7}{*}{Prompting} & Direct & The LLM answers with one option letter (A check or stop) without rationale, and the letter of highest probability, averaged over option orders, is taken. \\
 & CoT~\citep{wei2022chain} & The LLM reasons step by step about what it still needs, given the costs, and ends with ``ANSWER: X'' (Up to 1,024 tokens). \\
 & SC~\citep{wang2023selfconsistency} & Majority vote over five CoT samples (One request, temperature 0.7). \\
 & ReAct~\citep{yao2023react} & The LLM writes one Thought and one Action, either check[X] or a final decision (Up to 512 tokens per step). \\
\midrule
\multirow{8}{*}{\begin{tabular}[c]{@{}l@{}}Decision-aware\\prompting\end{tabular}} & Ask-P~\citep{zhong2026finriskatlas} & The LLM chooses between proceeding and asking for one more costly item, and if it asks with probability $\ge 0.5$, it acquires the item it names (Averaged over three orders), else it stops. \\
 & VoI-P & The LLM states each check's net value in JSON after brief reasoning, and the largest positive one is acquired. \\
 & VoI-Ask~\citep{dong2026value} & The LLM states in JSON its belief, the result distribution of every check, and its belief after each simulated result, and the check with the largest positive value of information net of cost is acquired. \\
\midrule
\multirow{2}{*}{Information gain} & \begin{tabular}[t]{@{}l@{}}BED-LLM\\\citep{choudhury2026bedllm}\end{tabular} & From the LLM's belief and result distributions under each state, it acquires the check of largest EIG if at least 0.05 bits, ignoring cost. \\
\midrule
\multirow{4}{*}{Program-aided} & PAL~\citep{gao2023pal} & The LLM writes a Python program that maximizes the expected payoff net of costs from the numbers in the text, and its printed choice is taken. \\
 & VoI-Tool$^{*}$ & The LLM fills in the probabilities for a correct VOI calculator in the prompt, which acquires the check with the largest positive net value. \\
\midrule
\multirow{2}{*}{Reference} & Never & It stops and decides at once. \\
 & Always ask & It acquires every check, in the order of \diva{}'s net values. \\
\bottomrule
\end{tabularx}
\caption{\textbf{Baseline methods.} How each baseline of Table~\ref{tab:know} chooses its first action. $^{*}$: Only in Appendix~\ref{app:prior}.}
\label{tab:baselines}
\end{table*}%
Figure~\ref{fig:incoh}b counts, on the scenarios where acquiring is optimal, how often $\arg\max_e\big[\sum_v P(v|E)\,V^*(\tilde b_{e,v})-V^*(\tilde b)-c_e\big]$ with the exact predictive probability $P(v|E)$ is an exact best check (Ties share credit).
Chance is the expected share of a random check, and \diva{} with stated reliabilities attains 1 by construction.

For Section~\ref{sec:why} and Figure~\ref{fig:incoh}a, the belief change (Swing) of a check is the difference $\tilde b_{e,1}-\tilde b_{e,0}$ of its beliefs after a positive and a negative finding, for the LLM and for the exact Bayesian counterpart.
The per-LLM correlation of the two swings ranges from \IncohCorrMin{} to \IncohCorrMax{}.
Pooled over the \NumIncoh{} LLMs, exact belief changes are \IncohShrink{} times larger in magnitude than elicited ones, and the gap $|P(e{=}1|E)\,\tilde b_{e,1}+P(e{=}0|E)\,\tilde b_{e,0}-\tilde b|$ between the average imagined belief and the current belief, which is zero for coherent beliefs, averages \IncohTP{} percentage points across all probed checks.

\subsection{Baseline Methods}
\label{app:baselines}
\textbf{Shared protocol.}
Every method sees the same scenario text with a stop option, and reliabilities are stated only in the controlled environments.
Native reasoning is disabled except for the native-reasoning baseline, decoding is greedy unless stated otherwise, and letter probabilities are read from the first answer token.
To reduce position bias \citep{zheng2024large,pezeshkpour2024large}, the direct policy averages over three option orders on the knowledge-required environments, and the other letter queries average over the listed order and its reverse.
An answer still unparsable after two resamples falls back to stopping. Table~\ref{tab:baselines} lists how each method picks its first action.

\textbf{Other baselines.}
Some appendix tables also include the following methods, each followed by the table or appendix section where it appears.
\begin{itemize}[leftmargin=0.4cm,itemsep=0pt,topsep=2pt]
\item \textit{Prompting}: Native reasoning (Reason.), where reasoning models think before the direct choice, with the stop option rotated over the three positions across scenarios (Appendix~\ref{app:full_ctrl_llm}), and Salience and Uncertainty, where the LLM picks the check that seems most important or that it is most uncertain about (Appendix~\ref{app:prior}).
\item \textit{Information gain}: EIG (Beliefs), a fixed EIG threshold on the LLM's beliefs (Appendix~\ref{app:prior}).
\item \textit{Reference}: Random picks a random check (Appendix~\ref{app:prior}), and Always ask $\rightarrow$ \diva{} takes the first decision of Always ask and then follows \diva{} (Appendix~\ref{app:episodes}). In Appendix~\ref{app:knowledge}, Generic \diva{} and Generic \diva{}-U give every check the same strength, Knowledge-free \diva{} and Knowledge-free \diva{}-U shuffle the elicited likelihoods across checks, Supervised ref.\ uses likelihoods estimated from training records, and Threshold heur.\ acquires the cheapest check only when the belief is close to the decision threshold. \diva{}-$\kappa^\ddagger$ and \diva{}-$\beta^\ddagger$ acquire a check when its value exceeds $\kappa c_e$ or $c_e-\beta$, with $\kappa$ and $\beta$ tuned on development data (Appendix~\ref{app:stop}).
\item \textit{Ablation}: \diva{} (Elic.), which also elicits the multi-state payoff contributions (Appendix~\ref{app:ms}).
\end{itemize}
\begin{figure*}[t]
\centering
\includegraphics[width=\textwidth]{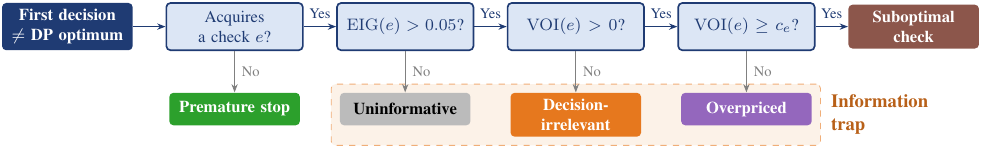}
\caption{\textbf{Error types of a first decision.} A first decision that differs from the DP-optimal one is classified by sequential tests: Each test either assigns an error type (Down) or passes to the next test (Right).}
\label{fig:taxonomy}
\end{figure*}%
\input{tables/m_know_detail}\textbf{Ablation variants.}
The ablations are not baselines, as they replace the decision value of \diva{} by another rule on the same elicited decision model.
\begin{itemize}[leftmargin=0.4cm,itemsep=0pt,topsep=2pt]
\item \textit{\diva{} w/ EIG}: It acquires the check with the largest EIG about the outcome if it is at least 0.05 bits, regardless of cost, and stops otherwise.
\item \textit{\diva{} w/ EIG$^\ddagger$}: It acquires the check with the largest $\mathrm{EIG}-\eta c_e/l$ if it exceeds $\tau$ and stops otherwise. The pair $\eta\in\{0,0.02,\ldots,20\}$ and $\tau\in\{0,0.01,0.02,0.05,0.1,0.2\}$ minimizes the development DP regret of the same LLM and environment, with $\eta=0.1$ and $\tau=0.02$ where development data are missing.
\item \textit{\diva{} w/ ES} and \textit{\diva{} w/ ES$^\ddagger$}: Entropy search over the optimal action \citep{truong2025gathering} with a fixed or a tuned threshold (Appendix~\ref{app:ms}).
\end{itemize}

\subsection{Table Conventions}
\label{app:notes}
\begin{itemize}[leftmargin=0.4cm,itemsep=0pt,topsep=2pt]
\item All regrets are $\times$100 and lower is better, the best results are in bold and the second best are underlined, and our methods are shaded.
\item Every first-decision table uses the first 300 test scenarios of each environment, every episode table test scenarios 0--149 of the knowledge-required environments and the first 100 of each controlled domain, and every method the same 20 LLMs, unless stated otherwise.
\item A cell ``$a$ / $b$'' reports the regret against the DP / myopic optimum.
\item $^\ddagger$ marks methods whose cost trade-off, threshold, cost factor $\kappa$, or bonus $\beta$ is tuned per LLM and environment on development scenarios with ground-truth decision values, whereas \diva{} and \diva{}-U use a single configuration.
\item LLM baselines are the prompting, decision-aware prompting, information gain, and program-aided groups, and native reasoning (Reason.) applies only to the nine reasoning-capable LLMs (Qwen3, gpt-oss, DeepSeek-R1).
\item Stop is the rate of stopping at the first decision, Stop opt.\ the rate at which stopping is DP-optimal, Acq.\ the expected number of acquired checks, and Uninformative the rate of acquiring the uninformative check.
\end{itemize}

\section{Additional Results}
\label{app:more}
Tables follow the conventions of Appendix~\ref{app:notes}.
\subsection{Acquisition and Stopping Behavior}
\label{app:stop}
Figure~\ref{fig:taxonomy} classifies every first decision that is not DP-optimal into the error types used below.

\textbf{Stopping behavior.}
Table~\ref{tab:know_detail} reports the stopping rates of \diva{}, \diva{}-U, and two variants tuned on development data, \diva{}-$\beta^\ddagger$ and \diva{}-$\kappa^\ddagger$ (Appendix~\ref{app:baselines}), together with the stop rate of VoI-Ask.
On the tabular records, \diva{} stops more often than the DP-optimal policy and \diva{}-U less often, while on SEC filings \diva{}-U rarely stops.
Without any tuning, \diva{}-U comes close to the tuned \diva{}-$\beta^\ddagger$ in most environments.%

\begin{figure}[t]%
\centering
\includegraphics[width=\columnwidth]{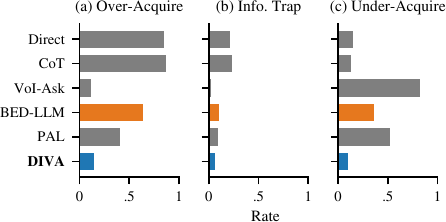}
\caption{\textbf{Acquisition behaviors} on controlled environments. Gray: LLMs, orange: BED-LLM, blue: \diva{}.}
\label{fig:behavior}
\end{figure}
\input{tables/m_episodes}\textbf{Over-acquisition when the evidence suffices.}
Figure~\ref{fig:desiderata} pools the first decisions where stopping is DP-optimal over the 20 LLMs and four controlled domains. Among them, the LLM baselines and EIG with a fixed threshold on \diva{}'s decision model acquire a check in most cases, whereas \diva{} acquires one in only a few cases.

\textbf{Acquisition behaviors.}\label{app:behavior}
Figure~\ref{fig:behavior} reports the over-acquisition, information-trap, and under-acquisition rates of LLM policies, BED-LLM, and \diva{}. The direct policy and CoT have the highest over-acquisition and information-trap rates, and VoI-Ask and PAL the highest under-acquisition rates, while \diva{} is the only policy with low rates of both over- and under-acquisition.
Uninformative checks cause \RecDivaUPlacebo\% of the remaining regret of \diva{}-U on the tabular records and \RecDivaPlacebo\% of that of \diva{}, as averaging over the uncertainty gives weak checks a positive value (Section~\ref{sec:divau}).

\input{tables/m_controls}\subsection{Full Episodes}
\label{app:episodes}
\textbf{Per-environment episodes.}\label{app:episodes_full}
Table~\ref{tab:episodes_full} reports the episode regret of Table~\ref{tab:episodes} for every environment, with the wins of \diva{}-U over LLM baselines.
In the controlled environments, it uses the first 100 test scenarios per domain: Every policy runs in the same two pre-sampled worlds per scenario, each fixing the hidden state and the outcome of every check.
The regret of an episode is the sum of the \textit{advantage losses}, i.e., the value lost at each step relative to the DP-optimal choice, over acquisition, stopping, and final decisions, normalized by $g+l$.

In the knowledge-required environments, Table~\ref{tab:episodes_all} takes the final action from the belief of the reference model and evaluates the \diva{} variants, the tuned baselines, and the reference policies exactly under the reference model on test scenarios 0--149.
The LLM baselines re-query the LLM at every step and are evaluated on \NumWorlds{} per scenario from the reference model, an unbiased estimate of the same expected episode regret.%

\textbf{Episode results.}
In the controlled environments, \diva{}-DP has the lowest and \diva{} the second lowest episode regret in every domain, and \diva{}-U beats every LLM baseline for all 20 LLMs.
On the tabular records, \diva{}-U beats every LLM baseline for at least half of the LLMs on every dataset, while on SEC filings its fixed uncertainty makes it over-acquire and fall behind never acquiring.

\textbf{Knowledge-required episodes.}
The DP and myopic optima agree on the best action in \GoldSameMin--\GoldSameMax\% of the scenarios where both acquire and differ mainly on cheap checks whose value arises only with later checks, motivating the episode regret.
On test scenarios 300--999 with the 18 pre-registered LLMs (Appendix~\ref{app:prereg}), \diva{}-DP has lower episode regret than never acquiring in \KfourA{} of the 8 environments and in \KfourC{} of \KfourCells{} LLM--environment cells. The paired 95\% CI of the difference excludes zero in six of the eight environments (Paired bootstrap, Appendix~\ref{app:stats}).
Planning with the same decision model and no new LLM call, \diva{}-DP also has lower regret than \diva{} in every environment.

\subsection{What Drives the Gains}
\label{app:knowledge}
\textbf{Contribution of the LLM's knowledge.}
Tables~\ref{tab:const} and~\ref{tab:controls} keep the computation of \diva{} fixed and replace the LLM's decision model by controls that shuffle the elicited likelihoods across checks (Knowledge-free) or give every check one strength selected on the development splits (Generic).
Both keep most of the gain of \diva{} over never acquiring on the tabular records, indicating that \diva{} is robust to weak LLM knowledge, as the gain comes mainly from the decision value.
The elicited strengths still matter, as the unshuffled \diva{} is significantly better in \KFWorse{} of \KFCells{} cells.
With each check's direction set by design, generic \diva{}-U stops more often and beats \diva{}-U in seven of the eight environments, whereas with stated reliabilities the LLM's strengths help (Section~\ref{sec:analysis}).

Supervised ref., with likelihoods from training records, shows the room left by elicitation, whereas Threshold heur.\ does not beat never acquiring.

\input{tables/abl_eig}
\textbf{Decision value vs.\ EIG.}\label{app:abl_eig}
Table~\ref{tab:abl_eig} replaces the decision value of \diva{} by EIG about the outcome on the same elicited decision model, on the full test sets of the knowledge-required environments.
With a fixed threshold and no cost term, EIG has higher regret than \diva{} in every environment and more than twice its regret on average.
Tuning its cost trade-off and threshold per LLM and environment on labeled development scenarios makes it better than \diva{}, and better than \diva{}-U on Taiwan, German, CDC, and Ret-5d.
\diva{}-U, which uses no such labels, still has the lowest regret on average and on Heart and three of the four SEC targets.%

\input{tables/transfer}
\textbf{Transfer without target labels.}
We hold out each data source (Taiwan, German, CDC, Heart, and SEC filings) in turn and select the shrinkage and the uncertainty scale of \diva{}-U on the development data of the other sources only.
The same configuration, $\lambda=0.5$ and $\sigma=1$, is selected for every held-out source.
On the first 400 test scenarios of each environment (Table~\ref{tab:transfer}), it has lower average regret than both our configuration ($\lambda=0.75$) and \diva{} without uncertainty ($\sigma=0$).%

\input{tables/ms}

\subsection{Multi-State Environments}
\label{app:ms}
In the multi-state environments, \diva{} elicits the current probability of every level of every hidden factor and the outcome distribution of every check given the level of the factor it reports on. It takes the payoff table directly from the task specification, as for $g$ and $l$ in the binary environments.
After an interim look at test scenarios 0--499, we report scenarios 500--999 as held-out results and the variant that elicits payoffs from the LLM as an ablation.%

Table~\ref{tab:ms} reports the held-out results (Scenarios 500--999 per domain), where \diva{} has far lower regret on average than the direct policy, never acquiring, and Always ask, which here acquires \diva{}'s top check. Replacing its decision value with EIG or entropy search over the optimal action on the same decision model increases the regret, even when their thresholds are tuned on development data.
With verbal beliefs, \diva{} still improves over the direct policy, whereas eliciting the payoffs from the LLM makes \diva{} no better than the direct policy, indicating that the payoffs must also be correct.%

\subsection{Other Metrics}
\label{app:realized}
\textbf{Realized utility.}
As the decision values of the knowledge-required environments come from a reference model, Table~\ref{tab:realized} replays the policies on the real records, where each check reveals the record's actual value. The final action, taken from the policy's own belief, is scored with its actual outcome (Normalized by $l$, net of costs, with clustered bootstrap CIs of the difference to never acquiring).
On the scenarios of Table~\ref{tab:know}, even the reference model's DP policy does not significantly beat never acquiring on three of the four datasets. We therefore also replay 5,000 Taiwan and 5,000 CDC test scenarios from distinct records, where it does.
The direct policy realizes a far lower utility than never acquiring, while \diva{} and \diva{}-U stay close to it without a reliable gain, significantly worse in more LLM--dataset cells than better, many of them on Heart, where the decision model is poorly calibrated.%

\input{tables/m_realized}
\input{tables/main_know_myo}
\input{tables/m_reference}
\textbf{Results under all metrics.}\label{app:allmetrics}
Table~\ref{tab:know_myo} reports the myopic regret of every method on the scenarios and LLMs of Table~\ref{tab:know}, and Table~\ref{tab:episodes_full} reports the corresponding episode regret per environment.%

\textbf{Reference models.}
Table~\ref{tab:alt_ref} re-scores every method under two alternative references, and Table~\ref{tab:ref_metrics} validates the reference models.
On held-out test records, every reference model has a lower Brier score than a constant base-rate prediction, with the lowest AUCs on SEC filings and the largest calibration errors on the two volatility targets, whose decision values are the least reliable.
The first alternative does not treat the findings as independent: It is a logistic regression with all pairwise interactions of the findings on the records and a joint model over all offered checks on SEC filings, and the second is a naive Bayes model on the same training records.
Under both, the ranking of the methods is preserved, and \diva{}-U has lower regret than every baseline in all eight environments under the first and than never acquiring in all but two LLM--dataset cells under the second.%

\textbf{Fresh scenarios.}
Table~\ref{tab:fresh} evaluates every method on fresh tabular scenarios drawn with an independent seed after all design decisions, including \diva{}-U, were fixed. For Taiwan and CDC, they come from test records that no earlier scenario used, and for German and Heart, whose test splits have 200 and 184 records, from the same records with fresh draws of checks, costs, payoffs, and known findings. On SEC filings, the held-out slice is the test scenarios beyond the first 1,000 of each target, reported in Appendix~\ref{app:heldout} for the main policies.%

\input{tables/m1_decomp}
\subsection{Further Analyses}
\label{app:further}

\textbf{Ablation details.}
Table~\ref{tab:ablation} uses the 496 controlled scenarios of the incoherence probe: Direct and the imagined-belief row use each LLM's probe answers, the 11-option row the same questions answered with option letters, and \diva{} its numeric elicitation. A missing answer counts as stopping.

\textbf{Check selection vs.\ stopping.}
Table~\ref{tab:m1_decomp} combines the stopping rule of one policy with the check selection of another on the first decision.
Giving the direct policy the stopping rule of \diva{}-U lowers its regret far more than giving it the check selection of \diva{}-U, except on SEC filings, where both lower it equally.
Under every stopping rule, the check of \diva{}-U has lower regret than the LLM's choice and the largest-EIG check.
This indicates that when to stop accounts for most of the gain and which check to acquire accounts for the rest.%

\section{Robustness and Diagnostics}
\label{app:robust}

\subsection{Additional Baselines}
\label{app:prior}
\input{tables/m_extra}

\textbf{Baselines.}
DeLLMa~\citep{liu2025dellma} forecasts the unknown state, fits a Bradley--Terry model to the LLM's ranking of state-action pairs as utilities, and takes the action with the largest expected utility, where the actions are acquiring each remaining check or stopping.
Uncertainty of Thoughts (UoT)~\citep{hu2024uncertainty} asks the question with the largest entropy reward accumulated over a simulation tree of possibilities and stops once every remaining possibility has the same label or no question reaches the 0.05-bit threshold of BED-LLM. As originally designed, its reward ignores check costs.
We do not include the related BIRD~\citep{feng2025bird}, as it outputs a probability without a rule for acquiring or stopping.
VoI-Tool, Salience, Uncertainty, EIG (Beliefs), and Random follow Appendix~\ref{app:baselines}, the direct policy and Ask-P get a per-LLM stop threshold by cross-fitting, and a cost-aware EIG rule uses the elicited decision model with one trade-off tuned on development data.

\textbf{Results.}
Table~\ref{tab:prior} reports all these baselines with the same 20 LLMs: \diva{}-U has lower regret than DeLLMa and UoT for every LLM in every environment, and lower regret than every other added baseline except the cost-aware EIG rule in all twelve environments.
DeLLMa and UoT stop at rates unrelated to how often stopping is optimal, VoI-Tool lowers the regret of the LLM in the controlled environments but stays above \diva{}, and tuned stopping moves the direct policy and Ask-P toward never acquiring.
The cost-aware EIG rule has lower regret than \diva{} on the tabular records and most SEC targets, while \diva{}-U has lower regret than it in seven of the eight knowledge-required environments. This indicates that cost-aware stopping on the elicited decision model carries much of the gain of \diva{}, consistent with Table~\ref{tab:m1_decomp}.%

\subsection{Further Controls}
\label{app:f8}

\input{tables/f8_mis}
\textbf{Misspecified payoffs and costs.}
Table~\ref{tab:f8_mis} lets \diva{} decide with $g$, $l$, and every cost multiplied by independent errors of up to $\pm$25\% or $\pm$50\% and scores the decisions with the true values.
Such errors raise the regret of \diva{} and \diva{}-U only moderately.
With $\pm$50\% errors, \diva{}-U still beats every LLM baseline in seven of the eight knowledge-required environments, all but Vol-20d, as does \diva{} in all four controlled domains.%

\input{tables/predvoi}
\input{tables/confirm}
\input{tables/sec_nofilter}
\textbf{SEC payoffs without the threshold filter.}
The SEC scenarios keep only payoffs whose threshold lies within 0.08 of the reference belief (Appendix~\ref{app:real}), which favors acquiring.
As the elicitation questions do not involve the payoffs, Table~\ref{tab:sec_nofilter} re-scores the stored decision models with payoffs drawn uniformly from the grid or only far from the reference belief.
With uniform payoffs, stopping is DP-optimal in most scenarios and \diva{}-U stops far less often, but it still beats never acquiring and Always ask on every target, while \diva{} and \diva{}-DP are best on the two return targets.
Far from the reference belief, \diva{} comes close to never acquiring, whereas \diva{}-U keeps acquiring and beats it on both return targets, so its advantage over \diva{} there is specific to decisions near the threshold.%

\textbf{Coherent predictive VOI.}
\diva{}-U averages the decision value over possible strengths, which values each check as if acquiring it also revealed its true strength. A coherent alternative, \diva{}-PV, first averages the likelihoods over the same strengths and then computes one decision value, which is never larger, as the decision value is convex in the likelihoods.
Table~\ref{tab:predvoi} evaluates \diva{}-PV with the same decision model, shrinkage, scale, and quadrature: At the first decision, it stays close to \diva{} and far above \diva{}-U, whereas over full episodes it has the lowest average regret in the table.
This indicates that the first-decision gain of \diva{}-U stems from an upward correction that offsets the premature stops of \diva{}, not from a coherent treatment of uncertainty, and that the same correction causes its over-acquisition in episodes.%

\subsection{Held-Out SEC Filings}
\label{app:heldout}
Table~\ref{tab:confirm} reports the SEC filings beyond the first 1,000 test scenarios of each target, where \diva{}-U has lower DP regret than \diva{} on three targets and equal regret on Ret-5d. On the volatility targets, it has the lowest DP regret of all policies, while the direct policy has the lowest myopic regret.%

\textbf{Pre-registration.}\label{app:prereg}
We fixed the main hypotheses, and every later addition to them, before scoring the corresponding test data.
The deviations are: The primary metric became DP regret instead of the pre-registered myopic regret (Both reported), a verbal-reliability hypothesis was withdrawn, and after interim inspections of the SEC, multi-state, and knowledge-required test results we tried redesigns on development data, of which only \diva{}-U and \diva{}-DP were adopted, each fixed before scoring on new test data.
On uninspected SEC filings, \diva{}-U has lower regret than \diva{} in \KthreeSecA{} of \KthreeSecN{} LLM--target cells and than never acquiring in \KthreeSecB{} (Averages in Table~\ref{tab:confirm}), whereas its results on the tabular records re-score inspected decision models and are \textit{post hoc}.
For \diva{}-DP, all four hypotheses on test scenarios 300--999 hold for the 18 pre-registered LLMs (Appendix~\ref{app:episodes}).

\subsection{Statistical Tests}
\label{app:stats}
All methods are scored on identical scenarios, and 95\% CIs of the mean regret difference use a clustered bootstrap over scenario groups with 1,000 resamples.
Table~\ref{tab:ci_know} tests, per LLM, the wins of \diva{}-U in Table~\ref{tab:know} against the best LLM baseline: The wins are significant for most LLMs on the tabular records and for fewer on the SEC targets.
It also reports the CIs of the differences of \diva{}, \mbox{\diva{}-U}, and \diva{}-DP from never acquiring.%
\input{tables/m_significance}

\subsection{Prompt Paraphrases}
\label{app:para}
To assess prompt sensitivity, we re-elicit the decision model of \diva{} with one paraphrase per question on the tabular records and two on 300 controlled test scenarios. Direct, CoT, VoI-P, and PAL are re-run with two instruction paraphrases on the first 300 controlled test scenarios per domain (Table~\ref{tab:para}).
The paraphrases change the average regret of every method little, and per LLM, the baselines vary more across prompts than \diva{}.
Under every prompt, every baseline has more than four times the regret of \diva{} under any prompt, indicating that prompt wording does not cause the gap.%
\input{tables/m_para}

\section{Qualitative Examples}
\label{app:examples}\label{app:qual}
Figure~\ref{fig:qual} shows the text of a controlled scenario as the LLM sees it, where the direct policy requests the most informative check, which cannot change the decision, whereas \diva{} finds that no check can change it and stops, which is DP-optimal.%
\input{tables/qual}

\section{Full Results}
\label{app:full}\label{app:full_know_llm}\label{app:full_ctrl_llm}
Per LLM and on the scenarios of the main tables (Conventions of Appendix~\ref{app:notes}), Tables~\ref{tab:full_know_llm_taiwan}--\ref{tab:full_know_llm_vol_20} report every knowledge-required environment, Table~\ref{tab:full_ctrl_llm} the controlled environments with native reasoning, Tables~\ref{tab:full_ctrl_llm_finance}--\ref{tab:full_ctrl_llm_legal} the four controlled domains, and Table~\ref{tab:full_episodes_llm} the controlled episodes. Never acquiring, which does not depend on the LLM, takes one row in every per-LLM table except Table~\ref{tab:full_episodes_llm}.%
\input{tables/full_know_llm}

The per-LLM tables show that the averages of Table~\ref{tab:know} are not driven by a few LLMs: \diva{}-U beats every LLM baseline for every LLM on Taiwan, German, and Heart, and its few losses lie mostly on the volatility targets. Without native reasoning in the best LLM baseline of Figure~\ref{fig:llms}, \diva{}'s average reduction rises from \RedBin\% to \RedBinNoReason\%.

\input{tables/full_ctrl_llm}
\input{tables/full_episodes_llm}

%% file: tables/notation.tex
\begin{table*}[t]
\centering
\small
\setlength{\tabcolsep}{4pt}
\begin{tabularx}{\textwidth}{@{}l >{\raggedright\arraybackslash}X@{}}
\toprule
Term & Meaning \\
\midrule
Episode & One decision case \\
State ($Y$) & The outcome of interest, unknown when deciding \\
Action, payoffs ($a$, $g$, $l$) & The two actions, the gain of acting if benign, and the loss of acting if adverse \\
Check, outcome, cost ($e$, $v$, $c_e$) & A piece of evidence that can be acquired, what it shows, and its price \\
Known findings ($\mathcal{K}$) & Outcomes already observed before deciding \\
Base rate ($\hat\pi$) & Probability of the adverse case before any finding \\
Belief ($\hat b$) & Probability of the adverse case given the known findings \\
Decision threshold ($b^\dagger$) & Belief above which \textsc{pass} is better than \textsc{buy}, $g/(g{+}l)$ \\
Reliability, likelihood ($\hat\ell_e(v|y)$) & Probability of each outcome of a check under each state \\
Strength & How strongly a check shifts the belief, the log-odds gap of its two likelihoods \\
Imagined belief ($\hat b_{e,v}$) & Belief if outcome $v$ of check $e$ were observed, by Bayes' rule \\
Decision value ($\mathrm{VOI}(e|E)$) & Expected improvement of the decision from observing a check \\
Net decision value ($\widehat\Delta(e)$) & Decision value minus cost, acquired only if positive \\
Information gain (EIG) & Expected reduction of uncertainty about $Y$, ignoring payoffs and costs \\
Regret & Value lost relative to the best possible decision (Appendix~\ref{app:setup}) \\
DP-optimal, myopic & Planning all future checks and stops exactly, or valuing only the next check \\
Shrinkage, uncertainty scale ($\lambda$, $\sigma$) & How much elicited strengths are weakened, and how widely \diva{}-U varies them \\
Reference model & Classifier fitted on training records that predicts $P(Y{=}1|\text{known findings})$ and gives the ground-truth decision values on real data \\
\bottomrule
\end{tabularx}
\caption{\textbf{Notation and terms.}}
\label{tab:notation}
\end{table*}

%% file: tables/spec_example.tex
\definecolor{specband}{RGB}{232,240,248}
\definecolor{specflip}{RGB}{226,240,228}
\definecolor{specred}{RGB}{170,0,30}
\definecolor{specblue}{RGB}{0,50,150}
\providecommand{\cR}[1]{\textcolor{specred}{#1}}
\providecommand{\cB}[1]{\textcolor{specblue}{#1}}
\definecolor{specgreen}{RGB}{0,110,40}
\providecommand{\cG}[1]{\textcolor{specgreen}{#1}}
\definecolor{specpurple}{RGB}{110,30,140}
\providecommand{\cP}[1]{\textcolor{specpurple}{#1}}
\definecolor{specnoflip}{RGB}{250,230,226}
\providecommand{\specq}[1]{\textcolor{llmgray}{\textit{#1}}}
\providecommand{\specstage}[1]{\rowcolor{specband}\multicolumn{3}{@{\hspace{4pt}}l}{\textbf{\textcolor{divablue}{#1}}}\\}
\begin{table*}[t]
\centering
\small
\renewcommand{\arraystretch}{1.1}
\setlength{\extrarowheight}{2pt}
\setlength{\aboverulesep}{0pt}
\setlength{\belowrulesep}{0pt}
\setlength{\tabcolsep}{5pt}
\begin{tabularx}{\textwidth}{@{\hspace{4pt}}>{\raggedright\arraybackslash}p{2.2cm}|>{\raggedright\arraybackslash}X|>{\raggedright\arraybackslash}X@{\hspace{4pt}}}
\toprule
& \textbf{(A) Controlled environments} & \textbf{(B) Knowledge-required environments} \\[-1pt]
& \textcolor{llmgray}{Numbers stated in the scenario} & \textcolor{llmgray}{No numbers in the scenario} \\
\midrule
\textbf{Common setup} & \multicolumn{2}{>{\raggedright\arraybackslash}p{\dimexpr\textwidth-2.2cm-22pt\relax}@{\hspace{4pt}}}{Payoffs: Gain $g=2$ and loss $l=3$, so the decision threshold is $b^\dagger=g/(g+l)=2/(2+3)=0.40$.\newline Known finding: Revenue $+20\%$ year over year, so the current belief is $\hat b=0.15$ (Figure~\ref{fig:overview}). The guidance costs $c_e=0.10$, and $V^*(b)=\max\{(1-b)g-bl,\,0\}$ is the value of the better action.}\rule[-1.2ex]{0pt}{0pt} \\
\cmidrule(lr){1-3}
\textbf{Scenario text} & \textit{``Underperforming stocks: 20\%. A guidance cut occurs in 51\% of underperforming stocks and 6.2\% of others.''} & \textit{``The company may cut or keep its forward guidance.''}\newline (No probabilities) \\
\midrule
\specstage{1) Elicitation: The LLM answers short numeric questions}
\textbf{Questions and answers} & The LLM \textit{copies} the numbers from the text. & The LLM \textit{answers from its own knowledge}, with the outcome-conditioned risk $\hat r_e(v)\approx P(Y{=}1|e{=}v)$.\rule[-1.6ex]{0pt}{0pt} \\
\cmidrule(lr){2-3}
& \specq{Base rate of underperformance?}\newline $\Rightarrow \hat\pi=0.20$ & \specq{Base rate of underperformance?}\newline $\Rightarrow \hat\pi=0.20$ \\
& \specq{Cut rate among underperforming stocks?}\newline $\Rightarrow \hat\ell(\text{cut}|1)=0.51$ & \specq{How often is guidance cut?}\newline $\Rightarrow \hat P(\text{cut})=0.15$ \\
& \specq{Cut rate among other stocks?}\newline $\Rightarrow \hat\ell(\text{cut}|0)=0.062$ & \specq{Share of underperformers among firms that cut?}\newline $\Rightarrow \hat r(\text{cut})\approx P(Y{=}1|\text{cut})=0.60$ \\
& & \specq{Share of underperformers among firms that keep?}\newline $\Rightarrow \hat r(\text{kept})\approx P(Y{=}1|\text{kept})=0.12$\rule[-1.6ex]{0pt}{0pt} \\
\cmidrule(lr){1-3}
\textbf{LLM calls} & $1+2$ per check & $1+3$ per check \\
\midrule
\specstage{2) Computation: No further LLM calls}
\textbf{Conversion to likelihoods} & Not needed. The kept outcome follows as $\hat\ell(\text{kept}|y)=1-\hat\ell(\text{cut}|y)$, i.e., $0.49$ and $0.938$. & With $p=\hat P(\text{cut})$, $r_{\text{c}}=\hat r(\text{cut})$, $r_{\text{k}}=\hat r(\text{kept})$, by Bayes' rule,\newline
\rule{0pt}{3.4ex}$\hat\ell(\text{cut}|1)=\dfrac{r_{\text{c}}\,p}{r_{\text{c}}\,p+r_{\text{k}}(1-p)}$\newline
\rule{0pt}{3.8ex}$\phantom{\hat\ell(\text{cut}|1)}=\dfrac{0.60\cdot0.15}{0.60\cdot0.15+0.12\cdot0.85}=0.47$,\newline
\rule{0pt}{3.8ex}$\hat\ell(\text{cut}|0)=\dfrac{(1-r_{\text{c}})\,p}{(1-r_{\text{c}})\,p+(1-r_{\text{k}})(1-p)}$\newline
\rule{0pt}{3.8ex}$\phantom{\hat\ell(\text{cut}|0)}=\dfrac{0.40\cdot0.15}{0.40\cdot0.15+0.88\cdot0.85}=0.07$.\rule[-3.2ex]{0pt}{0pt} \\
\cmidrule(lr){1-3}
\textbf{Shrinkage} & None ($\lambda=1$), as the copied numbers are exact. & $\lambda=0.75$ pulls both toward their mean $0.27$:\newline $0.27+0.75(0.47-0.27)=0.42$,\newline $0.27+0.75(0.07-0.27)=0.12$.\rule[-1.6ex]{0pt}{0pt} \\
\cmidrule(lr){1-3}
\textbf{Decision model} & $\cR{\hat\ell(\text{cut}|1)=0.51}$, $\cB{\hat\ell(\text{cut}|0)=0.062}$\newline strength $=\mathrm{logit}\,\hat\ell(\text{cut}|1)-\mathrm{logit}\,\hat\ell(\text{cut}|0)$\newline $\phantom{\text{strength}}=\mathrm{logit}\,\cR{0.51}-\mathrm{logit}\,\cB{0.062}$\newline $\phantom{\text{strength}}=\cR{0.04}-(\cB{-2.72})=\mathbf{2.76}$,\newline where $\mathrm{logit}\,p=\log\dfrac{p}{1-p}$ & $\cR{\hat\ell(\text{cut}|1)=0.42}$, $\cB{\hat\ell(\text{cut}|0)=0.12}$\newline strength $=\mathrm{logit}\,\hat\ell(\text{cut}|1)-\mathrm{logit}\,\hat\ell(\text{cut}|0)$\newline $\phantom{\text{strength}}=\mathrm{logit}\,\cR{0.42}-\mathrm{logit}\,\cB{0.12}$\newline $\phantom{\text{strength}}=\cR{-0.32}-(\cB{-1.99})=\mathbf{1.67}$\newline (Before shrinkage, $\mathrm{logit}\,0.47-\mathrm{logit}\,0.07=2.47$)\rule[-1.6ex]{0pt}{0pt} \\
\cmidrule(lr){1-3}
\textbf{\diva{}-U scale} & $\sigma=0$ & $\sigma=1$ \\
\midrule
\specstage{3) Valuation: Imagine each outcome, value the check, and decide}
\textbf{Imagined beliefs (Eq.~\ref{eq:imagine})} & \rule{0pt}{4.6ex}$\hat b_{\text{cut}}=\dfrac{0.15\cdot\cR{0.51}}{0.15\cdot\cR{0.51}+0.85\cdot\cB{0.062}}=\cG{\mathbf{0.59}}>b^\dagger$\rule[-2.2ex]{0pt}{0pt}\newline A cut flips \textsc{buy} to \textsc{pass} ($\hat b_{\text{kept}}=0.08$). & \rule{0pt}{4.6ex}$\hat b_{\text{cut}}=\dfrac{0.15\cdot\cR{0.42}}{0.15\cdot\cR{0.42}+0.85\cdot\cB{0.12}}=\cG{\mathbf{0.38}}<b^\dagger$\rule[-2.2ex]{0pt}{0pt}\newline No outcome flips \textsc{buy} ($\hat b_{\text{kept}}=0.10$).\rule[-1.6ex]{0pt}{0pt} \\
\cmidrule(lr){1-3}
\textbf{Net decision value (Eq.~\ref{eq:net})} & $\hat P(\text{cut}|E)=0.15\cdot\cR{0.51}+0.85\cdot\cB{0.062}=\cP{0.129}$\newline
$V^*(\hat b)=V^*(0.15)=1.250$\newline $V^*(\hat b_{\text{cut}})=V^*(\cG{0.59})=0$\newline $V^*(\hat b_{\text{kept}})=V^*(0.08)=1.578$\newline
\rule{0pt}{2.7ex}$\widehat{\mathrm{VOI}}=(\cP{0.129}\cdot0+\cP{0.871}\cdot1.578)-1.250=0.124$\newline
\rule{0pt}{2.7ex}$\widehat\Delta=\widehat{\mathrm{VOI}}-c_e=0.124-0.10=\mathbf{+0.024}$ &
$\hat P(\text{cut}|E)=0.15\cdot\cR{0.42}+0.85\cdot\cB{0.12}=\cP{0.165}$\newline
$V^*(\hat b)=V^*(0.15)=1.250$\newline $V^*(\hat b_{\text{cut}})=V^*(\cG{0.38})=0.091$\newline $V^*(\hat b_{\text{kept}})=V^*(0.10)=1.479$\newline
\rule{0pt}{2.7ex}$\widehat{\mathrm{VOI}}=(\cP{0.165}\cdot0.091+\cP{0.835}\cdot1.479)-1.250=0$\newline
\rule{0pt}{2.7ex}$\widehat\Delta=\widehat{\mathrm{VOI}}-c_e=0-0.10=\mathbf{-0.10}$\rule[-1.6ex]{0pt}{0pt} \\
\cmidrule(lr){1-3}
\textbf{Final decision} & \cellcolor{specflip}$\widehat\Delta>0$, so \diva{} \textbf{acquires the guidance}, as in Figure~\ref{fig:overview}. & \cellcolor{specnoflip}$\widehat\Delta<0$, so \diva{} \textbf{does not acquire it}. \diva{}-U (Section~\ref{sec:divau}) raises the VOI only to $0.03$, still below the cost.\rule[-1.6ex]{0pt}{0pt} \\
\bottomrule
\end{tabularx}
\caption{\textbf{Worked example of \diva{}.} The LLM answers only the \specq{gray} questions, and all else is computed exactly.}
\label{tab:spec_example}
\end{table*}

%% file: tables/models.tex
\begin{table*}[t]
\centering
\small
\begin{tabular*}{\textwidth}{@{\extracolsep{\fill}}ll|l|l|l@{}}
\toprule
Family & Size & Hugging Face checkpoint & Precision & License \\
\midrule
\multirow{6}{*}{\textbf{Qwen3}} & 0.6B & Qwen/Qwen3-0.6B & BF16 & Apache-2.0 \\
 & 1.7B & Qwen/Qwen3-1.7B & BF16 & Apache-2.0 \\
 & 4B & Qwen/Qwen3-4B & BF16 & Apache-2.0 \\
 & 8B & Qwen/Qwen3-8B & FP8 & Apache-2.0 \\
 & 14B & Qwen/Qwen3-14B & BF16 & Apache-2.0 \\
 & 32B & Qwen/Qwen3-32B-AWQ & AWQ 4-bit & Apache-2.0 \\
\cmidrule(lr){1-5}
\multirow{3}{*}{\textbf{Qwen2.5}} & 7B & Qwen/Qwen2.5-7B-Instruct & BF16 & Apache-2.0 \\
 & 14B & Qwen/Qwen2.5-14B-Instruct & BF16 & Apache-2.0 \\
 & 72B & Qwen/Qwen2.5-72B-Instruct-AWQ & AWQ 4-bit & Qwen License \\
\cmidrule(lr){1-5}
\multirow{2}{*}{\textbf{DeepSeek-R1}} & 7B & deepseek-ai/DeepSeek-R1-Distill-Qwen-7B & BF16 & MIT \\
 & 14B & deepseek-ai/DeepSeek-R1-Distill-Qwen-14B & BF16 & MIT \\
\cmidrule(lr){1-5}
\multirow{2}{*}{\textbf{Llama}} & 8B & hugging-quants/Meta-Llama-3.1-8B-Instruct-AWQ-INT4 & AWQ 4-bit & Llama 3.1 License \\
 & 70B & casperhansen/llama-3.3-70b-instruct-awq & AWQ 4-bit & Llama 3.3 License \\
\cmidrule(lr){1-5}
\multirow{3}{*}{\textbf{Gemma}} & 9B & google/gemma-2-9b-it & BF16 & Gemma Terms \\
 & 12B & google/gemma-3-12b-it & BF16 & Gemma Terms \\
 & 27B & google/gemma-2-27b-it & BF16 & Gemma Terms \\
\cmidrule(lr){1-5}
\multirow{2}{*}{\textbf{Mistral}} & 7B & mistralai/Mistral-7B-Instruct-v0.3 & BF16 & Apache-2.0 \\
 & 24B & casperhansen/mistral-small-24b-instruct-2501-awq & AWQ 4-bit & Apache-2.0 \\
\cmidrule(lr){1-5}
\textbf{Phi} & 14B & microsoft/phi-4 & BF16 & MIT \\
\cmidrule(lr){1-5}
\textbf{gpt-oss} & 20B & openai/gpt-oss-20b & MXFP4 & Apache-2.0 \\
\bottomrule
\end{tabular*}
\caption{\textbf{Model checkpoints} served with vLLM, with their precision and license. Later tables write Gemma for Gemma-2/3, Llama for Llama-3.1/3.3, Mistral 24B for Mistral-Small, and DeepSeek-R1 for R1-Distill-Qwen.}
\label{tab:models}
\end{table*}

%% file: tables/m_know_detail.tex
\begin{table*}[t]
\centering
\small
\setlength{\tabcolsep}{2pt}
\filltabular{\textwidth}{11}{>{\ftL}l|cccccccccc<{\ftR}}{
\toprule
Env. & Never & Always ask & \diva{}-$\kappa^\ddagger$ & \diva{}-$\beta^\ddagger$ & \diva{} & \diva{}-U & \begin{tabular}[c]{@{}c@{}}Stop\\opt.\end{tabular} & \begin{tabular}[c]{@{}c@{}}Stop\\\diva{}\end{tabular} & \begin{tabular}[c]{@{}c@{}}Stop\\\diva{}-U\end{tabular} & \begin{tabular}[c]{@{}c@{}}Stop\\VoI-Ask\end{tabular} \\
\midrule
Taiwan & 1.02 & 4.65 & 0.43 & \textbf{0.40} & 0.52 & \underline{0.41} & 0.65 & 0.69 & 0.56 & 0.93 \\
German & 0.91 & 4.93 & \textbf{0.49} & \underline{0.51} & 0.64 & \underline{0.51} & 0.69 & 0.75 & 0.57 & 0.98 \\
CDC & 0.36 & 4.96 & \underline{0.22} & \textbf{0.20} & 0.26 & \textbf{0.20} & 0.76 & 0.84 & 0.70 & 0.75 \\
Heart & 2.81 & 4.28 & 1.70 & \textbf{1.07} & 1.64 & \underline{1.08} & 0.61 & 0.85 & 0.72 & 0.86 \\
Ret-1d & 2.69 & 0.68 & 0.92 & \textbf{0.61} & 0.95 & \underline{0.64} & 0.34 & 0.32 & 0.01 & 0.65 \\
Ret-5d & 1.14 & 0.85 & \underline{0.74} & \textbf{0.64} & 0.80 & 0.77 & 0.52 & 0.39 & 0.02 & 0.71 \\
Vol-5d & 3.46 & \underline{0.81} & 2.24 & \underline{0.81} & 2.26 & \textbf{0.78} & 0.37 & 0.57 & 0.02 & 0.76 \\
Vol-20d & 3.91 & 0.77 & 2.49 & \underline{0.76} & 2.56 & \textbf{0.73} & 0.35 & 0.63 & 0.03 & 0.83 \\
\bottomrule
}
\caption{\textbf{Stopping behavior on knowledge-required environments.} DP regret and first-decision stop rates (All test scenarios, 20 LLMs), with \diva{}-$\beta^\ddagger$ and \diva{}-$\kappa^\ddagger$. VoI-Ask: First 300 test scenarios.}
\label{tab:know_detail}\label{tab:voiask_know}
\end{table*}

%% file: tables/m_episodes.tex
\begin{table*}[t]
\centering
\small
\setlength{\tabcolsep}{2pt}
\filltabular{\textwidth}{13}{>{\ftL}l|cccccccc|cccc<{\ftR}}{
\toprule
\multirow{3}{*}[-1.4ex]{\begin{tabular}[c]{@{}l@{}}Method\\(Avg.\ across 20 LLMs)\end{tabular}} & \multicolumn{8}{c|}{\textbf{Knowledge-required environments}} & \multicolumn{4}{c}{\textbf{Controlled environments}} \\
\cmidrule(lr){2-9}\cmidrule(lr){10-13}
 & \multicolumn{2}{c}{Credit} & \multicolumn{2}{c}{Clinical} & \multicolumn{4}{c|}{SEC filings (2025--2026)} & \multirow{2}{*}[-0.6ex]{Finance} & \multirow{2}{*}[-0.6ex]{Credit} & \multirow{2}{*}[-0.6ex]{Clinical} & \multirow{2}{*}[-0.6ex]{Legal} \\
\cmidrule(lr){2-3}\cmidrule(lr){4-5}\cmidrule(lr){6-9}
 & Taiwan & German & CDC & Heart & Ret-1d & Ret-5d & Vol-5d & \multicolumn{1}{c|}{Vol-20d} & & & & \\
\midrule
\multicolumn{13}{@{\hspace{3pt}}l}{\textit{Prompting}} \\
Direct & 31.24 & 31.23 & 30.69 & 34.40 & 6.73 & 5.47 & 6.00 & 6.10 & 28.60 & 23.92 & 30.04 & 31.29 \\
CoT \citep{wei2022chain} & 35.27 & 34.02 & 33.06 & 34.61 & 7.65 & 6.28 & 7.11 & 7.14 & 27.55 & 24.63 & 28.44 & 31.39 \\
SC \citep{wang2023selfconsistency} & 39.55 & 38.40 & 37.29 & 37.44 & 8.22 & 6.95 & 7.67 & 7.45 & 29.11 & 25.75 & 31.29 & 33.20 \\
ReAct \citep{yao2023react} & 37.29 & 34.89 & 33.82 & 32.35 & 5.97 & 4.86 & 5.20 & 5.51 & 22.19 & 21.41 & 23.65 & 28.67 \\
\midrule
\multicolumn{13}{@{\hspace{3pt}}l}{\textit{Decision-aware prompting}} \\
Ask-P \citep{zhong2026finriskatlas} & 29.12 & 31.29 & 31.14 & 33.07 & 6.95 & 5.74 & 6.18 & 6.36 & 28.33 & 24.56 & 29.78 & 30.68 \\
VoI-P & 13.39 & 15.69 & 14.46 & 18.61 & 4.24 & 3.13 & 4.40 & 4.39 & 16.99 & 14.10 & 17.28 & 21.00 \\
VoI-Ask \citep{dong2026value} & 0.73 & 1.35 & 1.37 & 3.66 & 2.48 & 1.35 & \underline{4.14} & \textbf{4.11} & 11.18 & 9.65 & 10.64 & 13.16 \\
\midrule
\multicolumn{13}{@{\hspace{3pt}}l}{\textit{Information gain}} \\
\begin{tabular}[c]{@{}l@{}}BED-LLM\\\citep{choudhury2026bedllm}\end{tabular} & 22.21 & 24.77 & 9.88 & 23.15 & 6.43 & 5.08 & 6.97 & 6.72 & 17.97 & 18.44 & 24.09 & 19.09 \\
\midrule
\multicolumn{13}{@{\hspace{3pt}}l}{\textit{Program-aided}} \\
PAL \citep{gao2023pal} & 6.58 & 6.00 & 5.21 & 8.23 & 3.20 & 2.18 & 4.16 & 4.25 & 13.95 & 11.65 & 13.37 & 15.70 \\
\midrule
\multicolumn{13}{@{\hspace{3pt}}l}{\textit{Reference}} \\
Never & 0.84 & 1.25 & 0.36 & 3.38 & 2.53 & 1.36 & \textbf{4.12} & \underline{4.17} & 10.99 & 9.41 & 10.44 & 12.88 \\
Always ask & 56.93 & 48.40 & 57.04 & 49.58 & 9.46 & 9.38 & 9.13 & 7.82 & 34.80 & 30.83 & 41.83 & 38.84 \\
Always ask $\rightarrow$ \diva{} & 5.08 & 5.33 & 5.36 & 5.84 & 1.78 & 1.53 & 4.65 & 4.46 & -- & -- & -- & -- \\
\diva{}-$\kappa^\ddagger$ & \underline{0.58} & 1.14 & \underline{0.30} & 2.80 & \underline{1.59} & \textbf{1.16} & 4.60 & 4.62 & -- & -- & -- & -- \\
\diva{}-$\beta^\ddagger$ & \textbf{0.50} & \textbf{0.79} & \textbf{0.24} & 3.39 & 2.04 & 1.37 & 6.52 & 4.79 & -- & -- & -- & -- \\
\midrule
\multicolumn{13}{@{\hspace{3pt}}l}{\textit{Ours}} \\
\rowcolor{oursrow} \diva{} & 0.69 & 1.26 & 0.35 & 2.66 & 1.65 & 1.24 & 4.47 & 4.39 &  &  &  &  \\
\rowcolor{oursrow} \diva{}-U & 0.68 & 0.96 & \underline{0.30} & \underline{2.07} & 4.40 & 4.43 & 4.97 & 4.39 & \multirow{-2}{*}{\underline{1.92}} & \multirow{-2}{*}{\underline{2.05}} & \multirow{-2}{*}{\underline{1.48}} & \multirow{-2}{*}{\underline{1.74}} \\
\rowcolor{oursrow} \diva{}-DP & \underline{0.58} & \underline{0.89} & \underline{0.30} & \textbf{2.01} & \textbf{1.58} & \underline{1.22} & 4.19 & \textbf{4.11} & \textbf{1.65} & \textbf{1.78} & \textbf{1.42} & \textbf{1.48} \\
\midrule
Wins of \diva{}-U & 10/20 & 17/20 & 18/20 & 20/20 & 1/20 & 0/20 & 0/20 & 3/20 & 20/20 & 20/20 & 20/20 & 20/20 \\
\bottomrule
}
\caption{\textbf{Episode regret on test scenarios 0--149} (First 100 per controlled domain), averaged over 20 LLMs. Final action: From the reference model's belief (Knowledge-required) or the policy's own belief (Controlled). --: Not run.}
\label{tab:episodes_full}\label{tab:episodes_all}
\end{table*}

%% file: tables/m_controls.tex
\begin{table*}[t]
\centering
\small
\setlength{\tabcolsep}{3pt}\setlength{\aboverulesep}{0pt}\setlength{\belowrulesep}{0pt}\renewcommand{\arraystretch}{1.2}
\filltabular{\textwidth}{9}{>{\ftL}l|cccc|cccc<{\ftR}}{
\toprule
\multirow{2}{*}[-0.5ex]{Method} & \multicolumn{4}{c|}{Records} & \multicolumn{4}{c}{SEC filings} \\
\cmidrule(lr){2-5}\cmidrule(lr){6-9}
 & Taiwan & German & CDC & Heart & Ret-1d & Ret-5d & Vol-5d & Vol-20d \\
\midrule
\rowcolor{gray!12}\multicolumn{9}{@{\hspace{3pt}}l}{\textit{Reference}} \\
Never & 0.89 & 1.06 & 0.31 & 2.96 & 2.64 & 1.76 & 3.71 & 4.38 \\
\rowcolor{gray!12}\multicolumn{9}{@{\hspace{3pt}}l}{\textit{Generic decision model (One strength for every check)}} \\
Generic \diva{} & 0.65 & \underline{0.51} & \underline{0.19} & 1.52 & 1.12 & 0.82 & 1.39 & 1.57 \\
Generic \diva{}-U & \underline{0.44} & \textbf{0.45} & \textbf{0.12} & \textbf{1.06} & \textbf{0.61} & \textbf{0.65} & \textbf{0.65} & \textbf{0.60} \\
\rowcolor{gray!12}\multicolumn{9}{@{\hspace{3pt}}l}{\textit{Elicited decision model (Ours)}} \\
\rowcolor{oursrow} \diva{} & 0.48 & 0.65 & 0.26 & 1.71 & 1.01 & 0.93 & 2.45 & 2.80 \\
\rowcolor{oursrow} \diva{}-U & \textbf{0.41} & 0.52 & 0.22 & \underline{1.12} & \underline{0.67} & \underline{0.71} & \underline{0.83} & \underline{0.74} \\
\bottomrule
}
\caption{\textbf{Generic decision model.} DP regret of \diva{} and \diva{}-U with one strength for every check, selected on development data, vs.\ the elicited decision model, first 300 test scenarios, averaged over 20 LLMs.}
\label{tab:const}
\end{table*}
\begin{table}[t]
\centering
\small
\setlength{\tabcolsep}{3pt}\setlength{\aboverulesep}{0pt}\setlength{\belowrulesep}{0pt}\renewcommand{\arraystretch}{1.2}
\filltabular{\columnwidth}{5}{>{\ftL}l|cccc<{\ftR}}{
\toprule
Method & Taiwan & German & CDC & Heart \\
\midrule
\rowcolor{gray!12}\multicolumn{5}{@{\hspace{3pt}}l}{\textit{Reference}} \\
Never & 1.02 & 0.91 & 0.36 & 2.81 \\
Threshold heur. & 1.02 & 0.91 & 0.38 & 3.07 \\
Supervised ref. & \textbf{0.05} & \textbf{0.11} & \textbf{0.04} & \textbf{0.29} \\
\rowcolor{gray!12}\multicolumn{5}{@{\hspace{3pt}}l}{\textit{Shuffled likelihoods}} \\
Knowledge-free \diva{} & 0.64 & 0.61 & 0.28 & 1.71 \\
Knowledge-free \diva{}-U & 0.52 & \underline{0.49} & 0.23 & 1.15 \\
\rowcolor{gray!12}\multicolumn{5}{@{\hspace{3pt}}l}{\textit{Ours}} \\
\rowcolor{oursrow} \diva{} & 0.52 & 0.64 & 0.26 & 1.64 \\
\rowcolor{oursrow} \diva{}-U & \underline{0.41} & 0.51 & \underline{0.20} & \underline{1.08} \\
\bottomrule
}
\caption{\textbf{Controls on real records.} DP regret on all test scenarios, averaged over 20 LLMs. Knowledge-free: Elicited likelihoods shuffled across checks.}
\label{tab:controls}
\end{table}

%% file: tables/abl_eig.tex
\begin{table*}[t]
\centering
\small
\setlength{\tabcolsep}{4.5pt}
\filltabular{\textwidth}{10}{>{\ftL}l|cccccccc|c<{\ftR}}{
\toprule
\multirow{2}{*}[-0.6ex]{\begin{tabular}[c]{@{}l@{}}Variant\\(Avg.\ across 20 LLMs)\end{tabular}} & \multicolumn{2}{c}{Credit} & \multicolumn{2}{c}{Clinical} & \multicolumn{4}{c|}{SEC filings (2025--2026)} & \multirow{2}{*}[-0.6ex]{Avg.} \\
\cmidrule(lr){2-3}\cmidrule(lr){4-5}\cmidrule(lr){6-9}
 & Taiwan & German & CDC & Heart & Ret-1d & Ret-5d & Vol-5d & Vol-20d & \\
\midrule
\diva{} w/ EIG (Fixed threshold) & 4.88 & 4.01 & 2.04 & 6.07 & 2.20 & 1.09 & 3.38 & 3.81 & 3.43 \\
\diva{} w/ EIG$^\ddagger$ (Tuned on labels) & \textbf{0.38} & \textbf{0.23} & \textbf{0.12} & \underline{1.43} & \underline{0.73} & \textbf{0.46} & \underline{1.43} & \underline{1.49} & \underline{0.78} \\
\midrule
\rowcolor{oursrow} \diva{} & 0.52 & 0.64 & 0.26 & 1.64 & 0.95 & 0.80 & 2.26 & 2.56 & 1.20 \\
\rowcolor{oursrow} \diva{}-U & \underline{0.41} & \underline{0.51} & \underline{0.20} & \textbf{1.08} & \textbf{0.64} & \underline{0.77} & \textbf{0.78} & \textbf{0.73} & \textbf{0.64} \\
\bottomrule
}
\caption{\textbf{Decision value vs.\ EIG on the same decision model.} DP regret on the full test sets, averaged over 20 LLMs. $^\ddagger$: Cost trade-off and threshold tuned on development scenarios with ground-truth decision values.}
\label{tab:abl_eig}
\end{table*}

%% file: tables/transfer.tex
\begin{table}[t]
\centering
\small
\setlength{\tabcolsep}{3pt}
\begin{tabular*}{\linewidth}{@{\extracolsep{\fill}}l|cccc}
\toprule
\begin{tabular}[c]{@{}c@{}}Held-out\\\relax source\end{tabular} & \begin{tabular}[c]{@{}c@{}}Selected\\\relax ($\lambda$, $\sigma$)\end{tabular} & \begin{tabular}[c]{@{}c@{}}\diva{}\\\relax (Point)\end{tabular} & \begin{tabular}[c]{@{}c@{}}\diva{}-U\\\relax ($\lambda{=}0.75$)\end{tabular} & \begin{tabular}[c]{@{}c@{}}\diva{}-U\\\relax (Selected)\end{tabular} \\
\midrule
Taiwan & (0.5, 1.0) & 0.53 & 0.44 & 0.32 \\
German & (0.5, 1.0) & 0.64 & 0.52 & 0.34 \\
CDC & (0.5, 1.0) & 0.81 & 0.70 & 0.40 \\
Heart & (0.5, 1.0) & 0.56 & 0.36 & 0.36 \\
SEC filings & (0.5, 1.0) & 0.53 & 0.26 & 0.24 \\
\bottomrule
\end{tabular*}
\caption{\textbf{Leave-one-source-out transfer.} Regret relative to Never on each held-out data source.}
\label{tab:transfer}
\end{table}

%% file: tables/ms.tex
\begin{table*}[t]
\centering
\small
\setlength{\tabcolsep}{3pt}
\begin{tabular}{@{}ll|ccccccccc|cc@{}}
\toprule
\multirow{2}{*}[-2ex]{Family} & \multirow{2}{*}[-2ex]{Size} & \multicolumn{1}{c}{Prompting} & \multicolumn{2}{c}{Reference} & \multicolumn{5}{c}{Ablation} & \multicolumn{1}{c|}{\cellcolor{oursrow}Ours} & \multicolumn{2}{c}{Verbal beliefs} \\
\cmidrule(lr){3-3}\cmidrule(lr){4-5}\cmidrule(lr){6-10}\cmidrule(lr){11-11}\cmidrule(lr){12-13}
 & & Direct & Never & \begin{tabular}[c]{@{}c@{}}Always\\ask\end{tabular} & \begin{tabular}[c]{@{}c@{}}\diva{} w/\\EIG\end{tabular} & \begin{tabular}[c]{@{}c@{}}\diva{} w/\\EIG$^\ddagger$\end{tabular} & \begin{tabular}[c]{@{}c@{}}\diva{} w/\\ES\end{tabular} & \begin{tabular}[c]{@{}c@{}}\diva{} w/\\ES$^\ddagger$\end{tabular} & \begin{tabular}[c]{@{}c@{}}\diva{}\\(Elic.)\end{tabular} & \cellcolor{oursrow}\diva{} & Direct & \cellcolor{oursrow}\diva{} \\
\midrule
\multirow{6}{*}{\textbf{Qwen3}} & 0.6B & 5.85 & 5.83 & \textbf{1.53} & 3.03 & 2.89 & 3.07 & 2.93 & 5.30 & \cellcolor{oursrow}\underline{2.83} & {6.05} & \cellcolor{oursrow}\textbf{3.63} \\
 & 1.7B & 5.77 & 5.83 & \textbf{1.37} & 3.00 & 2.89 & 2.59 & 2.69 & 4.74 & \cellcolor{oursrow}\underline{1.39} & {5.97} & \cellcolor{oursrow}\textbf{2.35} \\
 & 4B & 3.36 & 5.83 & \underline{1.14} & 2.89 & 2.28 & 2.04 & 1.86 & 3.27 & \cellcolor{oursrow}\textbf{0.22} & {3.76} & \cellcolor{oursrow}\textbf{1.11} \\
 & 8B & 3.10 & 5.83 & \underline{1.11} & 2.88 & 2.35 & 2.03 & 1.85 & 3.68 & \cellcolor{oursrow}\textbf{0.14} & {3.08} & \cellcolor{oursrow}\textbf{1.28} \\
 & 14B & 2.86 & 5.83 & \underline{1.10} & 2.88 & 2.37 & 2.02 & 1.85 & 4.43 & \cellcolor{oursrow}\textbf{0.13} & {2.58} & \cellcolor{oursrow}\textbf{0.90} \\
 & 32B & 3.40 & 5.83 & \underline{1.11} & 2.89 & 2.37 & 2.05 & 1.88 & 2.71 & \cellcolor{oursrow}\textbf{0.15} & {3.26} & \cellcolor{oursrow}\textbf{0.79} \\
\cmidrule(lr){1-13}
\multirow{3}{*}{\textbf{Qwen2.5}} & 7B & 3.35 & 5.83 & \underline{1.12} & 2.89 & 2.33 & 2.04 & 1.90 & 4.72 & \cellcolor{oursrow}\textbf{0.22} & {3.22} & \cellcolor{oursrow}\textbf{1.23} \\
 & 14B & 2.86 & 5.83 & \underline{1.10} & 2.86 & 2.35 & 2.01 & 1.84 & 2.67 & \cellcolor{oursrow}\textbf{0.14} & {2.78} & \cellcolor{oursrow}\textbf{0.87} \\
 & 72B & 2.57 & 5.83 & \underline{1.10} & 2.86 & 2.35 & 2.02 & 1.85 & 3.95 & \cellcolor{oursrow}\textbf{0.13} & {2.38} & \cellcolor{oursrow}\textbf{0.46} \\
\cmidrule(lr){1-13}
\multirow{2}{*}{\textbf{DeepSeek-R1}} & 7B & 4.35 & 5.83 & \underline{1.21} & 2.89 & 2.49 & 2.22 & 2.16 & 4.40 & \cellcolor{oursrow}\textbf{0.59} & {4.45} & \cellcolor{oursrow}\textbf{1.96} \\
 & 14B & 2.82 & 5.83 & \underline{1.12} & 2.90 & 2.39 & 2.04 & 1.90 & 3.59 & \cellcolor{oursrow}\textbf{0.35} & {2.83} & \cellcolor{oursrow}\textbf{1.10} \\
\cmidrule(lr){1-13}
\multirow{2}{*}{\textbf{Llama}} & 8B & 5.01 & 5.83 & \underline{1.11} & 2.89 & 2.37 & 2.05 & 1.86 & 3.14 & \cellcolor{oursrow}\textbf{0.16} & {5.09} & \cellcolor{oursrow}\textbf{0.89} \\
 & 70B & 2.44 & 5.83 & \underline{1.10} & 2.86 & 2.35 & 2.02 & 1.85 & 2.40 & \cellcolor{oursrow}\textbf{0.11} & {2.40} & \cellcolor{oursrow}\textbf{0.97} \\
\cmidrule(lr){1-13}
\multirow{3}{*}{\textbf{Gemma}} & 9B & 2.73 & 5.83 & \underline{1.11} & 2.89 & 2.36 & 2.03 & 1.87 & 4.95 & \cellcolor{oursrow}\textbf{0.17} & {2.59} & \cellcolor{oursrow}\textbf{1.11} \\
 & 12B & 2.88 & 5.83 & \underline{1.10} & 2.85 & 2.35 & 2.03 & 1.88 & 5.11 & \cellcolor{oursrow}\textbf{0.15} & {2.79} & \cellcolor{oursrow}\textbf{0.62} \\
 & 27B & 2.91 & 5.83 & \underline{1.11} & 2.89 & 2.39 & 2.02 & 1.87 & 4.28 & \cellcolor{oursrow}\textbf{0.23} & {2.74} & \cellcolor{oursrow}\textbf{0.71} \\
\cmidrule(lr){1-13}
\multirow{2}{*}{\textbf{Mistral}} & 7B & 2.74 & 5.83 & \underline{1.16} & 2.88 & 2.40 & 2.12 & 1.97 & 4.16 & \cellcolor{oursrow}\textbf{0.32} & {2.75} & \cellcolor{oursrow}\textbf{0.94} \\
 & 24B & 2.74 & 5.83 & \underline{1.10} & 2.86 & 2.35 & 2.02 & 1.85 & 1.21 & \cellcolor{oursrow}\textbf{0.12} & {2.82} & \cellcolor{oursrow}\textbf{0.53} \\
\cmidrule(lr){1-13}
\multirow{1}{*}{\textbf{Phi}} & 14B & 2.61 & 5.83 & \underline{1.10} & 2.86 & 2.36 & 2.02 & 1.86 & 3.71 & \cellcolor{oursrow}\textbf{0.16} & {2.60} & \cellcolor{oursrow}\textbf{1.01} \\
\cmidrule(lr){1-13}
\multirow{1}{*}{\textbf{gpt-oss}} & 20B & 5.35 & 5.83 & \underline{1.11} & 2.87 & 2.35 & 2.02 & 1.86 & 5.15 & \cellcolor{oursrow}\textbf{0.21} & {5.52} & \cellcolor{oursrow}\textbf{0.87} \\
\bottomrule
\end{tabular}
\caption{\textbf{Multi-state environments.} Regret on held-out scenarios (Scenarios 500--999 per domain), normalized by half the payoff range. Verbal beliefs (Scenarios 500--799): Best and second best marked within the group.}
\label{tab:ms}
\end{table*}

%% file: tables/m_realized.tex
\begin{table*}[t]
\centering
\small
\setlength{\tabcolsep}{3pt}
\filltabular{\textwidth}{9}{>{\ftL}l|ccccc|ccc<{\ftR}}{
\toprule
\multirow{2}{*}[-0.6ex]{Records} & \multicolumn{5}{c|}{LLM policies ($\times$100, normalized by $l$)} & \multicolumn{3}{c}{Reference DP policy (Payoff points)} \\
\cmidrule(lr){2-6}\cmidrule(lr){7-9}
 & Never & \cellcolor{oursrow}\diva{} & \cellcolor{oursrow}B / W & \cellcolor{oursrow}\diva{}-U & \cellcolor{oursrow}B / W & Never & Ref. DP & Difference ($\pm$ SE) \\
\midrule
Taiwan & \textbf{24.39} & \cellcolor{oursrow}\underline{23.91} & \cellcolor{oursrow}0 / 4 & \cellcolor{oursrow}23.78 & \cellcolor{oursrow}0 / 4 & 0.710 & 0.735 & 0.025 $\pm$ 0.021 \\
German & \textbf{16.56} & \cellcolor{oursrow}\underline{16.08} & \cellcolor{oursrow}0 / 2 & \cellcolor{oursrow}15.44 & \cellcolor{oursrow}0 / 5 & 0.394 & 0.428 & 0.034 $\pm$ 0.031 \\
CDC & 35.48 & \cellcolor{oursrow}\underline{35.55} & \cellcolor{oursrow}2 / 1 & \cellcolor{oursrow}\textbf{35.61} & \cellcolor{oursrow}0 / 1 & 1.112 & 1.141 & 0.029 $\pm$ 0.015 \\
Heart & \textbf{1.35} & \cellcolor{oursrow}\underline{0.64} & \cellcolor{oursrow}1 / 6 & \cellcolor{oursrow}0.34 & \cellcolor{oursrow}0 / 5 & 0.124 & 0.222 & 0.098 $\pm$ 0.031 \\
\midrule
Taiwan (5,000) & \textbf{25.18} & \cellcolor{oursrow}\underline{24.80} & \cellcolor{oursrow}0 / 6 & \cellcolor{oursrow}24.56 & \cellcolor{oursrow}0 / 8 & 25.94$^\dag$ & 27.04$^\dag$ & 1.10 $\pm$ 0.25$^\dag$ \\
CDC (5,000) & 36.20 & \cellcolor{oursrow}\textbf{36.27} & \cellcolor{oursrow}3 / 2 & \cellcolor{oursrow}\underline{36.25} & \cellcolor{oursrow}1 / 2 & 36.77$^\dag$ & 37.19$^\dag$ & 0.42 $\pm$ 0.16$^\dag$ \\
\bottomrule
}
\caption{\textbf{Realized utility on real records}, net of costs (Higher is better), averaged over 20 LLMs. B / W: LLMs significantly better / worse than never acquiring. Ref. DP: The DP policy of the reference model. Its Never row decides at once with the reference model's belief. $^\dag$: Multiplied by 100 and normalized by $l$.}
\label{tab:realized}\label{tab:realized_1}\label{tab:realized_ref}\label{tab:realized_big}
\end{table*}

%% file: tables/main_know_myo.tex
\begin{table*}[t]
\centering
\small
\setlength{\tabcolsep}{4.5pt}
\begin{tabular}{l|cccccccc|c}
\toprule
\multirow{2}{*}[-0.6ex]{\begin{tabular}[c]{@{}l@{}}Method\\(Avg.\ across 20 LLMs)\end{tabular}} & \multicolumn{2}{c}{Credit} & \multicolumn{2}{c}{Clinical} & \multicolumn{4}{c|}{SEC filings (2025--2026)} & \multirow{2}{*}[-0.6ex]{Avg.} \\
\cmidrule(lr){2-3}\cmidrule(lr){4-5}\cmidrule(lr){6-9}
 & Taiwan & German & CDC & Heart & Ret-1d & Ret-5d & Vol-5d & Vol-20d & \\
\midrule
\multicolumn{10}{l}{\textit{Prompting}} \\
Direct & 6.68 & 5.43 & 5.26 & 7.25 & 2.12 & 1.79 & \textbf{1.84} & \textbf{1.73} & 4.01 \\
CoT \citep{wei2022chain} & 7.72 & 6.98 & 6.62 & 7.98 & 1.94 & 1.76 & 2.08 & 2.12 & 4.65 \\
SC \citep{wang2023selfconsistency} & 7.73 & 7.15 & 6.89 & 7.91 & 1.79 & 1.72 & \underline{2.01} & \underline{1.85} & 4.63 \\
ReAct \citep{yao2023react} & 9.13 & 7.76 & 8.12 & 8.48 & 2.48 & 2.22 & 2.28 & 2.45 & 5.37 \\
\midrule
\multicolumn{10}{l}{\textit{Decision-aware prompting}} \\
Ask-P \citep{zhong2026finriskatlas} & 7.10 & 7.01 & 6.27 & 8.71 & 1.95 & 1.73 & 2.43 & 2.33 & 4.69 \\
VoI-P & 6.15 & 5.75 & 5.45 & 6.84 & 2.37 & 1.97 & 2.69 & 2.96 & 4.27 \\
VoI-Ask \citep{dong2026value} & 0.86 & \underline{0.80} & 1.27 & 2.68 & 2.18 & 1.55 & 3.55 & 4.24 & 2.14 \\
\midrule
\multicolumn{10}{l}{\textit{Information gain}} \\
BED-LLM \citep{choudhury2026bedllm} & 6.47 & 6.15 & 3.43 & 8.43 & 2.64 & 2.24 & 4.57 & 5.20 & 4.89 \\
\midrule
\multicolumn{10}{l}{\textit{Program-aided}} \\
PAL \citep{gao2023pal} & 3.70 & 3.06 & 2.51 & 4.42 & 2.55 & 1.95 & 3.11 & 3.66 & 3.12 \\
\midrule
\multicolumn{10}{l}{\textit{Reference}} \\
Never & 0.80 & \textbf{0.72} & \textbf{0.21} & 2.40 & 2.58 & 1.68 & 3.48 & 4.27 & 2.02 \\
Always ask & 5.48 & 5.02 & 4.87 & 5.26 & 1.71 & 1.55 & 3.84 & 4.42 & 4.02 \\
\midrule
\multicolumn{10}{l}{\textit{Ours}} \\
\rowcolor{oursrow} \diva{} & \textbf{0.65} & 0.94 & \underline{0.26} & \underline{1.98} & \textbf{1.60} & \textbf{1.31} & 3.71 & 4.43 & \textbf{1.86} \\
\rowcolor{oursrow} \diva{}-U & \underline{0.69} & 1.04 & 0.28 & \textbf{1.96} & \underline{1.64} & 1.46 & 3.69 & 4.19 & \underline{1.87} \\
\rowcolor{oursrow} \diva{}-DP & 0.77 & 0.90 & 0.29 & 1.99 & 1.83 & \underline{1.44} & 3.61 & 4.33 & 1.90 \\
\bottomrule
\end{tabular}
\caption{\textbf{Myopic regret on knowledge-required environments.} Regret of the first decision against the myopic optimum, on the same scenarios and LLMs as Table~\ref{tab:know}. Avg.: Mean over the eight environments.}
\label{tab:know_myo}
\end{table*}

%% file: tables/m_reference.tex
\begin{table*}[t]
\small
\setlength{\tabcolsep}{3pt}
\begin{minipage}[t]{0.53\textwidth}\vspace{0pt}\centering
\filltabular{\linewidth}{6}{>{\ftL}l|ccc|cc<{\ftR}}{
\toprule
\multirow{2}{*}[-0.6ex]{Method} & \multicolumn{3}{c|}{Records} & \multicolumn{2}{c}{SEC filings} \\
\cmidrule(lr){2-4}\cmidrule(lr){5-6}
& Main & Alt. & NB & Main & Alt. \\
\midrule
\multicolumn{6}{@{\hspace{3pt}}l}{\textit{Prompting}} \\
Direct & 5.79 & 5.72 & 5.76 & 1.20 & 1.17 \\
CoT \citep{wei2022chain} & 6.86 & 6.85 & 6.76 & 1.21 & 1.21 \\
SC \citep{wang2023selfconsistency} & 6.96 & 6.94 & 6.83 & 1.18 & 1.19 \\
ReAct \citep{yao2023react} & 7.79 & 7.80 & 7.63 & 1.12 & 1.14 \\
\midrule
\multicolumn{6}{@{\hspace{3pt}}l}{\textit{Decision-aware prompting}} \\
Ask-P \citep{zhong2026finriskatlas} & 6.90 & 6.85 & 6.89 & 1.62 & 1.58 \\
VoI-P & 5.77 & 5.70 & 5.76 & 1.86 & 1.78 \\
VoI-Ask \citep{dong2026value} & 1.49 & 1.31 & 1.74 & 2.48 & 2.31 \\
\midrule
\multicolumn{6}{@{\hspace{3pt}}l}{\textit{Information gain}} \\
\begin{tabular}[c]{@{}l@{}}BED-LLM\\\citep{choudhury2026bedllm}\end{tabular} & 5.68 & 5.62 & 5.72 & 1.57 & 1.56 \\
\midrule
\multicolumn{6}{@{\hspace{3pt}}l}{\textit{Program-aided}} \\
PAL \citep{gao2023pal} & 3.36 & 3.24 & 3.53 & 2.19 & 2.06 \\
\midrule
\multicolumn{6}{@{\hspace{3pt}}l}{\textit{Reference}} \\
Never & 1.31 & 1.10 & 1.67 & 3.12 & 2.87 \\
Always ask & 4.63 & 4.64 & 4.47 & \underline{0.79} & \underline{0.81} \\
\midrule
\multicolumn{6}{@{\hspace{3pt}}l}{\textit{Ours}} \\
\rowcolor{oursrow} \diva{} & 0.77 & 0.67 & 0.87 & 1.80 & 1.67 \\
\rowcolor{oursrow} \diva{}-U & \textbf{0.57} & \textbf{0.50} & \textbf{0.53} & \textbf{0.74} & \textbf{0.75} \\
\rowcolor{oursrow} \diva{}-DP & \underline{0.64} & \underline{0.55} & \underline{0.65} & 1.50 & 1.40 \\
\midrule
Rank correlation with Main & & 0.997 & 0.996 & & 0.995 \\
\bottomrule}
\captionof{table}{\textbf{Robustness to the reference model.} DP regret under the main (Main), an alternative (Alt.), and a naive Bayes (NB) reference model, averaged over 20 LLMs.}
\label{tab:altgold}\label{tab:alt_ref}
\end{minipage}\hfill
\begin{minipage}[t]{0.44\textwidth}\vspace{0pt}\centering
\filltabular{\linewidth}{5}{>{\ftL}l|cccc<{\ftR}}{
\toprule
Env. & AUC & Brier & Base & ECE \\
\midrule
Taiwan & 0.746 & 0.139 & 0.172 & 0.016 \\
German & 0.778 & 0.168 & 0.210 & 0.054 \\
CDC & 0.808 & 0.100 & 0.120 & 0.005 \\
Heart & 0.900 & 0.127 & 0.247 & 0.062 \\
\midrule
Ret-1d & 0.601 & 0.242 & 0.250 & 0.014 \\
Ret-5d & 0.595 & 0.243 & 0.249 & 0.009 \\
Vol-5d & 0.658 & 0.236 & 0.250 & 0.068 \\
Vol-20d & 0.658 & 0.245 & 0.250 & 0.118 \\
\bottomrule
}
\captionof{table}{\textbf{Validity of the reference models} on held-out records. ECE: Expected calibration error. Base: Brier score of a constant base-rate prediction.}
\label{tab:ref_metrics}
\vspace{6pt}
\filltabular{\linewidth}{5}{>{\ftL}l|cccc<{\ftR}}{
\toprule
Method & Taiwan & German & CDC & Heart \\
\midrule
Direct & 5.49 & 5.57 & 5.19 & 6.19 \\
CoT & 6.16 & 6.92 & 6.67 & 6.57 \\
SC & 6.27 & 7.07 & 6.88 & 6.39 \\
ReAct & 7.24 & 7.46 & 8.01 & 6.87 \\
\midrule
Ask-P & 6.00 & 6.98 & 6.27 & 7.28 \\
VoI-P & 5.03 & 5.62 & 5.35 & 5.93 \\
VoI-Ask & 0.83 & 0.94 & 1.15 & 2.56 \\
\midrule
BED-LLM & 5.39 & 6.14 & 3.36 & 6.62 \\
\midrule
PAL & 2.95 & 2.98 & 2.50 & 4.16 \\
\midrule
Never & 0.90 & 0.87 & 0.31 & 2.85 \\
Always ask & 4.32 & 4.77 & 4.69 & 3.95 \\
\midrule
\rowcolor{oursrow} \diva{} & \underline{0.54} & \underline{0.72} & \underline{0.21} & \underline{1.67} \\
\rowcolor{oursrow} \diva{}-U & \textbf{0.45} & \textbf{0.62} & \textbf{0.17} & \textbf{1.14} \\
\bottomrule
}
\captionof{table}{\textbf{Fresh tabular scenarios} not used in any design decision: DP regret over 20 LLMs.}
\label{tab:fresh}
\end{minipage}
\end{table*}

%% file: tables/m1_decomp.tex
\begin{table*}[t]
\small
\begin{minipage}[c]{\columnwidth}
\centering
\setlength{\tabcolsep}{3pt}
\filltabular{\columnwidth}{7}{>{\ftL}l|cccccc<{\ftR}}{
\toprule
\multirow{2}{*}[-0.6ex]{Selection} & \multicolumn{3}{c}{\textbf{Knowledge-required}} & \multicolumn{3}{c}{\textbf{Controlled}} \\
\cmidrule(lr){2-4}\cmidrule(lr){5-7}
 & LLM & \diva{}-U & DP & LLM & \diva{}-U & DP \\
\midrule
LLM & 3.49 & 0.91 & 0.38 & 5.26 & 1.50 & 1.04 \\
EIG & 3.97 & 1.08 & 0.53 & 4.90 & 1.14 & 0.67 \\
\rowcolor{oursrow} \diva{}-U & 2.23 & 0.65 & 0.25 & 2.90 & 0.86 & 0.28 \\
DP optimum & 1.92 & 0.38 & 0.00 & 2.55 & 0.64 & 0.00 \\
\bottomrule
}
\end{minipage}\hfill
\begin{minipage}[c]{\columnwidth}
\caption{\textbf{Which check vs.\ when to stop.} DP regret when stopping by the column's rule, else acquiring the row's top check. LLM: Direct policy. DP: DP optimum. \diva{}-U equals \diva{} on the controlled environments.}
\label{tab:m1_decomp}
\end{minipage}
\end{table*}

%% file: tables/m_extra.tex
\begin{table*}[t]
\centering
\small
\setlength{\tabcolsep}{2pt}
\filltabular{\textwidth}{13}{>{\ftL}l|cccccccc|cccc<{\ftR}}{
\toprule
\multirow{3}{*}[-1.4ex]{\begin{tabular}[c]{@{}l@{}}Method\\(Avg.\ across 20 LLMs)\end{tabular}} & \multicolumn{8}{c|}{\textbf{Knowledge-required environments}} & \multicolumn{4}{c}{\textbf{Controlled environments}} \\
\cmidrule(lr){2-9}\cmidrule(lr){10-13}
 & \multicolumn{2}{c}{Credit} & \multicolumn{2}{c}{Clinical} & \multicolumn{4}{c|}{SEC filings (2025--2026)} & \multirow{2}{*}[-0.6ex]{Finance} & \multirow{2}{*}[-0.6ex]{Credit} & \multirow{2}{*}[-0.6ex]{Clinical} & \multirow{2}{*}[-0.6ex]{Legal} \\
\cmidrule(lr){2-3}\cmidrule(lr){4-5}\cmidrule(lr){6-9}
 & Taiwan & German & CDC & Heart & Ret-1d & Ret-5d & Vol-5d & \multicolumn{1}{c|}{Vol-20d} & & & & \\
\midrule
\multicolumn{13}{@{\hspace{3pt}}l}{\textit{Prompting}} \\
Direct & 6.47 & 5.15 & 5.23 & 6.30 & 1.26 & 1.21 & \underline{1.24} & \underline{1.08} & 4.69 & 4.35 & 6.22 & 5.80 \\
Direct + tuned stop & 3.43 & 2.68 & 2.50 & 4.30 & 1.81 & 1.35 & 2.33 & 2.59 & 4.39 & 4.39 & 5.49 & 5.57 \\
Salience & -- & -- & -- & -- & -- & -- & -- & -- & 5.48 & 4.51 & 6.87 & 6.71 \\
Uncertainty & -- & -- & -- & -- & -- & -- & -- & -- & 5.28 & 4.40 & 6.39 & 6.69 \\
\midrule
\multicolumn{13}{@{\hspace{3pt}}l}{\textit{Decision-aware prompting}} \\
Ask-P \citep{zhong2026finriskatlas} & 6.97 & 6.69 & 6.25 & 7.71 & 1.49 & 1.39 & 1.87 & 1.74 & 5.08 & 4.46 & 6.38 & 6.11 \\
Ask-P + tuned stop & 0.89 & 1.07 & 0.31 & 2.97 & 2.62 & 1.76 & 3.70 & 4.37 & 4.50 & 4.43 & 4.75 & 5.23 \\
VoI-Ask \citep{dong2026value} & 0.80 & 1.12 & 1.30 & 2.73 & 1.88 & 1.39 & 3.00 & 3.66 & 4.39 & 4.03 & 3.40 & 4.43 \\
\midrule
\multicolumn{13}{@{\hspace{3pt}}l}{\textit{Elicited model}} \\
DeLLMa \citep{liu2025dellma} & 4.54 & 4.14 & 2.67 & 4.90 & 2.06 & 1.54 & 2.62 & 2.92 & 4.77 & 4.25 & 4.97 & 5.54 \\
\midrule
\multicolumn{13}{@{\hspace{3pt}}l}{\textit{Information gain}} \\
\begin{tabular}[c]{@{}l@{}}BED-LLM\\\citep{choudhury2026bedllm}\end{tabular} & 6.24 & 5.90 & 3.46 & 7.12 & 1.30 & 1.28 & 1.89 & 1.81 & 5.50 & 4.94 & 7.07 & 6.07 \\
UoT \citep{hu2024uncertainty} & 9.23 & 6.67 & 6.19 & 7.75 & 1.61 & 1.58 & 1.69 & 1.61 & 6.38 & 4.88 & 7.00 & 7.04 \\
EIG (Beliefs) & -- & -- & -- & -- & -- & -- & -- & -- & 4.94 & 4.72 & 6.28 & 5.58 \\
Cost-aware EIG & \underline{0.43} & \textbf{0.43} & \underline{0.25} & \underline{1.53} & 1.07 & \underline{0.80} & 2.31 & 2.72 & \underline{1.04} & \textbf{0.95} & \underline{0.81} & \underline{0.95} \\
\midrule
\multicolumn{13}{@{\hspace{3pt}}l}{\textit{Program-aided}} \\
VoI-Tool & 1.45 & 1.78 & 0.72 & 2.08 & 1.44 & 1.21 & 1.96 & 2.07 & 1.41 & 1.36 & 1.09 & 1.32 \\
\midrule
\multicolumn{13}{@{\hspace{3pt}}l}{\textit{Reference}} \\
Never & 0.89 & 1.06 & 0.31 & 2.96 & 2.64 & 1.76 & 3.71 & 4.38 & 4.63 & 4.54 & 4.52 & 4.98 \\
Random & -- & -- & -- & -- & -- & -- & -- & -- & 6.19 & 4.63 & 6.90 & 7.16 \\
\midrule
\multicolumn{13}{@{\hspace{3pt}}l}{\textit{Ours}} \\
\rowcolor{oursrow} \diva{} & 0.48 & 0.65 & 0.26 & 1.71 & \underline{1.01} & 0.93 & 2.45 & 2.80 &  &  &  &  \\
\rowcolor{oursrow} \diva{}-U & \textbf{0.41} & \underline{0.52} & \textbf{0.22} & \textbf{1.12} & \textbf{0.67} & \textbf{0.71} & \textbf{0.83} & \textbf{0.74} & \multirow{-2}{*}{\textbf{0.86}} & \multirow{-2}{*}{\underline{0.97}} & \multirow{-2}{*}{\textbf{0.73}} & \multirow{-2}{*}{\textbf{0.87}} \\
\bottomrule
}
\caption{\textbf{Additional baselines.} DP regret on the first 300 test scenarios, averaged over 20 LLMs. --: Not run. DeLLMa and UoT are adapted to acquire-or-stop. Tuned stop: Stop threshold of each LLM, cross-fitted over scenario groups. Cost-aware EIG: EIG minus $\eta c/l$ on the decision model, with $\eta=2$ from development data.}
\label{tab:prior}\label{tab:extra_bl}\label{tab:f8_stop}
\end{table*}

%% file: tables/f8_mis.tex
\begin{table*}[t]
\centering
\small
\setlength{\tabcolsep}{3pt}
\begin{tabular*}{\textwidth}{@{\extracolsep{\fill}}l|cccccccc|cccc}
\toprule
\multirow{3}{*}[-1.4ex]{Decision model} & \multicolumn{8}{c|}{\textbf{Knowledge-required environments}} & \multicolumn{4}{c}{\textbf{Controlled environments}} \\
\cmidrule(lr){2-9}\cmidrule(lr){10-13}
 & \multicolumn{2}{c}{Credit} & \multicolumn{2}{c}{Clinical} & \multicolumn{4}{c|}{SEC filings (2025--2026)} & \multirow{2}{*}[-0.6ex]{Finance} & \multirow{2}{*}[-0.6ex]{Credit} & \multirow{2}{*}[-0.6ex]{Clinical} & \multirow{2}{*}[-0.6ex]{Legal} \\
\cmidrule(lr){2-3}\cmidrule(lr){4-5}\cmidrule(lr){6-9}
 & Taiwan & German & CDC & Heart & Ret-1d & Ret-5d & Vol-5d & \multicolumn{1}{c|}{Vol-20d} & & & & \\
\midrule
\diva{} & 0.48 & 0.65 & 0.26 & 1.71 & 1.01 & 0.93 & 2.45 & 2.80 &  &  &  &  \\
\diva{}-U & \textbf{0.41} & \textbf{0.52} & \textbf{0.22} & \textbf{1.12} & \textbf{0.67} & \textbf{0.71} & \textbf{0.83} & \textbf{0.74} & \multirow{-2}{*}{\textbf{0.86}} & \multirow{-2}{*}{\textbf{0.97}} & \multirow{-2}{*}{\textbf{0.73}} & \multirow{-2}{*}{\textbf{0.87}} \\
\midrule
\diva{}, $\pm$25\% & 0.52 & 0.70 & 0.31 & 1.78 & 1.23 & 1.06 & 2.55 & 3.02 &  &  &  &  \\
\diva{}-U, $\pm$25\% & \underline{0.42} & \underline{0.56} & \underline{0.24} & \underline{1.24} & \underline{0.68} & \textbf{0.71} & \underline{0.85} & \underline{0.75} & \multirow{-2}{*}{\underline{1.00}} & \multirow{-2}{*}{\underline{0.99}} & \multirow{-2}{*}{\underline{0.86}} & \multirow{-2}{*}{\underline{0.97}} \\
\midrule
\diva{}, $\pm$50\% & 0.67 & 0.85 & 0.38 & 1.96 & 1.70 & 1.26 & 2.94 & 3.37 &  &  &  &  \\
\diva{}-U, $\pm$50\% & 0.59 & 0.66 & 0.35 & 1.36 & 0.70 & \underline{0.72} & 0.96 & 0.94 & \multirow{-2}{*}{1.37} & \multirow{-2}{*}{1.26} & \multirow{-2}{*}{1.04} & \multirow{-2}{*}{1.23} \\
\bottomrule
\end{tabular*}
\caption{\textbf{Misspecified payoffs and costs.} DP regret when \diva{} decides with $g$, $l$, and every cost multiplied by independent factors drawn uniformly from $[1-p,1+p]$ and is scored with the true values, averaged over 20 LLMs.}
\label{tab:f8_mis}
\end{table*}

%% file: tables/predvoi.tex
\begin{table*}[t]
\centering
\small
\setlength{\tabcolsep}{3pt}
\begin{tabular*}{\textwidth}{@{\extracolsep{\fill}}ll|cccccccc|c}
\toprule
\multicolumn{2}{l|}{\multirow{2}{*}[-0.6ex]{Policy}} & \multicolumn{2}{c}{Credit} & \multicolumn{2}{c}{Clinical} & \multicolumn{4}{c|}{SEC filings (2025--2026)} & \multirow{2}{*}[-0.6ex]{Avg.} \\
\cmidrule(lr){3-4}\cmidrule(lr){5-6}\cmidrule(lr){7-10}
\multicolumn{2}{l|}{} & Taiwan & German & CDC & Heart & Ret-1d & Ret-5d & Vol-5d & \multicolumn{1}{c|}{Vol-20d} & \\
\midrule
 & \diva{} & 0.48 & 0.65 & \underline{0.26} & \underline{1.71} & \underline{1.01} & \underline{0.93} & \underline{2.45} & \underline{2.80} & \underline{1.29} \\
 & \diva{}-U & \textbf{0.41} & \textbf{0.52} & \textbf{0.22} & \textbf{1.12} & \textbf{0.67} & \textbf{0.71} & \textbf{0.83} & \textbf{0.74} & \textbf{0.65} \\
\multirow{-3}{*}{First decision} & \diva{}-PV & \underline{0.46} & \underline{0.64} & \underline{0.26} & \underline{1.71} & 1.02 & 0.94 & 2.57 & 2.99 & 1.32 \\
\midrule
 & Never & 0.84 & 1.25 & 0.36 & 3.38 & \underline{2.53} & 1.36 & \textbf{4.12} & \textbf{4.17} & 2.25 \\
 & \diva{} & 0.69 & 1.26 & \underline{0.35} & 2.66 & \textbf{1.65} & \underline{1.24} & 4.47 & 4.39 & \underline{2.09} \\
 & \diva{}-U & \underline{0.68} & \textbf{0.96} & \textbf{0.30} & \textbf{2.07} & 4.40 & 4.43 & 4.97 & 4.39 & 2.77 \\
\multirow{-4}{*}{Episodes} & \diva{}-PV & \textbf{0.65} & \underline{1.24} & \underline{0.35} & \underline{2.64} & \textbf{1.65} & \textbf{1.20} & \underline{4.37} & \underline{4.33} & \textbf{2.05} \\
\bottomrule
\end{tabular*}
\caption{\textbf{Coherent predictive VOI.} DP regret of the first decision (Scenarios of Table~\ref{tab:know}) and over full episodes (Test scenarios 0--149, final action from the reference model's belief), averaged over 20 LLMs.}
\label{tab:predvoi}
\end{table*}

%% file: tables/confirm.tex
\begin{table*}[t]
\centering
\small
\setlength{\tabcolsep}{3pt}
\begin{tabular}{lccccccccc}
\toprule
Env. & Direct & Never & Always ask & \diva{}-$\kappa^\ddagger$ & \diva{}-$\beta^\ddagger$ & \cellcolor{oursrow}\diva{} & \cellcolor{oursrow}\diva{}-U & \cellcolor{oursrow}\begin{tabular}[c]{@{}c@{}}Stop\\\diva{}-U\end{tabular} & \begin{tabular}[c]{@{}c@{}}Uninformative\\(\%, \diva{}-U\\/ Direct)\end{tabular} \\
\midrule
Ret-1d & 1.19 / 2.00 & 2.61 / 2.54 & 0.69 / 1.73 & 0.91 / \textbf{1.53} & \textbf{0.60} / 1.60 & \cellcolor{oursrow}0.93 / \underline{1.56} & \cellcolor{oursrow}\underline{0.64} / 1.66 & \cellcolor{oursrow}0.01 & 4.9 / 7.2 \\
Ret-5d & 1.17 / 1.47 & 0.99 / \textbf{0.90} & 0.87 / 1.32 & \underline{0.72} / \underline{0.91} & \textbf{0.66} / 1.04 & \cellcolor{oursrow}0.79 / 0.99 & \cellcolor{oursrow}0.79 / 1.20 & \cellcolor{oursrow}0.02 & 4.7 / 7.2 \\
Vol-5d & 1.08 / \textbf{1.61} & 3.35 / \underline{3.23} & \underline{0.81} / 3.57 & 2.18 / 3.56 & \underline{0.81} / 3.57 & \cellcolor{oursrow}2.20 / 3.49 & \cellcolor{oursrow}\textbf{0.79} / 3.51 & \cellcolor{oursrow}0.02 & 7.6 / 4.8 \\
Vol-20d & 1.05 / \textbf{1.61} & 3.78 / \underline{3.69} & 0.77 / 3.95 & 2.43 / 4.02 & \underline{0.76} / 3.92 & \cellcolor{oursrow}2.48 / 3.88 & \cellcolor{oursrow}\textbf{0.72} / 3.72 & \cellcolor{oursrow}0.03 & 9.9 / 3.9 \\
\bottomrule
\end{tabular}
\caption{\textbf{Held-out SEC filings.} DP / myopic regret beyond the first 1,000 scenarios per target, over 20 LLMs.}
\label{tab:confirm}
\end{table*}

%% file: tables/sec_nofilter.tex
\begin{table*}[t]
\centering
\small
\setlength{\tabcolsep}{2.5pt}
\begin{tabular*}{\textwidth}{@{\extracolsep{\fill}}l|cccc|cccc|cccc}
\toprule
\multirow{2}{*}[-0.6ex]{Policy} & \multicolumn{4}{c|}{As built} & \multicolumn{4}{c|}{Uniform} & \multicolumn{4}{c}{Far from belief} \\
\cmidrule(lr){2-5}\cmidrule(lr){6-9}\cmidrule(lr){10-13}
 & Ret-1d & Ret-5d & Vol-5d & \multicolumn{1}{c|}{Vol-20d} & Ret-1d & Ret-5d & Vol-5d & \multicolumn{1}{c|}{Vol-20d} & Ret-1d & Ret-5d & Vol-5d & Vol-20d \\
\midrule
Never & 2.64 & 1.76 & 3.71 & 4.38 & 0.56 & 0.46 & 1.07 & 1.64 & \textbf{0.10} & \textbf{0.01} & 0.38 & 0.52 \\
Always ask & 0.72 & 0.77 & 0.90 & 0.77 & 0.53 & 0.54 & 0.53 & 0.56 & 0.46 & 0.44 & 0.46 & 0.45 \\
Generic \diva{}-U & \textbf{0.61} & \textbf{0.65} & \textbf{0.65} & \textbf{0.60} & 0.43 & 0.46 & \underline{0.39} & \underline{0.46} & 0.39 & 0.37 & 0.35 & \underline{0.36} \\
\rowcolor{oursrow} \diva{} & 1.01 & 0.93 & 2.45 & 2.80 & \underline{0.30} & \underline{0.29} & 0.78 & 1.13 & \underline{0.14} & \underline{0.08} & 0.37 & 0.52 \\
\rowcolor{oursrow} \diva{}-DP & 0.86 & 0.77 & 2.02 & 2.37 & \textbf{0.29} & \textbf{0.26} & 0.69 & 0.96 & 0.15 & 0.11 & \underline{0.32} & 0.46 \\
\rowcolor{oursrow} \diva{}-U & \underline{0.67} & \underline{0.71} & \underline{0.83} & \underline{0.74} & 0.41 & 0.41 & \textbf{0.35} & \textbf{0.41} & 0.34 & 0.31 & \textbf{0.31} & \textbf{0.31} \\
\midrule
Stop: DP optimum & 0.37 & 0.45 & 0.35 & 0.30 & 0.83 & 0.84 & 0.73 & 0.70 & 0.93 & 0.97 & 0.85 & 0.79 \\
Stop: \diva{}-U & 0.01 & 0.02 & 0.02 & 0.03 & 0.26 & 0.30 & 0.37 & 0.34 & 0.31 & 0.35 & 0.46 & 0.45 \\
\bottomrule
\end{tabular*}
\caption{\textbf{SEC payoffs without the threshold filter.} DP regret, averaged over 20 LLMs, with $(g,l)$ as built (Threshold within 0.08 of the reference belief), drawn uniformly, or drawn only far from the reference belief, re-scoring stored decision models. LLM baselines saw the original payoffs and cannot be re-scored.}
\label{tab:sec_nofilter}
\end{table*}

%% file: tables/m_significance.tex
\begin{table}[t]
\centering
\small
\setlength{\tabcolsep}{0pt}
\begin{tabular*}{\columnwidth}{@{\extracolsep{\fill}}@{}l@{\hspace{4pt}}|@{\hspace{2pt}}ccc@{\hspace{4pt}}|@{\hspace{2pt}}cc@{}}
\toprule
\multirow{2}{*}[-0.7ex]{Env.} & \multicolumn{3}{c|}{$\Delta$ to never acquiring} & \multicolumn{2}{c}{\diva{}-U wins} \\
\cmidrule(lr){2-4}\cmidrule(lr){5-6}
 & \diva{} & \diva{}-U & \diva{}-DP$^\dagger$ & All & Sig. \\
\midrule
Taiwan & $-$0.41$^*$ & $-$0.48$^*$ & $-$0.44$^*$ & 20/20 & 16/20 \\
German & $-$0.41$^*$ & $-$0.55$^*$ & $-$0.07 & 20/20 & 17/20 \\
CDC & $-$0.05 & $-$0.09 & $-$0.06 & 18/20 & 18/20 \\
Heart & $-$1.26$^*$ & $-$1.84$^*$ & $-$1.01$^*$ & 20/20 & 20/20 \\
Ret-1d & $-$1.64$^*$ & $-$1.97$^*$ & $-$1.21$^*$ & 16/20 & 15/20 \\
Ret-5d & $-$0.84$^*$ & $-$1.05$^*$ & $-$0.18$^*$ & 18/20 & 13/20 \\
Vol-5d & $-$1.26$^*$ & $-$2.87$^*$ & $-$0.22$^*$ & 15/20 & 8/20 \\
Vol-20d & $-$1.57$^*$ & $-$3.64$^*$ & $-$0.14$^*$ & 7/20 & 2/20 \\
\bottomrule
\end{tabular*}
\caption{\textbf{Statistical significance.} Paired DP-regret differences ($\Delta$), and the LLMs where \diva{}-U beats every LLM baseline (All) and significantly (Sig.). $^*$: CI excludes zero. $^\dagger$: Pre-registered episodes (18 LLMs).}
\label{tab:ci_know}\label{tab:ci_main}\label{tab:sig_wins}
\end{table}

%% file: tables/m_para.tex
\begin{table}[t]
\centering
\small
\setlength{\tabcolsep}{4pt}
\filltabular{\linewidth}{5}{>{\ftL}l|cccc<{\ftR}}{
\toprule
Method & Original & Para.~1 & Para.~2 & Spread \\
\midrule
Direct & 5.26 & 5.31 & 5.18 & 0.39 \\
CoT \citep{wei2022chain} & 5.16 & 5.21 & 5.14 & 0.41 \\
VoI-P & 5.00 & 4.97 & 4.99 & 0.56 \\
PAL \citep{gao2023pal} & 4.39 & 4.29 & 4.62 & 0.44 \\
\rowcolor{oursrow} \diva{} & 0.86 & 0.73 & 0.93 & 0.23 \\
\rowcolor{oursrow} \diva{}-U (Records) & 0.57 & 0.58 & -- & 0.06 \\
\bottomrule
}
\caption{\textbf{Prompt paraphrases.} DP regret averaged over 20 LLMs on the tabular records (\diva{}-U only) and the controlled environments. Spread: Range over the prompts per LLM, averaged over LLMs.}
\label{tab:para_bl}\label{tab:para}
\end{table}

%% file: tables/qual.tex
\begin{figure*}[t]
\centering
\fbox{\parbox{0.96\textwidth}{\footnotesize You are an equity analyst at an investment fund. A company has just released its quarterly results. You must decide whether to BUY its shares now or PASS.\\
Payoffs: if you buy and the stock outperforms, you gain 1 point; if you buy and it underperforms, you lose 8 points. Passing yields 0 points either way.\\
Base rate: in 45\% of comparable companies in this situation, the stock underperforms the market over the next quarter.\\
Already known:\\
- Peer company results (already done): it did not find that the main competitors reported weak demand. If the stock underperforms the market over the next quarter, this check reports this finding in roughly 25 percent of cases; otherwise in 3 out of 20 cases.\\
You may run any of the following checks (each at most once) before deciding.\\
(A) Inventory levels (costs 0.6 points): reports whether inventories grew much faster than sales. If the stock underperforms the market over the next quarter, it reports this 75\% of the time; otherwise about 35\% of the time.\\
(B) Share buyback activity (costs 0.8 points): reports whether the company paused its share buyback program. If the stock underperforms the market over the next quarter, it reports this in roughly 55 percent of cases; otherwise about 50\% of the time.\\
(C) Customer concentration (costs 0.2 points): reports whether a single customer accounts for more than a quarter of revenue. If the stock underperforms the market over the next quarter, it reports this in 9 out of 20 cases; otherwise 35\% of the time.\\
(D) Forward guidance (costs 0.8 points): reports whether management cut its forward guidance. If the stock underperforms the market over the next quarter, it reports this 80\% of the time; otherwise 80\% of the time.\\
(E) Cash flow versus earnings (costs 0.4 points): reports whether operating cash flow fell well short of reported earnings. If the stock underperforms the market over the next quarter, it reports this in 15 out of 20 cases; otherwise in 9 out of 20 cases.\\
(F) Short interest (costs 0.4 points): reports whether short interest rose sharply over the last month. If the stock underperforms the market over the next quarter, it reports this in roughly 80 percent of cases; otherwise 5\% of the time.\\
(S) Stop and decide now (BUY or PASS).}}
\caption{\textbf{A controlled finance scenario.} The direct policy of Qwen3-8B requests (F) Short interest, the most informative check (EIG 0.48 bits), but under every outcome of every check the probability of underperformance stays above the threshold $b^\dagger=1/(1+8)$, and no check has decision value. \diva{} stops and passes (DP-optimal).}
\label{fig:qual}
\end{figure*}

%% file: tables/full_know_llm.tex
\begin{table*}[t]
\centering
\small
\setlength{\tabcolsep}{1.8pt}
\filltabular{\textwidth}{15}{>{\ftL}ll|ccccccccccccc<{\ftR}}{
\toprule
Family & Size & \begin{tabular}[c]{@{}c@{}}Direct\end{tabular} & \begin{tabular}[c]{@{}c@{}}CoT\end{tabular} & \begin{tabular}[c]{@{}c@{}}SC\end{tabular} & \begin{tabular}[c]{@{}c@{}}ReAct\end{tabular} & \begin{tabular}[c]{@{}c@{}}Ask-P\end{tabular} & \begin{tabular}[c]{@{}c@{}}VoI-P\end{tabular} & \begin{tabular}[c]{@{}c@{}}VoI-Ask\end{tabular} & \begin{tabular}[c]{@{}c@{}}BED-LLM\end{tabular} & \begin{tabular}[c]{@{}c@{}}PAL\end{tabular} & \begin{tabular}[c]{@{}c@{}}Always ask\end{tabular} & \cellcolor{oursrow}\begin{tabular}[c]{@{}c@{}}\diva{}\end{tabular} & \cellcolor{oursrow}\begin{tabular}[c]{@{}c@{}}\diva{}-U\end{tabular} & \cellcolor{oursrow}\begin{tabular}[c]{@{}c@{}}\diva{}-DP\end{tabular} \\
\midrule
\multirow{6}{*}{\textbf{Qwen3}} & 0.6B & 0.89 & 1.29 & 0.94 & 10.38 & 1.42 & 0.91 & \underline{0.70} & 0.89 & 1.80 & 5.56 & \cellcolor{oursrow}0.90 & \cellcolor{oursrow}\textbf{0.67} & \cellcolor{oursrow}0.82 \\
 & 1.7B & 4.97 & 6.91 & 6.40 & 10.34 & 4.34 & 4.36 & 0.77 & 10.07 & 6.91 & 5.51 & \cellcolor{oursrow}0.67 & \cellcolor{oursrow}\textbf{0.57} & \cellcolor{oursrow}\underline{0.60} \\
 & 4B & 10.22 & 10.46 & 10.37 & 10.50 & 10.77 & 11.28 & 0.70 & 4.45 & 2.73 & 5.15 & \cellcolor{oursrow}0.34 & \cellcolor{oursrow}\underline{0.33} & \cellcolor{oursrow}\textbf{0.32} \\
 & 8B & 6.56 & 8.79 & 9.75 & 8.90 & 3.18 & 9.16 & 0.71 & 5.10 & 2.86 & 5.23 & \cellcolor{oursrow}0.48 & \cellcolor{oursrow}\underline{0.43} & \cellcolor{oursrow}\textbf{0.37} \\
 & 14B & 6.33 & 7.68 & 7.28 & 8.71 & 9.55 & 6.54 & 0.77 & 8.30 & 2.24 & 5.00 & \cellcolor{oursrow}\underline{0.31} & \cellcolor{oursrow}0.32 & \cellcolor{oursrow}\textbf{0.26} \\
 & 32B & 3.63 & 6.01 & 5.84 & 5.89 & 8.42 & 3.57 & 0.80 & 7.27 & 2.11 & 5.34 & \cellcolor{oursrow}\underline{0.67} & \cellcolor{oursrow}0.68 & \cellcolor{oursrow}\textbf{0.57} \\
\cmidrule(lr){1-15}
\multirow{3}{*}{\textbf{Qwen2.5}} & 7B & 9.71 & 9.65 & 9.75 & 10.16 & 9.99 & 3.74 & 0.70 & 0.96 & 3.88 & 5.16 & \cellcolor{oursrow}0.36 & \cellcolor{oursrow}\underline{0.27} & \cellcolor{oursrow}\textbf{0.25} \\
 & 14B & 8.86 & 9.39 & 9.07 & 9.14 & 9.49 & 6.80 & 0.70 & 9.40 & 3.18 & 5.04 & \cellcolor{oursrow}\underline{0.29} & \cellcolor{oursrow}\underline{0.29} & \cellcolor{oursrow}\textbf{0.28} \\
 & 72B & 7.20 & 6.72 & 6.56 & 7.20 & 6.80 & 1.71 & 0.78 & 10.06 & 3.29 & 5.06 & \cellcolor{oursrow}0.36 & \cellcolor{oursrow}\underline{0.30} & \cellcolor{oursrow}\textbf{0.26} \\
\cmidrule(lr){1-15}
\multirow{2}{*}{\textbf{DeepSeek-R1}} & 7B & 6.85 & 1.38 & 3.02 & 9.54 & 0.89 & 7.08 & 0.94 & 0.89 & 5.69 & 5.11 & \cellcolor{oursrow}0.40 & \cellcolor{oursrow}\underline{0.39} & \cellcolor{oursrow}\textbf{0.37} \\
 & 14B & 2.77 & 8.01 & 7.79 & 9.29 & 1.65 & 7.79 & 0.80 & 1.01 & 5.29 & 5.04 & \cellcolor{oursrow}0.44 & \cellcolor{oursrow}\underline{0.42} & \cellcolor{oursrow}\textbf{0.41} \\
\cmidrule(lr){1-15}
\multirow{2}{*}{\textbf{Llama}} & 8B & 0.91 & 9.90 & 10.01 & 10.50 & 6.94 & 5.20 & 0.95 & 2.66 & 4.47 & 5.29 & \cellcolor{oursrow}0.50 & \cellcolor{oursrow}\textbf{0.30} & \cellcolor{oursrow}\underline{0.37} \\
 & 70B & 7.07 & 6.78 & 6.77 & 6.42 & 8.66 & 7.27 & 0.87 & 9.02 & 3.09 & 5.37 & \cellcolor{oursrow}\underline{0.69} & \cellcolor{oursrow}\underline{0.69} & \cellcolor{oursrow}\textbf{0.68} \\
\cmidrule(lr){1-15}
\multirow{3}{*}{\textbf{Gemma}} & 9B & 8.42 & 10.11 & 9.83 & 9.10 & 10.32 & 10.67 & 0.76 & 9.50 & 7.95 & 5.02 & \cellcolor{oursrow}0.44 & \cellcolor{oursrow}\underline{0.34} & \cellcolor{oursrow}\textbf{0.29} \\
 & 12B & 9.27 & 9.48 & 9.76 & 8.13 & 10.72 & 4.39 & 0.68 & 7.08 & 4.95 & 5.02 & \cellcolor{oursrow}\underline{0.45} & \cellcolor{oursrow}\underline{0.45} & \cellcolor{oursrow}\textbf{0.42} \\
 & 27B & 9.54 & 8.85 & 8.80 & 8.59 & 9.70 & 7.83 & 0.74 & 9.32 & 2.27 & 4.99 & \cellcolor{oursrow}\underline{0.28} & \cellcolor{oursrow}0.34 & \cellcolor{oursrow}\textbf{0.25} \\
\cmidrule(lr){1-15}
\multirow{2}{*}{\textbf{Mistral}} & 7B & 8.99 & 6.75 & 7.87 & 9.95 & 1.21 & 10.48 & 0.74 & 8.98 & 2.16 & 5.12 & \cellcolor{oursrow}\underline{0.39} & \cellcolor{oursrow}\textbf{0.27} & \cellcolor{oursrow}\textbf{0.27} \\
 & 24B & 7.29 & 7.38 & 7.39 & 7.17 & 7.12 & 6.29 & 1.03 & 8.71 & 2.86 & 5.01 & \cellcolor{oursrow}0.36 & \cellcolor{oursrow}\underline{0.26} & \cellcolor{oursrow}\textbf{0.22} \\
\cmidrule(lr){1-15}
\multirow{1}{*}{\textbf{Phi}} & 14B & 8.88 & 7.00 & 6.98 & 8.50 & 9.47 & 3.40 & 1.06 & 9.94 & 2.60 & 5.02 & \cellcolor{oursrow}\underline{0.42} & \cellcolor{oursrow}\textbf{0.35} & \cellcolor{oursrow}\textbf{0.35} \\
\cmidrule(lr){1-15}
\multirow{1}{*}{\textbf{gpt-oss}} & 20B & 1.03 & 7.52 & 6.27 & 7.10 & 8.85 & 0.93 & 0.90 & 1.27 & 0.93 & 5.37 & \cellcolor{oursrow}0.79 & \cellcolor{oursrow}\textbf{0.45} & \cellcolor{oursrow}\underline{0.71} \\
\midrule
\multicolumn{2}{@{}l|}{Never} & \multicolumn{13}{c@{}}{0.89 (Identical for every LLM)} \\
\bottomrule
}
\caption{\textbf{DP regret per LLM: Taiwan.} Same first 300 test scenarios as Table~\ref{tab:know}. Baselines are cited in Table~\ref{tab:know}.}
\label{tab:full_know_llm_taiwan}
\end{table*}

\begin{table*}[t]
\centering
\small
\setlength{\tabcolsep}{1.8pt}
\filltabular{\textwidth}{15}{>{\ftL}ll|ccccccccccccc<{\ftR}}{
\toprule
Family & Size & \begin{tabular}[c]{@{}c@{}}Direct\end{tabular} & \begin{tabular}[c]{@{}c@{}}CoT\end{tabular} & \begin{tabular}[c]{@{}c@{}}SC\end{tabular} & \begin{tabular}[c]{@{}c@{}}ReAct\end{tabular} & \begin{tabular}[c]{@{}c@{}}Ask-P\end{tabular} & \begin{tabular}[c]{@{}c@{}}VoI-P\end{tabular} & \begin{tabular}[c]{@{}c@{}}VoI-Ask\end{tabular} & \begin{tabular}[c]{@{}c@{}}BED-LLM\end{tabular} & \begin{tabular}[c]{@{}c@{}}PAL\end{tabular} & \begin{tabular}[c]{@{}c@{}}Always ask\end{tabular} & \cellcolor{oursrow}\begin{tabular}[c]{@{}c@{}}\diva{}\end{tabular} & \cellcolor{oursrow}\begin{tabular}[c]{@{}c@{}}\diva{}-U\end{tabular} & \cellcolor{oursrow}\begin{tabular}[c]{@{}c@{}}\diva{}-DP\end{tabular} \\
\midrule
\multirow{6}{*}{\textbf{Qwen3}} & 0.6B & 1.06 & 1.13 & 1.19 & 7.95 & 1.06 & 1.11 & 1.05 & 1.45 & 1.39 & 4.77 & \cellcolor{oursrow}1.09 & \cellcolor{oursrow}\underline{0.99} & \cellcolor{oursrow}\textbf{0.95} \\
 & 1.7B & 2.46 & 5.29 & 5.29 & 8.03 & 2.18 & 4.03 & 2.02 & 8.32 & 4.06 & 4.41 & \cellcolor{oursrow}0.61 & \cellcolor{oursrow}\underline{0.49} & \cellcolor{oursrow}\textbf{0.44} \\
 & 4B & 8.92 & 8.93 & 9.67 & 8.16 & 10.05 & 10.77 & 1.06 & 2.47 & 3.42 & 4.69 & \cellcolor{oursrow}\underline{0.85} & \cellcolor{oursrow}0.87 & \cellcolor{oursrow}\textbf{0.79} \\
 & 8B & 3.76 & 8.25 & 8.45 & 8.51 & 2.41 & 10.06 & 1.06 & 5.38 & 2.34 & 4.44 & \cellcolor{oursrow}0.68 & \cellcolor{oursrow}\textbf{0.46} & \cellcolor{oursrow}\underline{0.51} \\
 & 14B & 4.74 & 6.97 & 6.94 & 7.73 & 9.83 & 4.56 & 1.06 & 7.95 & 1.70 & 4.47 & \cellcolor{oursrow}0.54 & \cellcolor{oursrow}\underline{0.42} & \cellcolor{oursrow}\textbf{0.37} \\
 & 32B & 2.83 & 5.12 & 4.86 & 5.13 & 8.08 & 4.61 & 1.06 & 6.30 & 1.29 & 4.61 & \cellcolor{oursrow}0.78 & \cellcolor{oursrow}\underline{0.70} & \cellcolor{oursrow}\textbf{0.69} \\
\cmidrule(lr){1-15}
\multirow{3}{*}{\textbf{Qwen2.5}} & 7B & 7.33 & 8.21 & 8.60 & 8.88 & 8.32 & 3.82 & 1.07 & 7.60 & 4.54 & 4.30 & \cellcolor{oursrow}\underline{0.46} & \cellcolor{oursrow}\textbf{0.29} & \cellcolor{oursrow}\textbf{0.29} \\
 & 14B & 8.93 & 7.38 & 8.42 & 8.25 & 9.57 & 5.68 & 1.06 & 7.70 & 3.16 & 4.82 & \cellcolor{oursrow}0.75 & \cellcolor{oursrow}\underline{0.73} & \cellcolor{oursrow}\textbf{0.69} \\
 & 72B & 5.46 & 5.51 & 5.30 & 5.53 & 6.36 & 1.29 & 1.13 & 8.49 & 2.00 & 4.35 & \cellcolor{oursrow}0.47 & \cellcolor{oursrow}\underline{0.44} & \cellcolor{oursrow}\textbf{0.26} \\
\cmidrule(lr){1-15}
\multirow{2}{*}{\textbf{DeepSeek-R1}} & 7B & 5.52 & 1.23 & 3.49 & 8.50 & 1.06 & 5.15 & 0.98 & 1.06 & 4.39 & 4.34 & \cellcolor{oursrow}0.45 & \cellcolor{oursrow}\underline{0.38} & \cellcolor{oursrow}\textbf{0.33} \\
 & 14B & 2.63 & 7.44 & 6.53 & 7.70 & 2.53 & 6.33 & 1.06 & 1.33 & 2.72 & 4.33 & \cellcolor{oursrow}0.83 & \cellcolor{oursrow}\textbf{0.28} & \cellcolor{oursrow}\underline{0.76} \\
\cmidrule(lr){1-15}
\multirow{2}{*}{\textbf{Llama}} & 8B & 1.02 & 7.90 & 7.83 & 8.14 & 9.20 & 4.38 & 1.20 & 1.30 & 4.53 & 4.42 & \cellcolor{oursrow}0.51 & \cellcolor{oursrow}\underline{0.37} & \cellcolor{oursrow}\textbf{0.34} \\
 & 70B & 5.91 & 6.57 & 6.29 & 6.21 & 8.07 & 6.66 & 1.07 & 7.09 & 3.48 & 4.66 & \cellcolor{oursrow}\underline{0.82} & \cellcolor{oursrow}0.86 & \cellcolor{oursrow}\textbf{0.80} \\
\cmidrule(lr){1-15}
\multirow{3}{*}{\textbf{Gemma}} & 9B & 6.24 & 9.10 & 9.44 & 7.08 & 10.34 & 9.87 & 1.06 & 8.87 & 6.47 & 4.43 & \cellcolor{oursrow}0.52 & \cellcolor{oursrow}\underline{0.38} & \cellcolor{oursrow}\textbf{0.32} \\
 & 12B & 8.11 & 7.79 & 7.93 & 7.11 & 9.88 & 3.86 & 1.06 & 8.17 & 3.72 & 4.41 & \cellcolor{oursrow}0.51 & \cellcolor{oursrow}\underline{0.50} & \cellcolor{oursrow}\textbf{0.44} \\
 & 27B & 6.91 & 7.98 & 8.21 & 6.79 & 8.70 & 7.69 & 1.06 & 8.31 & 1.73 & 4.56 & \cellcolor{oursrow}\underline{0.65} & \cellcolor{oursrow}\textbf{0.50} & \cellcolor{oursrow}\textbf{0.50} \\
\cmidrule(lr){1-15}
\multirow{2}{*}{\textbf{Mistral}} & 7B & 7.05 & 8.19 & 8.52 & 7.96 & 1.30 & 9.18 & 1.06 & 8.07 & 3.44 & 4.45 & \cellcolor{oursrow}0.56 & \cellcolor{oursrow}\underline{0.43} & \cellcolor{oursrow}\textbf{0.32} \\
 & 24B & 5.79 & 5.92 & 5.88 & 5.77 & 8.67 & 6.23 & 1.08 & 8.32 & 3.14 & 4.37 & \cellcolor{oursrow}0.51 & \cellcolor{oursrow}\underline{0.43} & \cellcolor{oursrow}\textbf{0.35} \\
\cmidrule(lr){1-15}
\multirow{1}{*}{\textbf{Phi}} & 14B & 7.09 & 5.78 & 5.81 & 6.70 & 8.24 & 3.14 & 1.07 & 8.14 & 2.40 & 4.41 & \cellcolor{oursrow}\underline{0.55} & \cellcolor{oursrow}0.56 & \cellcolor{oursrow}\textbf{0.45} \\
\cmidrule(lr){1-15}
\multirow{1}{*}{\textbf{gpt-oss}} & 20B & 1.34 & 5.88 & 4.78 & 4.21 & 7.89 & 1.32 & 1.08 & 1.76 & 1.11 & 4.29 & \cellcolor{oursrow}0.91 & \cellcolor{oursrow}\textbf{0.24} & \cellcolor{oursrow}\underline{0.70} \\
\midrule
\multicolumn{2}{@{}l|}{Never} & \multicolumn{13}{c@{}}{1.06 (Identical for every LLM)} \\
\bottomrule
}
\caption{\textbf{DP regret per LLM: German.} Same first 300 test scenarios as Table~\ref{tab:know}. Baselines are cited in Table~\ref{tab:know}.}
\label{tab:full_know_llm_german}
\end{table*}

\begin{table*}[t]
\centering
\small
\setlength{\tabcolsep}{1.8pt}
\filltabular{\textwidth}{15}{>{\ftL}ll|ccccccccccccc<{\ftR}}{
\toprule
Family & Size & \begin{tabular}[c]{@{}c@{}}Direct\end{tabular} & \begin{tabular}[c]{@{}c@{}}CoT\end{tabular} & \begin{tabular}[c]{@{}c@{}}SC\end{tabular} & \begin{tabular}[c]{@{}c@{}}ReAct\end{tabular} & \begin{tabular}[c]{@{}c@{}}Ask-P\end{tabular} & \begin{tabular}[c]{@{}c@{}}VoI-P\end{tabular} & \begin{tabular}[c]{@{}c@{}}VoI-Ask\end{tabular} & \begin{tabular}[c]{@{}c@{}}BED-LLM\end{tabular} & \begin{tabular}[c]{@{}c@{}}PAL\end{tabular} & \begin{tabular}[c]{@{}c@{}}Always ask\end{tabular} & \cellcolor{oursrow}\begin{tabular}[c]{@{}c@{}}\diva{}\end{tabular} & \cellcolor{oursrow}\begin{tabular}[c]{@{}c@{}}\diva{}-U\end{tabular} & \cellcolor{oursrow}\begin{tabular}[c]{@{}c@{}}\diva{}-DP\end{tabular} \\
\midrule
\multirow{6}{*}{\textbf{Qwen3}} & 0.6B & \textbf{0.31} & 1.31 & \underline{1.00} & 9.26 & \textbf{0.31} & \underline{1.00} & 1.87 & \textbf{0.31} & 1.49 & 5.38 & \cellcolor{oursrow}1.41 & \cellcolor{oursrow}1.64 & \cellcolor{oursrow}1.68 \\
 & 1.7B & 2.88 & 5.80 & 6.23 & 8.95 & 0.46 & 4.81 & 0.45 & 7.76 & 2.12 & 4.82 & \cellcolor{oursrow}\underline{0.24} & \cellcolor{oursrow}\textbf{0.22} & \cellcolor{oursrow}0.25 \\
 & 4B & 9.19 & 8.58 & 9.20 & 8.70 & 9.16 & 9.91 & 0.31 & 0.31 & 2.78 & 4.75 & \cellcolor{oursrow}\textbf{0.16} & \cellcolor{oursrow}\underline{0.19} & \cellcolor{oursrow}\textbf{0.16} \\
 & 8B & 4.84 & 8.75 & 8.48 & 9.03 & 4.00 & 6.50 & 0.72 & 0.31 & 1.43 & 4.81 & \cellcolor{oursrow}0.24 & \cellcolor{oursrow}\textbf{0.14} & \cellcolor{oursrow}\underline{0.21} \\
 & 14B & 6.17 & 7.58 & 7.59 & 7.96 & 8.21 & 3.81 & 1.28 & 0.38 & 1.05 & 4.73 & \cellcolor{oursrow}0.11 & \cellcolor{oursrow}\underline{0.08} & \cellcolor{oursrow}\textbf{0.07} \\
 & 32B & 1.91 & 6.73 & 6.88 & 6.26 & 8.31 & 4.96 & 1.36 & 0.31 & 0.84 & 4.68 & \cellcolor{oursrow}0.11 & \cellcolor{oursrow}\underline{0.08} & \cellcolor{oursrow}\textbf{0.06} \\
\cmidrule(lr){1-15}
\multirow{3}{*}{\textbf{Qwen2.5}} & 7B & 9.24 & 8.64 & 8.83 & 8.88 & 9.51 & 4.99 & 0.31 & 2.30 & 3.30 & 4.76 & \cellcolor{oursrow}\underline{0.17} & \cellcolor{oursrow}\textbf{0.12} & \cellcolor{oursrow}\textbf{0.12} \\
 & 14B & 9.04 & 8.88 & 8.81 & 8.31 & 9.10 & 7.51 & 1.79 & 7.08 & 3.18 & 4.71 & \cellcolor{oursrow}0.16 & \cellcolor{oursrow}\underline{0.10} & \cellcolor{oursrow}\textbf{0.08} \\
 & 72B & 6.48 & 6.50 & 6.49 & 5.57 & 8.09 & 2.03 & 0.31 & 3.32 & 1.51 & 4.67 & \cellcolor{oursrow}\underline{0.15} & \cellcolor{oursrow}\textbf{0.06} & \cellcolor{oursrow}\textbf{0.06} \\
\cmidrule(lr){1-15}
\multirow{2}{*}{\textbf{DeepSeek-R1}} & 7B & 1.35 & 0.51 & 2.36 & 7.89 & \textbf{0.31} & 0.42 & 2.95 & \underline{0.34} & 2.72 & 5.03 & \cellcolor{oursrow}0.74 & \cellcolor{oursrow}0.68 & \cellcolor{oursrow}0.73 \\
 & 14B & 1.24 & 7.64 & 8.11 & 8.86 & 0.36 & 8.31 & 0.32 & 5.81 & 4.07 & 4.76 & \cellcolor{oursrow}0.15 & \cellcolor{oursrow}\underline{0.14} & \cellcolor{oursrow}\textbf{0.09} \\
\cmidrule(lr){1-15}
\multirow{2}{*}{\textbf{Llama}} & 8B & 1.56 & 8.53 & 8.56 & 8.82 & 0.33 & 2.03 & 1.98 & 6.87 & 3.76 & 4.72 & \cellcolor{oursrow}0.19 & \cellcolor{oursrow}\textbf{0.07} & \cellcolor{oursrow}\underline{0.14} \\
 & 70B & 6.13 & 6.80 & 7.02 & 6.67 & 8.40 & 5.71 & 0.35 & 8.31 & 4.09 & 4.73 & \cellcolor{oursrow}0.14 & \cellcolor{oursrow}\textbf{0.08} & \cellcolor{oursrow}\underline{0.10} \\
\cmidrule(lr){1-15}
\multirow{3}{*}{\textbf{Gemma}} & 9B & 7.98 & 9.44 & 9.07 & 9.38 & 10.65 & 10.85 & 2.47 & 0.64 & 5.82 & 4.74 & \cellcolor{oursrow}0.23 & \cellcolor{oursrow}\underline{0.16} & \cellcolor{oursrow}\textbf{0.12} \\
 & 12B & 5.52 & 7.27 & 7.37 & 7.39 & 8.57 & 6.25 & 0.73 & 0.31 & 3.29 & 4.77 & \cellcolor{oursrow}\underline{0.19} & \cellcolor{oursrow}\textbf{0.14} & \cellcolor{oursrow}\textbf{0.14} \\
 & 27B & 8.06 & 8.37 & 8.28 & 7.78 & 8.38 & 7.78 & 1.45 & 8.68 & 2.28 & 4.71 & \cellcolor{oursrow}0.16 & \cellcolor{oursrow}\underline{0.09} & \cellcolor{oursrow}\textbf{0.07} \\
\cmidrule(lr){1-15}
\multirow{2}{*}{\textbf{Mistral}} & 7B & 5.64 & 0.41 & 2.90 & 8.92 & 5.43 & 9.85 & 3.14 & 8.89 & 2.27 & 4.75 & \cellcolor{oursrow}\underline{0.16} & \cellcolor{oursrow}\underline{0.16} & \cellcolor{oursrow}\textbf{0.13} \\
 & 24B & 8.34 & 7.68 & 7.45 & 7.44 & 8.73 & 6.29 & 2.46 & 6.56 & 2.55 & 4.72 & \cellcolor{oursrow}0.21 & \cellcolor{oursrow}\textbf{0.10} & \cellcolor{oursrow}\underline{0.11} \\
\cmidrule(lr){1-15}
\multirow{1}{*}{\textbf{Phi}} & 14B & 8.33 & 6.06 & 6.93 & 7.91 & 8.21 & 2.88 & 1.40 & 0.31 & 1.51 & 4.69 & \cellcolor{oursrow}0.13 & \cellcolor{oursrow}\textbf{0.06} & \cellcolor{oursrow}\underline{0.07} \\
\cmidrule(lr){1-15}
\multirow{1}{*}{\textbf{gpt-oss}} & 20B & 0.32 & 5.66 & 4.93 & 6.09 & 8.38 & 2.44 & 0.31 & 0.31 & 0.45 & 4.72 & \cellcolor{oursrow}0.14 & \cellcolor{oursrow}\underline{0.10} & \cellcolor{oursrow}\textbf{0.06} \\
\midrule
\multicolumn{2}{@{}l|}{Never} & \multicolumn{13}{c@{}}{0.31 (Identical for every LLM)} \\
\bottomrule
}
\caption{\textbf{DP regret per LLM: CDC.} Same first 300 test scenarios as Table~\ref{tab:know}. Baselines are cited in Table~\ref{tab:know}.}
\label{tab:full_know_llm_cdc}
\end{table*}

\begin{table*}[t]
\centering
\small
\setlength{\tabcolsep}{1.8pt}
\filltabular{\textwidth}{15}{>{\ftL}ll|ccccccccccccc<{\ftR}}{
\toprule
Family & Size & \begin{tabular}[c]{@{}c@{}}Direct\end{tabular} & \begin{tabular}[c]{@{}c@{}}CoT\end{tabular} & \begin{tabular}[c]{@{}c@{}}SC\end{tabular} & \begin{tabular}[c]{@{}c@{}}ReAct\end{tabular} & \begin{tabular}[c]{@{}c@{}}Ask-P\end{tabular} & \begin{tabular}[c]{@{}c@{}}VoI-P\end{tabular} & \begin{tabular}[c]{@{}c@{}}VoI-Ask\end{tabular} & \begin{tabular}[c]{@{}c@{}}BED-LLM\end{tabular} & \begin{tabular}[c]{@{}c@{}}PAL\end{tabular} & \begin{tabular}[c]{@{}c@{}}Always ask\end{tabular} & \cellcolor{oursrow}\begin{tabular}[c]{@{}c@{}}\diva{}\end{tabular} & \cellcolor{oursrow}\begin{tabular}[c]{@{}c@{}}\diva{}-U\end{tabular} & \cellcolor{oursrow}\begin{tabular}[c]{@{}c@{}}\diva{}-DP\end{tabular} \\
\midrule
\multirow{6}{*}{\textbf{Qwen3}} & 0.6B & 2.96 & 4.13 & 3.55 & 8.33 & 2.96 & 3.28 & 2.97 & 2.98 & 4.21 & 4.45 & \cellcolor{oursrow}2.21 & \cellcolor{oursrow}\textbf{1.45} & \cellcolor{oursrow}\underline{1.94} \\
 & 1.7B & 4.01 & 8.15 & 8.67 & 8.33 & 7.02 & 5.74 & 2.96 & 7.63 & 5.38 & 3.96 & \cellcolor{oursrow}2.22 & \cellcolor{oursrow}\textbf{1.14} & \cellcolor{oursrow}\underline{2.03} \\
 & 4B & 8.66 & 8.46 & 8.20 & 7.65 & 10.24 & 9.26 & 2.98 & 5.34 & 3.71 & 4.05 & \cellcolor{oursrow}1.59 & \cellcolor{oursrow}\textbf{1.06} & \cellcolor{oursrow}\underline{1.27} \\
 & 8B & 7.64 & 8.02 & 7.76 & 8.35 & 5.95 & 8.90 & 2.76 & 7.20 & 3.20 & 4.50 & \cellcolor{oursrow}1.80 & \cellcolor{oursrow}\textbf{1.32} & \cellcolor{oursrow}\underline{1.52} \\
 & 14B & 6.87 & 7.68 & 7.57 & 8.11 & 9.14 & 7.08 & 3.13 & 7.37 & 4.54 & 4.06 & \cellcolor{oursrow}1.72 & \cellcolor{oursrow}\textbf{1.05} & \cellcolor{oursrow}\underline{1.35} \\
 & 32B & 4.00 & 5.54 & 4.78 & 5.90 & 7.25 & 5.33 & 2.75 & 7.09 & 3.34 & 3.98 & \cellcolor{oursrow}1.54 & \cellcolor{oursrow}\textbf{1.04} & \cellcolor{oursrow}\underline{1.36} \\
\cmidrule(lr){1-15}
\multirow{3}{*}{\textbf{Qwen2.5}} & 7B & 7.92 & 7.90 & 8.23 & 8.85 & 9.51 & 5.51 & 2.96 & 6.78 & 4.86 & 3.97 & \cellcolor{oursrow}1.39 & \cellcolor{oursrow}\textbf{0.96} & \cellcolor{oursrow}\underline{1.13} \\
 & 14B & 8.36 & 8.24 & 8.26 & 8.24 & 9.09 & 7.20 & 2.65 & 7.92 & 4.46 & 4.04 & \cellcolor{oursrow}1.52 & \cellcolor{oursrow}\textbf{1.06} & \cellcolor{oursrow}\underline{1.19} \\
 & 72B & 6.21 & 5.74 & 5.83 & 5.11 & 7.45 & 2.42 & 1.86 & 8.42 & 3.87 & 4.01 & \cellcolor{oursrow}1.43 & \cellcolor{oursrow}\textbf{0.99} & \cellcolor{oursrow}\underline{1.14} \\
\cmidrule(lr){1-15}
\multirow{2}{*}{\textbf{DeepSeek-R1}} & 7B & 7.94 & 2.88 & 4.94 & 7.52 & 2.96 & 4.87 & 2.98 & 3.71 & 4.91 & 4.52 & \cellcolor{oursrow}2.25 & \cellcolor{oursrow}\textbf{1.39} & \cellcolor{oursrow}\underline{1.88} \\
 & 14B & 4.74 & 7.27 & 7.29 & 8.15 & 9.00 & 7.49 & 2.88 & 8.07 & 5.21 & 4.16 & \cellcolor{oursrow}1.48 & \cellcolor{oursrow}\underline{1.15} & \cellcolor{oursrow}\textbf{1.12} \\
\cmidrule(lr){1-15}
\multirow{2}{*}{\textbf{Llama}} & 8B & 4.09 & 8.01 & 7.63 & 7.32 & 6.33 & 3.83 & 2.24 & 8.00 & 4.16 & 4.01 & \cellcolor{oursrow}1.52 & \cellcolor{oursrow}\textbf{1.00} & \cellcolor{oursrow}\underline{1.49} \\
 & 70B & 5.80 & 5.53 & 5.81 & 5.10 & 7.15 & 6.01 & 2.23 & 7.73 & 2.98 & 4.07 & \cellcolor{oursrow}1.36 & \cellcolor{oursrow}\underline{1.19} & \cellcolor{oursrow}\textbf{1.14} \\
\cmidrule(lr){1-15}
\multirow{3}{*}{\textbf{Gemma}} & 9B & 6.36 & 8.56 & 8.34 & 7.72 & 10.14 & 7.40 & 2.72 & 8.25 & 7.26 & 4.02 & \cellcolor{oursrow}1.88 & \cellcolor{oursrow}\textbf{1.37} & \cellcolor{oursrow}\underline{1.41} \\
 & 12B & 7.51 & 7.19 & 6.56 & 5.91 & 9.83 & 6.77 & 3.31 & 7.72 & 4.72 & 4.07 & \cellcolor{oursrow}1.81 & \cellcolor{oursrow}\textbf{1.13} & \cellcolor{oursrow}\underline{1.35} \\
 & 27B & 8.03 & 7.66 & 7.55 & 6.76 & 8.92 & 8.12 & 2.81 & 8.42 & 4.62 & 3.99 & \cellcolor{oursrow}1.55 & \cellcolor{oursrow}\textbf{1.02} & \cellcolor{oursrow}\underline{1.22} \\
\cmidrule(lr){1-15}
\multirow{2}{*}{\textbf{Mistral}} & 7B & 8.34 & 9.14 & 8.92 & 7.93 & 5.81 & 9.18 & 2.83 & 7.38 & 4.26 & 3.81 & \cellcolor{oursrow}1.10 & \cellcolor{oursrow}\underline{0.97} & \cellcolor{oursrow}\textbf{0.94} \\
 & 24B & 6.79 & 5.19 & 5.48 & 6.39 & 8.97 & 6.11 & 2.25 & 7.70 & 3.89 & 4.05 & \cellcolor{oursrow}2.00 & \cellcolor{oursrow}\textbf{1.06} & \cellcolor{oursrow}\underline{1.59} \\
\cmidrule(lr){1-15}
\multirow{1}{*}{\textbf{Phi}} & 14B & 6.72 & 5.80 & 5.79 & 6.39 & 8.58 & 5.87 & 2.59 & 7.55 & 3.62 & 3.97 & \cellcolor{oursrow}1.58 & \cellcolor{oursrow}\textbf{1.01} & \cellcolor{oursrow}\underline{1.32} \\
\cmidrule(lr){1-15}
\multirow{1}{*}{\textbf{gpt-oss}} & 20B & 3.13 & 6.28 & 5.04 & 4.85 & 7.84 & 3.46 & 2.83 & 7.16 & 2.96 & 3.81 & \cellcolor{oursrow}2.16 & \cellcolor{oursrow}\textbf{1.00} & \cellcolor{oursrow}\underline{1.89} \\
\midrule
\multicolumn{2}{@{}l|}{Never} & \multicolumn{13}{c@{}}{2.96 (Identical for every LLM)} \\
\bottomrule
}
\caption{\textbf{DP regret per LLM: Heart.} Same first 300 test scenarios as Table~\ref{tab:know}. Baselines are cited in Table~\ref{tab:know}.}
\label{tab:full_know_llm_heart}
\end{table*}

\begin{table*}[t]
\centering
\small
\setlength{\tabcolsep}{1.8pt}
\filltabular{\textwidth}{15}{>{\ftL}ll|ccccccccccccc<{\ftR}}{
\toprule
Family & Size & \begin{tabular}[c]{@{}c@{}}Direct\end{tabular} & \begin{tabular}[c]{@{}c@{}}CoT\end{tabular} & \begin{tabular}[c]{@{}c@{}}SC\end{tabular} & \begin{tabular}[c]{@{}c@{}}ReAct\end{tabular} & \begin{tabular}[c]{@{}c@{}}Ask-P\end{tabular} & \begin{tabular}[c]{@{}c@{}}VoI-P\end{tabular} & \begin{tabular}[c]{@{}c@{}}VoI-Ask\end{tabular} & \begin{tabular}[c]{@{}c@{}}BED-LLM\end{tabular} & \begin{tabular}[c]{@{}c@{}}PAL\end{tabular} & \begin{tabular}[c]{@{}c@{}}Always ask\end{tabular} & \cellcolor{oursrow}\begin{tabular}[c]{@{}c@{}}\diva{}\end{tabular} & \cellcolor{oursrow}\begin{tabular}[c]{@{}c@{}}\diva{}-U\end{tabular} & \cellcolor{oursrow}\begin{tabular}[c]{@{}c@{}}\diva{}-DP\end{tabular} \\
\midrule
\multirow{6}{*}{\textbf{Qwen3}} & 0.6B & 2.64 & 2.37 & 2.42 & 1.53 & 2.63 & 2.54 & 1.96 & 0.84 & 2.52 & \textbf{0.71} & \cellcolor{oursrow}2.42 & \cellcolor{oursrow}\underline{0.74} & \cellcolor{oursrow}2.35 \\
 & 1.7B & 2.22 & 1.76 & 1.76 & 1.55 & 2.64 & 2.25 & 2.29 & 1.36 & 2.38 & \underline{1.22} & \cellcolor{oursrow}1.61 & \cellcolor{oursrow}\textbf{1.11} & \cellcolor{oursrow}1.41 \\
 & 4B & \textbf{0.71} & 1.03 & 1.07 & 1.24 & 1.15 & 2.45 & 2.64 & 1.47 & 2.16 & 0.98 & \cellcolor{oursrow}1.04 & \cellcolor{oursrow}\underline{0.91} & \cellcolor{oursrow}0.99 \\
 & 8B & 0.82 & 1.08 & 0.93 & 1.07 & 1.73 & 2.51 & 1.41 & 1.58 & 2.27 & 0.67 & \cellcolor{oursrow}0.70 & \cellcolor{oursrow}\underline{0.61} & \cellcolor{oursrow}\textbf{0.55} \\
 & 14B & 0.81 & 0.83 & 0.75 & 0.81 & 0.82 & 2.02 & 1.94 & 1.19 & 1.73 & 0.59 & \cellcolor{oursrow}0.68 & \cellcolor{oursrow}\textbf{0.52} & \cellcolor{oursrow}\underline{0.55} \\
 & 32B & 1.32 & 0.61 & \underline{0.59} & 0.84 & 0.91 & 0.82 & 1.94 & 1.22 & 1.53 & 0.66 & \cellcolor{oursrow}\underline{0.59} & \cellcolor{oursrow}0.60 & \cellcolor{oursrow}\textbf{0.58} \\
\cmidrule(lr){1-15}
\multirow{3}{*}{\textbf{Qwen2.5}} & 7B & 1.67 & 1.09 & 1.09 & 1.50 & 1.50 & 1.22 & 2.64 & 1.25 & 1.88 & \underline{0.69} & \cellcolor{oursrow}1.13 & \cellcolor{oursrow}\textbf{0.68} & \cellcolor{oursrow}0.94 \\
 & 14B & 1.10 & 1.00 & 0.94 & 1.16 & 0.99 & 1.22 & 2.45 & 1.33 & 1.95 & 0.64 & \cellcolor{oursrow}0.84 & \cellcolor{oursrow}\underline{0.62} & \cellcolor{oursrow}\textbf{0.60} \\
 & 72B & 0.62 & 0.70 & 0.71 & 0.79 & 0.70 & 1.04 & 0.89 & 0.99 & 1.91 & 0.58 & \cellcolor{oursrow}0.57 & \cellcolor{oursrow}\underline{0.54} & \cellcolor{oursrow}\textbf{0.46} \\
\cmidrule(lr){1-15}
\multirow{2}{*}{\textbf{DeepSeek-R1}} & 7B & 1.37 & 2.64 & 2.66 & 1.61 & 2.64 & 2.60 & 1.91 & 2.51 & 1.64 & \textbf{1.05} & \cellcolor{oursrow}1.46 & \cellcolor{oursrow}\underline{1.06} & \cellcolor{oursrow}1.36 \\
 & 14B & 2.08 & 1.00 & 0.97 & 1.20 & 2.49 & 2.73 & 1.61 & 0.97 & 1.42 & \underline{0.68} & \cellcolor{oursrow}1.03 & \cellcolor{oursrow}\textbf{0.65} & \cellcolor{oursrow}0.94 \\
\cmidrule(lr){1-15}
\multirow{2}{*}{\textbf{Llama}} & 8B & 1.34 & 1.25 & 1.31 & 1.49 & 1.60 & 1.57 & 2.13 & 1.13 & 1.49 & \underline{0.65} & \cellcolor{oursrow}2.09 & \cellcolor{oursrow}\textbf{0.62} & \cellcolor{oursrow}1.69 \\
 & 70B & 0.92 & 0.83 & 0.76 & 0.82 & 1.12 & 0.90 & \textbf{0.52} & 1.04 & 1.39 & 0.66 & \cellcolor{oursrow}0.66 & \cellcolor{oursrow}0.62 & \cellcolor{oursrow}\underline{0.61} \\
\cmidrule(lr){1-15}
\multirow{3}{*}{\textbf{Gemma}} & 9B & 1.08 & 1.13 & 1.16 & 1.31 & 1.22 & 1.19 & 2.12 & 1.53 & 1.54 & \underline{0.58} & \cellcolor{oursrow}0.75 & \cellcolor{oursrow}\textbf{0.56} & \cellcolor{oursrow}0.59 \\
 & 12B & 1.09 & 1.10 & 0.87 & 1.23 & 0.96 & 1.34 & 2.28 & 1.15 & 1.86 & 0.74 & \cellcolor{oursrow}0.99 & \cellcolor{oursrow}\textbf{0.66} & \cellcolor{oursrow}\underline{0.67} \\
 & 27B & 0.82 & 0.88 & 0.80 & 0.85 & 0.93 & 0.98 & 1.66 & 1.28 & 1.55 & 0.64 & \cellcolor{oursrow}0.65 & \cellcolor{oursrow}\underline{0.61} & \cellcolor{oursrow}\textbf{0.53} \\
\cmidrule(lr){1-15}
\multirow{2}{*}{\textbf{Mistral}} & 7B & 1.41 & 1.89 & 1.97 & 1.39 & 1.83 & 2.59 & 2.05 & 1.24 & 2.58 & \underline{0.69} & \cellcolor{oursrow}0.93 & \cellcolor{oursrow}\textbf{0.63} & \cellcolor{oursrow}0.79 \\
 & 24B & 1.08 & 1.00 & 1.05 & 1.14 & 1.33 & 1.35 & 1.81 & 1.25 & 1.57 & 0.48 & \cellcolor{oursrow}\underline{0.38} & \cellcolor{oursrow}0.45 & \cellcolor{oursrow}\textbf{0.26} \\
\cmidrule(lr){1-15}
\multirow{1}{*}{\textbf{Phi}} & 14B & 0.93 & 0.78 & 0.73 & 0.85 & 1.65 & 1.16 & 2.64 & 1.50 & 1.62 & \underline{0.58} & \cellcolor{oursrow}0.61 & \cellcolor{oursrow}\textbf{0.55} & \cellcolor{oursrow}\textbf{0.55} \\
\cmidrule(lr){1-15}
\multirow{1}{*}{\textbf{gpt-oss}} & 20B & 1.20 & 0.86 & 0.77 & 0.94 & 0.87 & 2.47 & \textbf{0.66} & 1.22 & 2.62 & 0.83 & \cellcolor{oursrow}0.98 & \cellcolor{oursrow}\underline{0.71} & \cellcolor{oursrow}0.72 \\
\midrule
\multicolumn{2}{@{}l|}{Never} & \multicolumn{13}{c@{}}{2.64 (Identical for every LLM)} \\
\bottomrule
}
\caption{\textbf{DP regret per LLM: Ret-1d.} Same first 300 test scenarios as Table~\ref{tab:know}. Baselines are cited in Table~\ref{tab:know}.}
\label{tab:full_know_llm_ret_1}
\end{table*}

\begin{table*}[t]
\centering
\small
\setlength{\tabcolsep}{1.8pt}
\filltabular{\textwidth}{15}{>{\ftL}ll|ccccccccccccc<{\ftR}}{
\toprule
Family & Size & \begin{tabular}[c]{@{}c@{}}Direct\end{tabular} & \begin{tabular}[c]{@{}c@{}}CoT\end{tabular} & \begin{tabular}[c]{@{}c@{}}SC\end{tabular} & \begin{tabular}[c]{@{}c@{}}ReAct\end{tabular} & \begin{tabular}[c]{@{}c@{}}Ask-P\end{tabular} & \begin{tabular}[c]{@{}c@{}}VoI-P\end{tabular} & \begin{tabular}[c]{@{}c@{}}VoI-Ask\end{tabular} & \begin{tabular}[c]{@{}c@{}}BED-LLM\end{tabular} & \begin{tabular}[c]{@{}c@{}}PAL\end{tabular} & \begin{tabular}[c]{@{}c@{}}Always ask\end{tabular} & \cellcolor{oursrow}\begin{tabular}[c]{@{}c@{}}\diva{}\end{tabular} & \cellcolor{oursrow}\begin{tabular}[c]{@{}c@{}}\diva{}-U\end{tabular} & \cellcolor{oursrow}\begin{tabular}[c]{@{}c@{}}\diva{}-DP\end{tabular} \\
\midrule
\multirow{6}{*}{\textbf{Qwen3}} & 0.6B & 1.76 & 1.70 & 1.70 & 1.60 & 1.69 & 1.74 & 1.56 & 0.76 & 1.65 & \underline{0.73} & \cellcolor{oursrow}1.67 & \cellcolor{oursrow}\textbf{0.68} & \cellcolor{oursrow}1.57 \\
 & 1.7B & 1.72 & 1.65 & 1.69 & 1.72 & 1.81 & 1.89 & 1.24 & 1.53 & 1.72 & \underline{1.03} & \cellcolor{oursrow}1.51 & \cellcolor{oursrow}\textbf{1.00} & \cellcolor{oursrow}1.38 \\
 & 4B & \textbf{0.82} & 1.33 & 1.22 & 1.49 & 1.29 & 1.60 & 1.70 & 1.63 & 1.69 & 0.99 & \cellcolor{oursrow}1.07 & \cellcolor{oursrow}0.91 & \cellcolor{oursrow}\underline{0.86} \\
 & 8B & 0.92 & 1.16 & 1.14 & 1.22 & 1.47 & 1.93 & 1.07 & 1.49 & 1.72 & 0.77 & \cellcolor{oursrow}0.72 & \cellcolor{oursrow}\underline{0.71} & \cellcolor{oursrow}\textbf{0.61} \\
 & 14B & 0.82 & 1.04 & 0.96 & 1.11 & 0.98 & 1.79 & 1.52 & 1.29 & 1.57 & 0.65 & \cellcolor{oursrow}0.67 & \cellcolor{oursrow}\underline{0.63} & \cellcolor{oursrow}\textbf{0.58} \\
 & 32B & 1.05 & 0.82 & 0.79 & 0.94 & 1.06 & 1.05 & 1.33 & 1.31 & 1.34 & 0.73 & \cellcolor{oursrow}\underline{0.61} & \cellcolor{oursrow}0.69 & \cellcolor{oursrow}\textbf{0.57} \\
\cmidrule(lr){1-15}
\multirow{3}{*}{\textbf{Qwen2.5}} & 7B & 1.45 & 1.22 & 1.30 & 1.55 & 1.49 & 1.49 & 1.76 & 1.19 & 1.62 & \underline{0.79} & \cellcolor{oursrow}1.06 & \cellcolor{oursrow}\textbf{0.74} & \cellcolor{oursrow}0.80 \\
 & 14B & 1.10 & 1.23 & 1.17 & 1.31 & 1.18 & 1.35 & 1.60 & 1.39 & 1.44 & 0.70 & \cellcolor{oursrow}\underline{0.69} & \cellcolor{oursrow}\underline{0.69} & \cellcolor{oursrow}\textbf{0.55} \\
 & 72B & 0.82 & 0.81 & 0.88 & 0.94 & 0.86 & 0.75 & 0.79 & 1.12 & 1.56 & 0.75 & \cellcolor{oursrow}0.67 & \cellcolor{oursrow}\underline{0.63} & \cellcolor{oursrow}\textbf{0.53} \\
\cmidrule(lr){1-15}
\multirow{2}{*}{\textbf{DeepSeek-R1}} & 7B & 1.30 & 1.76 & 1.82 & 1.47 & 1.76 & 1.96 & 1.33 & 1.40 & 1.38 & \underline{0.91} & \cellcolor{oursrow}1.32 & \cellcolor{oursrow}\textbf{0.84} & \cellcolor{oursrow}1.13 \\
 & 14B & 1.48 & 1.18 & 1.18 & 1.48 & 1.62 & 2.03 & 1.49 & 1.00 & 1.21 & \underline{0.68} & \cellcolor{oursrow}0.85 & \cellcolor{oursrow}\textbf{0.66} & \cellcolor{oursrow}0.77 \\
\cmidrule(lr){1-15}
\multirow{2}{*}{\textbf{Llama}} & 8B & 1.20 & 1.32 & 1.48 & 1.53 & 1.57 & 1.35 & 1.71 & 1.22 & 1.31 & \underline{0.66} & \cellcolor{oursrow}1.64 & \cellcolor{oursrow}\textbf{0.64} & \cellcolor{oursrow}1.40 \\
 & 70B & 1.07 & 0.93 & 0.99 & 0.93 & 1.21 & 1.03 & \underline{0.72} & 1.08 & 1.31 & 0.81 & \cellcolor{oursrow}0.82 & \cellcolor{oursrow}\underline{0.72} & \cellcolor{oursrow}\textbf{0.61} \\
\cmidrule(lr){1-15}
\multirow{3}{*}{\textbf{Gemma}} & 9B & 1.28 & 1.56 & 1.52 & 1.61 & 1.49 & 1.54 & 1.41 & 1.53 & 1.41 & 0.65 & \cellcolor{oursrow}0.72 & \cellcolor{oursrow}\underline{0.61} & \cellcolor{oursrow}\textbf{0.54} \\
 & 12B & 1.30 & 1.21 & 1.28 & 1.39 & 1.17 & 1.34 & 1.52 & 1.27 & 1.45 & 0.84 & \cellcolor{oursrow}0.85 & \cellcolor{oursrow}\underline{0.74} & \cellcolor{oursrow}\textbf{0.70} \\
 & 27B & 1.18 & 1.20 & 1.12 & 1.00 & 1.18 & 1.22 & 1.40 & 1.32 & 1.27 & 0.69 & \cellcolor{oursrow}\underline{0.61} & \cellcolor{oursrow}0.62 & \cellcolor{oursrow}\textbf{0.48} \\
\cmidrule(lr){1-15}
\multirow{2}{*}{\textbf{Mistral}} & 7B & 1.49 & 1.82 & 1.82 & 1.42 & 1.80 & 2.24 & 1.64 & 1.24 & 1.71 & 0.80 & \cellcolor{oursrow}0.90 & \cellcolor{oursrow}\textbf{0.71} & \cellcolor{oursrow}\underline{0.75} \\
 & 24B & 1.22 & 0.99 & 1.07 & 1.38 & 1.37 & 1.38 & 1.56 & 1.12 & 1.55 & 0.60 & \cellcolor{oursrow}\underline{0.46} & \cellcolor{oursrow}0.55 & \cellcolor{oursrow}\textbf{0.35} \\
\cmidrule(lr){1-15}
\multirow{1}{*}{\textbf{Phi}} & 14B & 1.09 & 0.97 & 0.96 & 1.00 & 1.69 & 1.23 & 1.76 & 1.56 & 1.06 & 0.68 & \cellcolor{oursrow}0.68 & \cellcolor{oursrow}\underline{0.64} & \cellcolor{oursrow}\textbf{0.56} \\
\cmidrule(lr){1-15}
\multirow{1}{*}{\textbf{gpt-oss}} & 20B & 1.16 & 1.17 & 1.07 & 1.03 & 1.07 & 1.81 & \underline{0.72} & 1.19 & 1.75 & 0.88 & \cellcolor{oursrow}1.01 & \cellcolor{oursrow}0.79 & \cellcolor{oursrow}\textbf{0.70} \\
\midrule
\multicolumn{2}{@{}l|}{Never} & \multicolumn{13}{c@{}}{1.76 (Identical for every LLM)} \\
\bottomrule
}
\caption{\textbf{DP regret per LLM: Ret-5d.} Same first 300 test scenarios as Table~\ref{tab:know}. Baselines are cited in Table~\ref{tab:know}.}
\label{tab:full_know_llm_ret_5}
\end{table*}

\begin{table*}[t]
\centering
\small
\setlength{\tabcolsep}{1.8pt}
\filltabular{\textwidth}{15}{>{\ftL}ll|ccccccccccccc<{\ftR}}{
\toprule
Family & Size & \begin{tabular}[c]{@{}c@{}}Direct\end{tabular} & \begin{tabular}[c]{@{}c@{}}CoT\end{tabular} & \begin{tabular}[c]{@{}c@{}}SC\end{tabular} & \begin{tabular}[c]{@{}c@{}}ReAct\end{tabular} & \begin{tabular}[c]{@{}c@{}}Ask-P\end{tabular} & \begin{tabular}[c]{@{}c@{}}VoI-P\end{tabular} & \begin{tabular}[c]{@{}c@{}}VoI-Ask\end{tabular} & \begin{tabular}[c]{@{}c@{}}BED-LLM\end{tabular} & \begin{tabular}[c]{@{}c@{}}PAL\end{tabular} & \begin{tabular}[c]{@{}c@{}}Always ask\end{tabular} & \cellcolor{oursrow}\begin{tabular}[c]{@{}c@{}}\diva{}\end{tabular} & \cellcolor{oursrow}\begin{tabular}[c]{@{}c@{}}\diva{}-U\end{tabular} & \cellcolor{oursrow}\begin{tabular}[c]{@{}c@{}}\diva{}-DP\end{tabular} \\
\midrule
\multirow{6}{*}{\textbf{Qwen3}} & 0.6B & 3.71 & 2.83 & 3.32 & 1.50 & 3.59 & 3.71 & 1.79 & 3.71 & 3.40 & \textbf{0.75} & \cellcolor{oursrow}3.49 & \cellcolor{oursrow}\underline{0.82} & \cellcolor{oursrow}3.44 \\
 & 1.7B & 1.11 & 1.21 & 1.06 & 1.06 & 3.64 & 2.69 & 3.42 & 1.65 & 3.41 & \textbf{1.00} & \cellcolor{oursrow}2.76 & \cellcolor{oursrow}\underline{1.02} & \cellcolor{oursrow}2.67 \\
 & 4B & 1.29 & 1.17 & 1.11 & 1.06 & 1.22 & 1.76 & 3.31 & 1.60 & 2.89 & \underline{0.86} & \cellcolor{oursrow}2.41 & \cellcolor{oursrow}\textbf{0.79} & \cellcolor{oursrow}2.03 \\
 & 8B & 0.93 & 1.08 & 0.99 & 1.01 & 1.73 & 2.40 & 2.64 & 1.69 & 3.09 & \underline{0.78} & \cellcolor{oursrow}3.01 & \cellcolor{oursrow}\textbf{0.76} & \cellcolor{oursrow}2.54 \\
 & 14B & 1.54 & 0.94 & 1.02 & 0.94 & 1.28 & 3.76 & 3.44 & 1.65 & 2.60 & \underline{0.90} & \cellcolor{oursrow}2.44 & \cellcolor{oursrow}\textbf{0.84} & \cellcolor{oursrow}1.88 \\
 & 32B & 2.05 & \underline{0.91} & 0.98 & \textbf{0.90} & 1.10 & 1.17 & 3.26 & 1.64 & 2.47 & 0.98 & \cellcolor{oursrow}2.08 & \cellcolor{oursrow}0.92 & \cellcolor{oursrow}1.65 \\
\cmidrule(lr){1-15}
\multirow{3}{*}{\textbf{Qwen2.5}} & 7B & \textbf{0.76} & 1.17 & 1.08 & 1.24 & 1.00 & 1.62 & 3.71 & 1.68 & 2.96 & 0.98 & \cellcolor{oursrow}2.11 & \cellcolor{oursrow}\underline{0.85} & \cellcolor{oursrow}1.59 \\
 & 14B & 1.18 & 1.11 & 1.16 & 1.33 & 1.45 & 1.32 & 3.52 & 1.28 & 2.33 & \underline{0.83} & \cellcolor{oursrow}2.56 & \cellcolor{oursrow}\textbf{0.76} & \cellcolor{oursrow}2.18 \\
 & 72B & \underline{0.97} & 1.11 & 1.15 & \textbf{0.93} & 1.24 & 1.37 & 3.35 & 1.65 & 2.37 & 1.11 & \cellcolor{oursrow}1.95 & \cellcolor{oursrow}\textbf{0.93} & \cellcolor{oursrow}1.37 \\
\cmidrule(lr){1-15}
\multirow{2}{*}{\textbf{DeepSeek-R1}} & 7B & 1.55 & 3.22 & 2.84 & 1.39 & 3.71 & 2.64 & 3.13 & 3.35 & 2.32 & \underline{1.07} & \cellcolor{oursrow}1.84 & \cellcolor{oursrow}\textbf{0.90} & \cellcolor{oursrow}1.23 \\
 & 14B & 1.49 & 1.06 & 1.01 & 1.21 & 3.61 & 3.53 & 2.34 & 2.32 & 2.68 & \underline{0.71} & \cellcolor{oursrow}2.99 & \cellcolor{oursrow}\textbf{0.70} & \cellcolor{oursrow}2.72 \\
\cmidrule(lr){1-15}
\multirow{2}{*}{\textbf{Llama}} & 8B & 0.73 & 1.21 & 1.13 & 1.16 & 2.15 & 1.88 & 2.27 & 2.34 & 2.20 & \underline{0.69} & \cellcolor{oursrow}3.47 & \cellcolor{oursrow}\textbf{0.67} & \cellcolor{oursrow}3.06 \\
 & 70B & \underline{0.80} & 0.97 & 1.07 & \textbf{0.79} & 1.16 & 1.23 & 1.59 & 1.69 & 1.82 & 1.11 & \cellcolor{oursrow}1.73 & \cellcolor{oursrow}1.00 & \cellcolor{oursrow}1.18 \\
\cmidrule(lr){1-15}
\multirow{3}{*}{\textbf{Gemma}} & 9B & \textbf{0.76} & 1.12 & 1.10 & 0.94 & 1.12 & 1.38 & 2.65 & 1.60 & 1.51 & \underline{0.78} & \cellcolor{oursrow}2.27 & \cellcolor{oursrow}\textbf{0.76} & \cellcolor{oursrow}1.79 \\
 & 12B & \underline{0.85} & 1.01 & 1.01 & 1.22 & 1.11 & 1.37 & 3.42 & 1.61 & 2.57 & 1.01 & \cellcolor{oursrow}1.87 & \cellcolor{oursrow}\textbf{0.84} & \cellcolor{oursrow}1.42 \\
 & 27B & 1.10 & 1.11 & 1.02 & 1.01 & 1.11 & 1.30 & 3.37 & 1.64 & 1.21 & \underline{0.73} & \cellcolor{oursrow}2.37 & \cellcolor{oursrow}\textbf{0.72} & \cellcolor{oursrow}1.76 \\
\cmidrule(lr){1-15}
\multirow{2}{*}{\textbf{Mistral}} & 7B & 0.98 & 1.34 & 1.28 & 1.08 & 3.20 & 1.43 & 3.29 & 1.69 & 3.23 & \underline{0.75} & \cellcolor{oursrow}3.03 & \cellcolor{oursrow}\textbf{0.70} & \cellcolor{oursrow}2.47 \\
 & 24B & 0.97 & 0.92 & 0.97 & 0.94 & 0.98 & 1.41 & 3.56 & 1.74 & 2.23 & \underline{0.80} & \cellcolor{oursrow}2.85 & \cellcolor{oursrow}\textbf{0.71} & \cellcolor{oursrow}2.60 \\
\cmidrule(lr){1-15}
\multirow{1}{*}{\textbf{Phi}} & 14B & 0.96 & \textbf{0.86} & 0.94 & 1.06 & 1.92 & 0.99 & 3.71 & 1.61 & 1.82 & 1.04 & \cellcolor{oursrow}1.75 & \cellcolor{oursrow}\underline{0.91} & \cellcolor{oursrow}1.25 \\
\cmidrule(lr){1-15}
\multirow{1}{*}{\textbf{gpt-oss}} & 20B & 1.14 & 1.20 & 1.10 & \underline{1.04} & 1.16 & 3.62 & 2.34 & 1.73 & 3.71 & 1.09 & \cellcolor{oursrow}1.99 & \cellcolor{oursrow}\textbf{1.02} & \cellcolor{oursrow}1.48 \\
\midrule
\multicolumn{2}{@{}l|}{Never} & \multicolumn{13}{c@{}}{3.71 (Identical for every LLM)} \\
\bottomrule
}
\caption{\textbf{DP regret per LLM: Vol-5d.} Same first 300 test scenarios as Table~\ref{tab:know}. Baselines are cited in Table~\ref{tab:know}.}
\label{tab:full_know_llm_vol_5}
\end{table*}

\begin{table*}[t]
\centering
\small
\setlength{\tabcolsep}{1.8pt}
\filltabular{\textwidth}{15}{>{\ftL}ll|ccccccccccccc<{\ftR}}{
\toprule
Family & Size & \begin{tabular}[c]{@{}c@{}}Direct\end{tabular} & \begin{tabular}[c]{@{}c@{}}CoT\end{tabular} & \begin{tabular}[c]{@{}c@{}}SC\end{tabular} & \begin{tabular}[c]{@{}c@{}}ReAct\end{tabular} & \begin{tabular}[c]{@{}c@{}}Ask-P\end{tabular} & \begin{tabular}[c]{@{}c@{}}VoI-P\end{tabular} & \begin{tabular}[c]{@{}c@{}}VoI-Ask\end{tabular} & \begin{tabular}[c]{@{}c@{}}BED-LLM\end{tabular} & \begin{tabular}[c]{@{}c@{}}PAL\end{tabular} & \begin{tabular}[c]{@{}c@{}}Always ask\end{tabular} & \cellcolor{oursrow}\begin{tabular}[c]{@{}c@{}}\diva{}\end{tabular} & \cellcolor{oursrow}\begin{tabular}[c]{@{}c@{}}\diva{}-U\end{tabular} & \cellcolor{oursrow}\begin{tabular}[c]{@{}c@{}}\diva{}-DP\end{tabular} \\
\midrule
\multirow{6}{*}{\textbf{Qwen3}} & 0.6B & 4.38 & 3.49 & 3.84 & 1.17 & 4.30 & 4.38 & 2.41 & 4.38 & 4.16 & \textbf{0.80} & \cellcolor{oursrow}4.06 & \cellcolor{oursrow}\underline{0.81} & \cellcolor{oursrow}4.05 \\
 & 1.7B & 0.99 & 1.05 & \textbf{0.75} & \underline{0.86} & 3.84 & 2.68 & 4.25 & 1.55 & 3.75 & 1.03 & \cellcolor{oursrow}3.23 & \cellcolor{oursrow}0.98 & \cellcolor{oursrow}3.12 \\
 & 4B & 0.85 & 0.78 & \underline{0.74} & 0.79 & 1.01 & 1.91 & 4.13 & 1.56 & 3.02 & 0.76 & \cellcolor{oursrow}2.62 & \cellcolor{oursrow}\textbf{0.70} & \cellcolor{oursrow}2.31 \\
 & 8B & \textbf{0.62} & 0.80 & 0.69 & 0.68 & 1.63 & 3.00 & 3.12 & 1.52 & 3.88 & 0.68 & \cellcolor{oursrow}3.27 & \cellcolor{oursrow}\underline{0.63} & \cellcolor{oursrow}2.55 \\
 & 14B & 1.34 & \underline{0.67} & \textbf{0.65} & 0.79 & 0.80 & 3.98 & 4.24 & 1.42 & 3.15 & 0.69 & \cellcolor{oursrow}2.89 & \cellcolor{oursrow}0.82 & \cellcolor{oursrow}2.30 \\
 & 32B & 1.92 & \textbf{0.70} & \underline{0.73} & 0.94 & 0.86 & 0.95 & 4.24 & 1.55 & 2.56 & 0.85 & \cellcolor{oursrow}2.29 & \cellcolor{oursrow}0.79 & \cellcolor{oursrow}1.92 \\
\cmidrule(lr){1-15}
\multirow{3}{*}{\textbf{Qwen2.5}} & 7B & \textbf{0.66} & 0.75 & 0.79 & 0.98 & 0.83 & 1.61 & 4.38 & 1.48 & 3.05 & 0.76 & \cellcolor{oursrow}2.77 & \cellcolor{oursrow}\underline{0.71} & \cellcolor{oursrow}2.09 \\
 & 14B & 0.83 & 0.92 & \underline{0.69} & 0.93 & 1.01 & 1.26 & 4.24 & 1.18 & 3.05 & \underline{0.69} & \cellcolor{oursrow}3.03 & \cellcolor{oursrow}\textbf{0.63} & \cellcolor{oursrow}2.47 \\
 & 72B & \textbf{0.71} & 0.95 & 0.92 & 0.91 & \underline{0.74} & 1.20 & 4.27 & 1.47 & 2.88 & 0.86 & \cellcolor{oursrow}2.13 & \cellcolor{oursrow}0.82 & \cellcolor{oursrow}1.61 \\
\cmidrule(lr){1-15}
\multirow{2}{*}{\textbf{DeepSeek-R1}} & 7B & 1.33 & 3.71 & 3.21 & 1.13 & 4.38 & 3.24 & 3.93 & 4.18 & 3.08 & \underline{0.85} & \cellcolor{oursrow}1.39 & \cellcolor{oursrow}\textbf{0.78} & \cellcolor{oursrow}1.20 \\
 & 14B & 1.63 & 0.79 & 0.67 & 1.03 & 4.24 & 4.04 & 2.14 & 2.27 & 3.08 & \underline{0.65} & \cellcolor{oursrow}3.64 & \cellcolor{oursrow}\textbf{0.62} & \cellcolor{oursrow}3.50 \\
\cmidrule(lr){1-15}
\multirow{2}{*}{\textbf{Llama}} & 8B & \textbf{0.63} & 1.01 & 0.97 & 1.07 & 1.83 & 2.01 & 2.05 & 1.96 & 2.56 & \underline{0.64} & \cellcolor{oursrow}3.99 & \cellcolor{oursrow}\underline{0.64} & \cellcolor{oursrow}3.58 \\
 & 70B & \underline{0.59} & 0.70 & 0.67 & \textbf{0.57} & 0.78 & 0.84 & 1.56 & 1.40 & 1.97 & 0.98 & \cellcolor{oursrow}1.95 & \cellcolor{oursrow}0.84 & \cellcolor{oursrow}1.27 \\
\cmidrule(lr){1-15}
\multirow{3}{*}{\textbf{Gemma}} & 9B & \textbf{0.49} & 0.86 & 0.79 & 0.91 & 0.73 & 1.20 & 3.49 & 1.40 & 1.03 & 0.70 & \cellcolor{oursrow}3.17 & \cellcolor{oursrow}\underline{0.65} & \cellcolor{oursrow}2.45 \\
 & 12B & \textbf{0.65} & 0.81 & 0.87 & 1.17 & \underline{0.70} & 1.19 & 4.31 & 1.50 & 2.17 & 0.71 & \cellcolor{oursrow}2.69 & \cellcolor{oursrow}0.74 & \cellcolor{oursrow}2.06 \\
 & 27B & 0.69 & \textbf{0.65} & 0.74 & 0.81 & 0.71 & 1.14 & 4.23 & 1.36 & 1.19 & 0.71 & \cellcolor{oursrow}2.51 & \cellcolor{oursrow}\underline{0.66} & \cellcolor{oursrow}2.09 \\
\cmidrule(lr){1-15}
\multirow{2}{*}{\textbf{Mistral}} & 7B & 0.77 & 1.01 & 0.82 & 0.78 & 3.22 & 1.40 & 4.28 & 1.56 & 3.51 & \underline{0.67} & \cellcolor{oursrow}3.60 & \cellcolor{oursrow}\textbf{0.64} & \cellcolor{oursrow}3.07 \\
 & 24B & \underline{0.76} & 0.80 & 0.79 & 0.99 & \underline{0.76} & 1.17 & 4.27 & 1.50 & 2.34 & \textbf{0.68} & \cellcolor{oursrow}3.22 & \cellcolor{oursrow}\textbf{0.68} & \cellcolor{oursrow}2.90 \\
\cmidrule(lr){1-15}
\multirow{1}{*}{\textbf{Phi}} & 14B & \underline{0.59} & 0.65 & \textbf{0.58} & 0.78 & 1.66 & 0.82 & 4.38 & 1.46 & 1.94 & 0.80 & \cellcolor{oursrow}1.78 & \cellcolor{oursrow}0.78 & \cellcolor{oursrow}1.44 \\
\cmidrule(lr){1-15}
\multirow{1}{*}{\textbf{gpt-oss}} & 20B & 1.11 & 1.00 & \underline{0.75} & 0.99 & \textbf{0.73} & 4.21 & 3.36 & 1.45 & 4.35 & 0.84 & \cellcolor{oursrow}1.87 & \cellcolor{oursrow}0.80 & \cellcolor{oursrow}1.50 \\
\midrule
\multicolumn{2}{@{}l|}{Never} & \multicolumn{13}{c@{}}{4.38 (Identical for every LLM)} \\
\bottomrule
}
\caption{\textbf{DP regret per LLM: Vol-20d.} Same first 300 test scenarios as Table~\ref{tab:know}. Baselines are cited in Table~\ref{tab:know}.}
\label{tab:full_know_llm_vol_20}
\end{table*}

%% file: tables/full_ctrl_llm.tex
\begin{table*}[t]
\centering
\small
\setlength{\tabcolsep}{1.8pt}
\filltabular{\textwidth}{15}{>{\ftL}ll|ccccccccccccc<{\ftR}}{
\toprule
Family & Size & \begin{tabular}[c]{@{}c@{}}Direct\end{tabular} & \begin{tabular}[c]{@{}c@{}}CoT\end{tabular} & \begin{tabular}[c]{@{}c@{}}SC\end{tabular} & \begin{tabular}[c]{@{}c@{}}ReAct\end{tabular} & \begin{tabular}[c]{@{}c@{}}Reason.\end{tabular} & \begin{tabular}[c]{@{}c@{}}Ask-P\end{tabular} & \begin{tabular}[c]{@{}c@{}}VoI-P\end{tabular} & \begin{tabular}[c]{@{}c@{}}VoI-Ask\end{tabular} & \begin{tabular}[c]{@{}c@{}}BED-LLM\end{tabular} & \begin{tabular}[c]{@{}c@{}}PAL\end{tabular} & \begin{tabular}[c]{@{}c@{}}Always ask\end{tabular} & \cellcolor{oursrow}\begin{tabular}[c]{@{}c@{}}\diva{}\end{tabular} & \cellcolor{oursrow}\begin{tabular}[c]{@{}c@{}}\diva{}-DP\end{tabular} \\
\midrule
\multirow{6}{*}{\textbf{Qwen3}} & 0.6B & 4.67 & 5.77 & 5.55 & 6.42 & 5.98 & 4.40 & 4.74 & 4.33 & 4.67 & 4.80 & 3.15 & \cellcolor{oursrow}\underline{2.77} & \cellcolor{oursrow}\textbf{2.51} \\
 & 1.7B & 5.99 & 6.36 & 6.06 & 6.58 & 5.86 & 7.08 & 4.39 & 4.39 & 5.25 & 5.03 & 2.92 & \cellcolor{oursrow}\underline{1.46} & \cellcolor{oursrow}\textbf{0.97} \\
 & 4B & 6.00 & 5.52 & 5.50 & 5.64 & 3.65 & 6.04 & 6.22 & 4.24 & 5.39 & 4.49 & 2.58 & \cellcolor{oursrow}\underline{0.86} & \cellcolor{oursrow}\textbf{0.52} \\
 & 8B & 5.76 & 5.37 & 5.29 & 5.72 & 3.02 & 5.64 & 6.28 & 3.90 & 5.45 & 4.72 & 2.45 & \cellcolor{oursrow}\underline{0.66} & \cellcolor{oursrow}\textbf{0.33} \\
 & 14B & 5.57 & 5.35 & 4.98 & 5.16 & 2.24 & 5.53 & 5.28 & 4.11 & 6.57 & 4.08 & 2.44 & \cellcolor{oursrow}\underline{0.57} & \cellcolor{oursrow}\textbf{0.29} \\
 & 32B & 4.48 & 3.97 & 4.08 & 3.74 & 1.04 & 5.06 & 3.69 & 4.08 & 6.40 & 4.35 & 2.47 & \cellcolor{oursrow}\underline{0.60} & \cellcolor{oursrow}\textbf{0.27} \\
\cmidrule(lr){1-15}
\multirow{3}{*}{\textbf{Qwen2.5}} & 7B & 5.85 & 5.90 & 5.95 & 6.33 & -- & 5.40 & 4.72 & 4.29 & 6.06 & 4.77 & 2.64 & \cellcolor{oursrow}\underline{0.81} & \cellcolor{oursrow}\textbf{0.57} \\
 & 14B & 5.14 & 4.98 & 4.80 & 5.15 & -- & 5.47 & 5.07 & 4.30 & 6.73 & 4.72 & 2.54 & \cellcolor{oursrow}\underline{0.66} & \cellcolor{oursrow}\textbf{0.38} \\
 & 72B & 4.39 & 4.19 & 4.06 & 4.38 & -- & 4.84 & 4.60 & 4.24 & 6.59 & 4.36 & 2.41 & \cellcolor{oursrow}\underline{0.59} & \cellcolor{oursrow}\textbf{0.26} \\
\cmidrule(lr){1-15}
\multirow{2}{*}{\textbf{DeepSeek-R1}} & 7B & 5.82 & 5.24 & 5.68 & 6.13 & 4.54 & 4.80 & 4.69 & 3.75 & 4.72 & 4.56 & 2.77 & \cellcolor{oursrow}\underline{1.39} & \cellcolor{oursrow}\textbf{0.92} \\
 & 14B & 5.20 & 4.36 & 4.34 & 5.14 & 3.17 & 5.95 & 4.88 & 4.41 & 5.74 & 4.33 & 2.57 & \cellcolor{oursrow}\underline{0.76} & \cellcolor{oursrow}\textbf{0.45} \\
\cmidrule(lr){1-15}
\multirow{2}{*}{\textbf{Llama}} & 8B & 5.52 & 6.16 & 5.90 & 6.47 & -- & 5.45 & 4.66 & 3.82 & 5.14 & 4.81 & 3.02 & \cellcolor{oursrow}\underline{1.21} & \cellcolor{oursrow}\textbf{0.86} \\
 & 70B & 4.12 & 3.71 & 3.76 & 4.17 & -- & 4.70 & 4.25 & 3.47 & 6.50 & 3.64 & 2.40 & \cellcolor{oursrow}\underline{0.60} & \cellcolor{oursrow}\textbf{0.26} \\
\cmidrule(lr){1-15}
\multirow{3}{*}{\textbf{Gemma}} & 9B & 5.91 & 5.73 & 5.61 & 5.53 & -- & 5.98 & 5.99 & 3.51 & 6.50 & 4.73 & 2.50 & \cellcolor{oursrow}\underline{0.68} & \cellcolor{oursrow}\textbf{0.37} \\
 & 12B & 5.27 & 5.37 & 5.21 & 5.32 & -- & 5.62 & 4.85 & 4.14 & 6.44 & 4.98 & 2.54 & \cellcolor{oursrow}\underline{0.63} & \cellcolor{oursrow}\textbf{0.36} \\
 & 27B & 4.78 & 5.51 & 5.12 & 4.76 & -- & 5.22 & 5.56 & 3.82 & 6.47 & 4.01 & 2.32 & \cellcolor{oursrow}\underline{0.35} & \cellcolor{oursrow}\textbf{0.09} \\
\cmidrule(lr){1-15}
\multirow{2}{*}{\textbf{Mistral}} & 7B & 6.71 & 6.46 & 6.67 & 5.88 & -- & 7.32 & 6.03 & 4.19 & 4.90 & 4.83 & 2.85 & \cellcolor{oursrow}\underline{1.13} & \cellcolor{oursrow}\textbf{0.79} \\
 & 24B & 4.70 & 4.16 & 4.00 & 4.39 & -- & 5.44 & 4.73 & 3.79 & 6.25 & 4.10 & 2.47 & \cellcolor{oursrow}\underline{0.51} & \cellcolor{oursrow}\textbf{0.22} \\
\cmidrule(lr){1-15}
\multirow{1}{*}{\textbf{Phi}} & 14B & 4.83 & 4.80 & 4.09 & 4.60 & -- & 5.40 & 4.89 & 4.48 & 6.46 & 3.39 & 2.38 & \cellcolor{oursrow}\underline{0.55} & \cellcolor{oursrow}\textbf{0.22} \\
\cmidrule(lr){1-15}
\multirow{1}{*}{\textbf{gpt-oss}} & 20B & 4.53 & 4.34 & 3.66 & 3.66 & 1.54 & 4.83 & 4.44 & 3.98 & 5.63 & 3.11 & 2.30 & \cellcolor{oursrow}\underline{0.38} & \cellcolor{oursrow}\textbf{0.10} \\
\midrule
\multicolumn{2}{@{}l|}{Never} & \multicolumn{13}{c@{}}{4.67 (Identical for every LLM)} \\
\bottomrule
}
\caption{\textbf{DP regret per LLM} on controlled environments, averaged over the four domains, first 300 test scenarios per domain. Reason.: Native reasoning (Reasoning models only). Baselines are cited in Table~\ref{tab:know}.}
\label{tab:full_ctrl_llm}
\end{table*}

\begin{table*}[t]
\centering
\small
\setlength{\tabcolsep}{1.8pt}
\filltabular{\textwidth}{15}{>{\ftL}ll|ccccccccccccc<{\ftR}}{
\toprule
Family & Size & \begin{tabular}[c]{@{}c@{}}Direct\end{tabular} & \begin{tabular}[c]{@{}c@{}}CoT\end{tabular} & \begin{tabular}[c]{@{}c@{}}SC\end{tabular} & \begin{tabular}[c]{@{}c@{}}ReAct\end{tabular} & \begin{tabular}[c]{@{}c@{}}Reason.\end{tabular} & \begin{tabular}[c]{@{}c@{}}Ask-P\end{tabular} & \begin{tabular}[c]{@{}c@{}}VoI-P\end{tabular} & \begin{tabular}[c]{@{}c@{}}VoI-Ask\end{tabular} & \begin{tabular}[c]{@{}c@{}}BED-LLM\end{tabular} & \begin{tabular}[c]{@{}c@{}}PAL\end{tabular} & \begin{tabular}[c]{@{}c@{}}Always ask\end{tabular} & \cellcolor{oursrow}\begin{tabular}[c]{@{}c@{}}\diva{}\end{tabular} & \cellcolor{oursrow}\begin{tabular}[c]{@{}c@{}}\diva{}-DP\end{tabular} \\
\midrule
\multirow{6}{*}{\textbf{Qwen3}} & 0.6B & 4.63 & 5.99 & 5.14 & 6.71 & 6.02 & 4.63 & 4.80 & 4.60 & 4.63 & 5.00 & \textbf{3.08} & \cellcolor{oursrow}3.36 & \cellcolor{oursrow}\underline{3.29} \\
 & 1.7B & 5.36 & 5.81 & 5.41 & 6.73 & 5.25 & 7.49 & 4.27 & 4.60 & 5.23 & 5.49 & 2.92 & \cellcolor{oursrow}\underline{1.53} & \cellcolor{oursrow}\textbf{1.08} \\
 & 4B & 5.59 & 5.12 & 5.27 & 5.54 & 3.03 & 6.02 & 5.92 & 4.54 & 5.50 & 4.76 & 2.41 & \cellcolor{oursrow}\underline{0.85} & \cellcolor{oursrow}\textbf{0.54} \\
 & 8B & 5.22 & 4.47 & 4.90 & 5.33 & 2.68 & 4.98 & 5.47 & 4.60 & 5.21 & 4.13 & 2.30 & \cellcolor{oursrow}\underline{0.80} & \cellcolor{oursrow}\textbf{0.51} \\
 & 14B & 4.50 & 5.25 & 4.89 & 4.72 & 2.45 & 4.45 & 4.88 & 4.50 & 5.92 & 4.38 & 2.18 & \cellcolor{oursrow}\underline{0.61} & \cellcolor{oursrow}\textbf{0.41} \\
 & 32B & 3.57 & 3.82 & 3.63 & 3.58 & 0.84 & 4.26 & 4.06 & 4.47 & 5.99 & 4.10 & 2.21 & \cellcolor{oursrow}\underline{0.41} & \cellcolor{oursrow}\textbf{0.24} \\
\cmidrule(lr){1-15}
\multirow{3}{*}{\textbf{Qwen2.5}} & 7B & 5.39 & 5.16 & 5.66 & 6.35 & -- & 5.03 & 4.39 & 4.43 & 4.92 & 4.69 & 2.43 & \cellcolor{oursrow}\underline{0.69} & \cellcolor{oursrow}\textbf{0.49} \\
 & 14B & 4.39 & 4.39 & 3.71 & 4.87 & -- & 5.00 & 4.82 & 4.61 & 6.24 & 4.73 & 2.26 & \cellcolor{oursrow}\underline{0.66} & \cellcolor{oursrow}\textbf{0.37} \\
 & 72B & 3.19 & 3.75 & 3.13 & 3.66 & -- & 3.96 & 4.63 & 4.47 & 6.40 & 4.21 & 2.25 & \cellcolor{oursrow}\underline{0.72} & \cellcolor{oursrow}\textbf{0.35} \\
\cmidrule(lr){1-15}
\multirow{2}{*}{\textbf{DeepSeek-R1}} & 7B & 5.79 & 4.51 & 4.59 & 5.96 & 4.26 & 4.78 & 4.21 & 4.50 & 4.63 & 4.28 & 2.40 & \cellcolor{oursrow}\underline{0.87} & \cellcolor{oursrow}\textbf{0.54} \\
 & 14B & 4.40 & 3.85 & 3.45 & 5.05 & 3.00 & 5.33 & 4.67 & 4.65 & 5.04 & 3.68 & 2.30 & \cellcolor{oursrow}\underline{0.70} & \cellcolor{oursrow}\textbf{0.45} \\
\cmidrule(lr){1-15}
\multirow{2}{*}{\textbf{Llama}} & 8B & 4.62 & 5.75 & 5.30 & 6.36 & -- & 4.63 & 4.48 & 4.28 & 4.63 & 4.80 & 2.95 & \cellcolor{oursrow}\underline{1.50} & \cellcolor{oursrow}\textbf{1.12} \\
 & 70B & 3.65 & 2.98 & 3.24 & 3.77 & -- & 4.65 & 4.12 & 3.73 & 5.50 & 3.91 & 2.12 & \cellcolor{oursrow}\underline{0.58} & \cellcolor{oursrow}\textbf{0.30} \\
\cmidrule(lr){1-15}
\multirow{3}{*}{\textbf{Gemma}} & 9B & 5.79 & 4.77 & 4.92 & 5.80 & -- & 5.74 & 5.22 & 4.14 & 5.97 & 4.85 & 2.32 & \cellcolor{oursrow}\underline{0.55} & \cellcolor{oursrow}\textbf{0.31} \\
 & 12B & 4.64 & 4.35 & 4.40 & 5.47 & -- & 5.12 & 4.26 & 4.16 & 5.93 & 4.63 & 2.44 & \cellcolor{oursrow}\underline{0.53} & \cellcolor{oursrow}\textbf{0.35} \\
 & 27B & 4.16 & 4.77 & 3.78 & 4.51 & -- & 4.67 & 4.46 & 4.02 & 6.29 & 3.85 & 2.10 & \cellcolor{oursrow}\underline{0.28} & \cellcolor{oursrow}\textbf{0.08} \\
\cmidrule(lr){1-15}
\multirow{2}{*}{\textbf{Mistral}} & 7B & 6.86 & 6.76 & 6.37 & 5.40 & -- & 6.96 & 4.99 & 4.51 & 4.82 & 4.79 & 2.66 & \cellcolor{oursrow}\underline{1.27} & \cellcolor{oursrow}\textbf{0.93} \\
 & 24B & 4.19 & 3.68 & 3.48 & 3.81 & -- & 4.98 & 4.93 & 4.18 & 6.65 & 4.23 & 2.22 & \cellcolor{oursrow}\underline{0.50} & \cellcolor{oursrow}\textbf{0.27} \\
\cmidrule(lr){1-15}
\multirow{1}{*}{\textbf{Phi}} & 14B & 3.51 & 4.74 & 3.79 & 3.87 & -- & 4.69 & 4.86 & 4.58 & 5.85 & 3.35 & 2.09 & \cellcolor{oursrow}\underline{0.45} & \cellcolor{oursrow}\textbf{0.21} \\
\cmidrule(lr){1-15}
\multirow{1}{*}{\textbf{gpt-oss}} & 20B & 4.26 & 4.30 & 3.34 & 3.29 & 1.33 & 4.34 & 4.64 & 4.29 & 4.63 & 3.24 & 2.03 & \cellcolor{oursrow}\underline{0.40} & \cellcolor{oursrow}\textbf{0.19} \\
\midrule
\multicolumn{2}{@{}l|}{Never} & \multicolumn{13}{c@{}}{4.63 (Identical for every LLM)} \\
\bottomrule
}
\caption{\textbf{DP regret per LLM: Finance.} Controlled, first 300 test scenarios. Baselines are cited in Table~\ref{tab:know}.}
\label{tab:full_ctrl_llm_finance}
\end{table*}

\begin{table*}[t]
\centering
\small
\setlength{\tabcolsep}{1.8pt}
\filltabular{\textwidth}{15}{>{\ftL}ll|ccccccccccccc<{\ftR}}{
\toprule
Family & Size & \begin{tabular}[c]{@{}c@{}}Direct\end{tabular} & \begin{tabular}[c]{@{}c@{}}CoT\end{tabular} & \begin{tabular}[c]{@{}c@{}}SC\end{tabular} & \begin{tabular}[c]{@{}c@{}}ReAct\end{tabular} & \begin{tabular}[c]{@{}c@{}}Reason.\end{tabular} & \begin{tabular}[c]{@{}c@{}}Ask-P\end{tabular} & \begin{tabular}[c]{@{}c@{}}VoI-P\end{tabular} & \begin{tabular}[c]{@{}c@{}}VoI-Ask\end{tabular} & \begin{tabular}[c]{@{}c@{}}BED-LLM\end{tabular} & \begin{tabular}[c]{@{}c@{}}PAL\end{tabular} & \begin{tabular}[c]{@{}c@{}}Always ask\end{tabular} & \cellcolor{oursrow}\begin{tabular}[c]{@{}c@{}}\diva{}\end{tabular} & \cellcolor{oursrow}\begin{tabular}[c]{@{}c@{}}\diva{}-DP\end{tabular} \\
\midrule
\multirow{6}{*}{\textbf{Qwen3}} & 0.6B & 4.54 & 4.84 & 4.93 & 5.21 & 5.43 & 3.48 & 4.39 & 3.58 & 4.54 & 4.46 & \textbf{2.48} & \cellcolor{oursrow}2.92 & \cellcolor{oursrow}\underline{2.65} \\
 & 1.7B & 4.93 & 5.24 & 5.92 & 5.04 & 4.97 & 5.95 & 4.50 & 4.24 & 4.64 & 4.53 & 2.33 & \cellcolor{oursrow}\underline{1.80} & \cellcolor{oursrow}\textbf{1.24} \\
 & 4B & 4.95 & 4.29 & 4.22 & 4.58 & 3.43 & 4.90 & 4.89 & 4.24 & 4.10 & 3.82 & 1.88 & \cellcolor{oursrow}\underline{0.90} & \cellcolor{oursrow}\textbf{0.57} \\
 & 8B & 4.64 & 4.56 & 3.98 & 4.47 & 2.66 & 4.62 & 4.66 & 3.88 & 4.81 & 3.84 & 1.76 & \cellcolor{oursrow}\underline{0.69} & \cellcolor{oursrow}\textbf{0.31} \\
 & 14B & 4.33 & 3.92 & 4.10 & 4.23 & 2.13 & 4.38 & 4.47 & 3.97 & 5.45 & 3.71 & 1.88 & \cellcolor{oursrow}\underline{0.75} & \cellcolor{oursrow}\textbf{0.32} \\
 & 32B & 3.52 & 3.72 & 3.49 & 2.66 & 0.96 & 3.98 & 3.12 & 4.35 & 5.21 & 4.35 & 1.95 & \cellcolor{oursrow}\underline{0.91} & \cellcolor{oursrow}\textbf{0.35} \\
\cmidrule(lr){1-15}
\multirow{3}{*}{\textbf{Qwen2.5}} & 7B & 4.18 & 4.24 & 4.35 & 5.01 & -- & 4.28 & 4.41 & 4.15 & 4.90 & 3.84 & 2.13 & \cellcolor{oursrow}\underline{0.73} & \cellcolor{oursrow}\textbf{0.50} \\
 & 14B & 4.01 & 3.61 & 3.78 & 4.19 & -- & 4.26 & 4.34 & 4.36 & 4.90 & 4.65 & 1.94 & \cellcolor{oursrow}\underline{0.74} & \cellcolor{oursrow}\textbf{0.46} \\
 & 72B & 3.47 & 3.42 & 3.42 & 3.31 & -- & 3.76 & 4.54 & 4.30 & 5.26 & 3.94 & 1.76 & \cellcolor{oursrow}\underline{0.54} & \cellcolor{oursrow}\textbf{0.19} \\
\cmidrule(lr){1-15}
\multirow{2}{*}{\textbf{DeepSeek-R1}} & 7B & 4.48 & 5.06 & 5.42 & 5.13 & 3.60 & 4.54 & 4.92 & 3.47 & 4.54 & 4.42 & 2.27 & \cellcolor{oursrow}\underline{1.61} & \cellcolor{oursrow}\textbf{1.09} \\
 & 14B & 5.02 & 3.44 & 3.42 & 3.99 & 2.31 & 5.35 & 4.19 & 4.61 & 4.89 & 4.05 & 2.09 & \cellcolor{oursrow}\underline{0.90} & \cellcolor{oursrow}\textbf{0.53} \\
\cmidrule(lr){1-15}
\multirow{2}{*}{\textbf{Llama}} & 8B & 4.28 & 4.83 & 5.00 & 5.37 & -- & 4.58 & 4.42 & 3.48 & 4.54 & 4.34 & 2.63 & \cellcolor{oursrow}\underline{1.40} & \cellcolor{oursrow}\textbf{1.04} \\
 & 70B & 3.40 & 2.99 & 3.16 & 3.67 & -- & 3.58 & 3.61 & 3.81 & 5.49 & 3.02 & 1.82 & \cellcolor{oursrow}\underline{0.69} & \cellcolor{oursrow}\textbf{0.30} \\
\cmidrule(lr){1-15}
\multirow{3}{*}{\textbf{Gemma}} & 9B & 4.96 & 4.46 & 4.21 & 4.50 & -- & 4.64 & 4.82 & 3.58 & 5.57 & 3.63 & 1.83 & \cellcolor{oursrow}\underline{0.83} & \cellcolor{oursrow}\textbf{0.48} \\
 & 12B & 4.18 & 3.98 & 3.81 & 4.57 & -- & 4.39 & 3.83 & 4.02 & 5.16 & 4.35 & 1.81 & \cellcolor{oursrow}\underline{0.66} & \cellcolor{oursrow}\textbf{0.30} \\
 & 27B & 4.04 & 4.08 & 4.21 & 4.01 & -- & 4.24 & 4.16 & 3.92 & 5.08 & 4.15 & 1.66 & \cellcolor{oursrow}\underline{0.43} & \cellcolor{oursrow}\textbf{0.07} \\
\cmidrule(lr){1-15}
\multirow{2}{*}{\textbf{Mistral}} & 7B & 5.94 & 6.00 & 6.06 & 4.38 & -- & 5.93 & 4.87 & 4.15 & 4.76 & 4.77 & 2.25 & \cellcolor{oursrow}\underline{1.26} & \cellcolor{oursrow}\textbf{0.78} \\
 & 24B & 3.84 & 3.39 & 3.10 & 3.24 & -- & 4.26 & 4.85 & 3.99 & 4.97 & 3.25 & 1.82 & \cellcolor{oursrow}\underline{0.57} & \cellcolor{oursrow}\textbf{0.18} \\
\cmidrule(lr){1-15}
\multirow{1}{*}{\textbf{Phi}} & 14B & 3.80 & 4.27 & 3.49 & 3.19 & -- & 4.16 & 4.49 & 4.41 & 5.23 & 3.18 & 1.82 & \cellcolor{oursrow}\underline{0.53} & \cellcolor{oursrow}\textbf{0.15} \\
\cmidrule(lr){1-15}
\multirow{1}{*}{\textbf{gpt-oss}} & 20B & 4.45 & 3.74 & 3.26 & 2.92 & 1.05 & 3.92 & 4.19 & 4.02 & 4.70 & 3.21 & 1.72 & \cellcolor{oursrow}\underline{0.45} & \cellcolor{oursrow}\textbf{0.12} \\
\midrule
\multicolumn{2}{@{}l|}{Never} & \multicolumn{13}{c@{}}{4.54 (Identical for every LLM)} \\
\bottomrule
}
\caption{\textbf{DP regret per LLM: Credit.} Controlled, first 300 test scenarios. Baselines are cited in Table~\ref{tab:know}.}
\label{tab:full_ctrl_llm_credit}
\end{table*}

\begin{table*}[t]
\centering
\small
\setlength{\tabcolsep}{1.8pt}
\filltabular{\textwidth}{15}{>{\ftL}ll|ccccccccccccc<{\ftR}}{
\toprule
Family & Size & \begin{tabular}[c]{@{}c@{}}Direct\end{tabular} & \begin{tabular}[c]{@{}c@{}}CoT\end{tabular} & \begin{tabular}[c]{@{}c@{}}SC\end{tabular} & \begin{tabular}[c]{@{}c@{}}ReAct\end{tabular} & \begin{tabular}[c]{@{}c@{}}Reason.\end{tabular} & \begin{tabular}[c]{@{}c@{}}Ask-P\end{tabular} & \begin{tabular}[c]{@{}c@{}}VoI-P\end{tabular} & \begin{tabular}[c]{@{}c@{}}VoI-Ask\end{tabular} & \begin{tabular}[c]{@{}c@{}}BED-LLM\end{tabular} & \begin{tabular}[c]{@{}c@{}}PAL\end{tabular} & \begin{tabular}[c]{@{}c@{}}Always ask\end{tabular} & \cellcolor{oursrow}\begin{tabular}[c]{@{}c@{}}\diva{}\end{tabular} & \cellcolor{oursrow}\begin{tabular}[c]{@{}c@{}}\diva{}-DP\end{tabular} \\
\midrule
\multirow{6}{*}{\textbf{Qwen3}} & 0.6B & 4.52 & 7.01 & 6.63 & 6.51 & 6.00 & 4.52 & 4.77 & 4.27 & 4.52 & 4.72 & 3.48 & \cellcolor{oursrow}\underline{1.74} & \cellcolor{oursrow}\textbf{1.25} \\
 & 1.7B & 6.84 & 6.84 & 6.89 & 7.10 & 6.39 & 9.01 & 3.80 & 4.13 & 6.56 & 4.82 & 3.38 & \cellcolor{oursrow}\underline{1.13} & \cellcolor{oursrow}\textbf{0.71} \\
 & 4B & 7.19 & 5.73 & 6.02 & 6.33 & 2.77 & 7.22 & 7.11 & 4.15 & 6.95 & 4.71 & 3.21 & \cellcolor{oursrow}\underline{0.94} & \cellcolor{oursrow}\textbf{0.53} \\
 & 8B & 6.54 & 5.75 & 5.70 & 6.30 & 2.92 & 5.22 & 7.28 & 2.65 & 6.63 & 4.82 & 2.90 & \cellcolor{oursrow}\underline{0.58} & \cellcolor{oursrow}\textbf{0.20} \\
 & 14B & 6.93 & 6.22 & 5.48 & 6.11 & 1.83 & 6.87 & 5.46 & 3.25 & 8.03 & 3.75 & 2.98 & \cellcolor{oursrow}\underline{0.45} & \cellcolor{oursrow}\textbf{0.22} \\
 & 32B & 6.07 & 4.27 & 5.07 & 4.16 & 0.78 & 6.01 & 3.74 & 2.89 & 7.45 & 3.57 & 3.02 & \cellcolor{oursrow}\underline{0.57} & \cellcolor{oursrow}\textbf{0.26} \\
\cmidrule(lr){1-15}
\multirow{3}{*}{\textbf{Qwen2.5}} & 7B & 6.46 & 6.84 & 6.92 & 7.23 & -- & 6.59 & 4.85 & 4.26 & 7.67 & 5.22 & 3.23 & \cellcolor{oursrow}\underline{0.88} & \cellcolor{oursrow}\textbf{0.61} \\
 & 14B & 6.46 & 5.59 & 6.18 & 5.75 & -- & 6.65 & 5.30 & 3.53 & 7.88 & 4.50 & 3.09 & \cellcolor{oursrow}\underline{0.60} & \cellcolor{oursrow}\textbf{0.31} \\
 & 72B & 6.90 & 5.01 & 5.25 & 5.61 & -- & 6.53 & 4.52 & 3.41 & 7.96 & 4.62 & 2.97 & \cellcolor{oursrow}\underline{0.57} & \cellcolor{oursrow}\textbf{0.27} \\
\cmidrule(lr){1-15}
\multirow{2}{*}{\textbf{DeepSeek-R1}} & 7B & 6.51 & 5.22 & 6.53 & 6.99 & 5.05 & 5.22 & 4.58 & 3.83 & 4.72 & 4.83 & 3.31 & \cellcolor{oursrow}\underline{1.48} & \cellcolor{oursrow}\textbf{1.04} \\
 & 14B & 5.67 & 5.31 & 5.52 & 6.00 & 3.75 & 6.46 & 4.88 & 3.70 & 7.97 & 4.91 & 3.09 & \cellcolor{oursrow}\underline{0.65} & \cellcolor{oursrow}\textbf{0.34} \\
\cmidrule(lr){1-15}
\multirow{2}{*}{\textbf{Llama}} & 8B & 6.70 & 6.51 & 6.34 & 7.17 & -- & 6.10 & 4.49 & 3.27 & 6.40 & 4.47 & 3.22 & \cellcolor{oursrow}\underline{0.72} & \cellcolor{oursrow}\textbf{0.48} \\
 & 70B & 4.28 & 4.17 & 4.32 & 4.56 & -- & 5.17 & 5.18 & 2.31 & 7.52 & 3.56 & 3.00 & \cellcolor{oursrow}\underline{0.67} & \cellcolor{oursrow}\textbf{0.27} \\
\cmidrule(lr){1-15}
\multirow{3}{*}{\textbf{Gemma}} & 9B & 6.64 & 6.87 & 6.59 & 6.25 & -- & 7.12 & 6.98 & 1.93 & 7.57 & 5.10 & 3.08 & \cellcolor{oursrow}\underline{0.73} & \cellcolor{oursrow}\textbf{0.36} \\
 & 12B & 5.93 & 6.27 & 6.01 & 5.15 & -- & 6.11 & 5.82 & 3.49 & 7.86 & 5.53 & 3.09 & \cellcolor{oursrow}\underline{0.58} & \cellcolor{oursrow}\textbf{0.29} \\
 & 27B & 5.49 & 6.51 & 6.20 & 5.51 & -- & 6.11 & 6.67 & 3.44 & 7.77 & 3.84 & 2.89 & \cellcolor{oursrow}\underline{0.33} & \cellcolor{oursrow}\textbf{0.07} \\
\cmidrule(lr){1-15}
\multirow{2}{*}{\textbf{Mistral}} & 7B & 8.15 & 7.24 & 7.85 & 6.95 & -- & 8.45 & 7.39 & 3.46 & 4.71 & 4.81 & 3.35 & \cellcolor{oursrow}\underline{0.88} & \cellcolor{oursrow}\textbf{0.69} \\
 & 24B & 5.90 & 4.67 & 5.19 & 5.38 & -- & 6.18 & 4.32 & 2.88 & 7.90 & 3.47 & 2.98 & \cellcolor{oursrow}\underline{0.47} & \cellcolor{oursrow}\textbf{0.16} \\
\cmidrule(lr){1-15}
\multirow{1}{*}{\textbf{Phi}} & 14B & 6.63 & 5.38 & 4.67 & 6.38 & -- & 6.78 & 5.07 & 4.25 & 7.28 & 3.42 & 2.89 & \cellcolor{oursrow}\underline{0.39} & \cellcolor{oursrow}\textbf{0.06} \\
\cmidrule(lr){1-15}
\multirow{1}{*}{\textbf{gpt-oss}} & 20B & 4.52 & 4.09 & 3.47 & 4.30 & 1.30 & 5.27 & 4.52 & 2.83 & 7.98 & 2.11 & 2.87 & \cellcolor{oursrow}\underline{0.33} & \cellcolor{oursrow}\textbf{0.05} \\
\midrule
\multicolumn{2}{@{}l|}{Never} & \multicolumn{13}{c@{}}{4.52 (Identical for every LLM)} \\
\bottomrule
}
\caption{\textbf{DP regret per LLM: Clinical.} Controlled, first 300 test scenarios. Baselines are cited in Table~\ref{tab:know}.}
\label{tab:full_ctrl_llm_clinical}
\end{table*}

\begin{table*}[t]
\centering
\small
\setlength{\tabcolsep}{1.8pt}
\filltabular{\textwidth}{15}{>{\ftL}ll|ccccccccccccc<{\ftR}}{
\toprule
Family & Size & \begin{tabular}[c]{@{}c@{}}Direct\end{tabular} & \begin{tabular}[c]{@{}c@{}}CoT\end{tabular} & \begin{tabular}[c]{@{}c@{}}SC\end{tabular} & \begin{tabular}[c]{@{}c@{}}ReAct\end{tabular} & \begin{tabular}[c]{@{}c@{}}Reason.\end{tabular} & \begin{tabular}[c]{@{}c@{}}Ask-P\end{tabular} & \begin{tabular}[c]{@{}c@{}}VoI-P\end{tabular} & \begin{tabular}[c]{@{}c@{}}VoI-Ask\end{tabular} & \begin{tabular}[c]{@{}c@{}}BED-LLM\end{tabular} & \begin{tabular}[c]{@{}c@{}}PAL\end{tabular} & \begin{tabular}[c]{@{}c@{}}Always ask\end{tabular} & \cellcolor{oursrow}\begin{tabular}[c]{@{}c@{}}\diva{}\end{tabular} & \cellcolor{oursrow}\begin{tabular}[c]{@{}c@{}}\diva{}-DP\end{tabular} \\
\midrule
\multirow{6}{*}{\textbf{Qwen3}} & 0.6B & 4.98 & 5.22 & 5.51 & 7.25 & 6.48 & 4.98 & 5.00 & 4.87 & 4.98 & 5.01 & 3.56 & \cellcolor{oursrow}\underline{3.05} & \cellcolor{oursrow}\textbf{2.86} \\
 & 1.7B & 6.83 & 7.55 & 6.03 & 7.45 & 6.82 & 5.86 & 4.99 & 4.58 & 4.59 & 5.26 & 3.04 & \cellcolor{oursrow}\underline{1.38} & \cellcolor{oursrow}\textbf{0.84} \\
 & 4B & 6.28 & 6.91 & 6.50 & 6.11 & 5.35 & 6.04 & 6.94 & 4.03 & 5.03 & 4.69 & 2.82 & \cellcolor{oursrow}\underline{0.75} & \cellcolor{oursrow}\textbf{0.43} \\
 & 8B & 6.65 & 6.68 & 6.58 & 6.80 & 3.83 & 7.72 & 7.73 & 4.46 & 5.15 & 6.10 & 2.83 & \cellcolor{oursrow}\underline{0.58} & \cellcolor{oursrow}\textbf{0.30} \\
 & 14B & 6.52 & 6.03 & 5.46 & 5.57 & 2.54 & 6.43 & 6.31 & 4.71 & 6.88 & 4.47 & 2.72 & \cellcolor{oursrow}\underline{0.47} & \cellcolor{oursrow}\textbf{0.22} \\
 & 32B & 4.76 & 4.09 & 4.15 & 4.54 & 1.59 & 5.97 & 3.85 & 4.63 & 6.96 & 5.38 & 2.70 & \cellcolor{oursrow}\underline{0.49} & \cellcolor{oursrow}\textbf{0.25} \\
\cmidrule(lr){1-15}
\multirow{3}{*}{\textbf{Qwen2.5}} & 7B & 7.36 & 7.36 & 6.86 & 6.71 & -- & 5.72 & 5.24 & 4.32 & 6.74 & 5.34 & 2.78 & \cellcolor{oursrow}\underline{0.95} & \cellcolor{oursrow}\textbf{0.67} \\
 & 14B & 5.71 & 6.35 & 5.51 & 5.78 & -- & 5.98 & 5.81 & 4.71 & 7.90 & 5.01 & 2.86 & \cellcolor{oursrow}\underline{0.64} & \cellcolor{oursrow}\textbf{0.39} \\
 & 72B & 4.01 & 4.57 & 4.45 & 4.95 & -- & 5.11 & 4.71 & 4.77 & 6.74 & 4.66 & 2.68 & \cellcolor{oursrow}\underline{0.54} & \cellcolor{oursrow}\textbf{0.23} \\
\cmidrule(lr){1-15}
\multirow{2}{*}{\textbf{DeepSeek-R1}} & 7B & 6.51 & 6.19 & 6.17 & 6.46 & 5.28 & 4.67 & 5.06 & 3.21 & 4.98 & 4.71 & 3.11 & \cellcolor{oursrow}\underline{1.59} & \cellcolor{oursrow}\textbf{1.00} \\
 & 14B & 5.70 & 4.84 & 4.99 & 5.50 & 3.60 & 6.66 & 5.79 & 4.68 & 5.06 & 4.70 & 2.82 & \cellcolor{oursrow}\underline{0.79} & \cellcolor{oursrow}\textbf{0.50} \\
\cmidrule(lr){1-15}
\multirow{2}{*}{\textbf{Llama}} & 8B & 6.48 & 7.55 & 6.97 & 6.99 & -- & 6.49 & 5.23 & 4.26 & 4.98 & 5.64 & 3.28 & \cellcolor{oursrow}\underline{1.20} & \cellcolor{oursrow}\textbf{0.81} \\
 & 70B & 5.15 & 4.69 & 4.34 & 4.67 & -- & 5.40 & 4.10 & 4.05 & 7.51 & 4.07 & 2.67 & \cellcolor{oursrow}\underline{0.46} & \cellcolor{oursrow}\textbf{0.17} \\
\cmidrule(lr){1-15}
\multirow{3}{*}{\textbf{Gemma}} & 9B & 6.26 & 6.81 & 6.72 & 5.55 & -- & 6.40 & 6.95 & 4.41 & 6.87 & 5.32 & 2.77 & \cellcolor{oursrow}\underline{0.58} & \cellcolor{oursrow}\textbf{0.35} \\
 & 12B & 6.35 & 6.88 & 6.64 & 6.08 & -- & 6.85 & 5.51 & 4.88 & 6.82 & 5.42 & 2.82 & \cellcolor{oursrow}\underline{0.76} & \cellcolor{oursrow}\textbf{0.50} \\
 & 27B & 5.44 & 6.69 & 6.31 & 5.01 & -- & 5.87 & 6.94 & 3.89 & 6.76 & 4.18 & 2.64 & \cellcolor{oursrow}\underline{0.35} & \cellcolor{oursrow}\textbf{0.13} \\
\cmidrule(lr){1-15}
\multirow{2}{*}{\textbf{Mistral}} & 7B & 5.91 & 5.86 & 6.39 & 6.78 & -- & 7.94 & 6.87 & 4.64 & 5.32 & 4.96 & 3.12 & \cellcolor{oursrow}\underline{1.12} & \cellcolor{oursrow}\textbf{0.76} \\
 & 24B & 4.88 & 4.92 & 4.26 & 5.13 & -- & 6.36 & 4.81 & 4.12 & 5.48 & 5.45 & 2.86 & \cellcolor{oursrow}\underline{0.52} & \cellcolor{oursrow}\textbf{0.27} \\
\cmidrule(lr){1-15}
\multirow{1}{*}{\textbf{Phi}} & 14B & 5.38 & 4.79 & 4.42 & 4.95 & -- & 5.97 & 5.14 & 4.68 & 7.47 & 3.62 & 2.73 & \cellcolor{oursrow}\underline{0.82} & \cellcolor{oursrow}\textbf{0.47} \\
\cmidrule(lr){1-15}
\multirow{1}{*}{\textbf{gpt-oss}} & 20B & 4.90 & 5.26 & 4.57 & 4.13 & 2.49 & 5.79 & 4.40 & 4.79 & 5.22 & 3.90 & 2.60 & \cellcolor{oursrow}\underline{0.33} & \cellcolor{oursrow}\textbf{0.05} \\
\midrule
\multicolumn{2}{@{}l|}{Never} & \multicolumn{13}{c@{}}{4.98 (Identical for every LLM)} \\
\bottomrule
}
\caption{\textbf{DP regret per LLM: Legal.} Controlled, first 300 test scenarios. Baselines are cited in Table~\ref{tab:know}.}
\label{tab:full_ctrl_llm_legal}
\end{table*}

%% file: tables/full_episodes_llm.tex
\begin{table*}[t]
\centering
\small
\setlength{\tabcolsep}{1.8pt}
\filltabular{\textwidth}{15}{>{\ftL}ll|ccccccccccccc<{\ftR}}{
\toprule
Family & Size & \begin{tabular}[c]{@{}c@{}}Direct\end{tabular} & \begin{tabular}[c]{@{}c@{}}CoT\end{tabular} & \begin{tabular}[c]{@{}c@{}}SC\end{tabular} & \begin{tabular}[c]{@{}c@{}}ReAct\end{tabular} & \begin{tabular}[c]{@{}c@{}}Ask-P\end{tabular} & \begin{tabular}[c]{@{}c@{}}VoI-P\end{tabular} & \begin{tabular}[c]{@{}c@{}}VoI-Ask\end{tabular} & \begin{tabular}[c]{@{}c@{}}BED-LLM\end{tabular} & \begin{tabular}[c]{@{}c@{}}PAL\end{tabular} & \begin{tabular}[c]{@{}c@{}}Never\end{tabular} & \begin{tabular}[c]{@{}c@{}}Always ask\end{tabular} & \cellcolor{oursrow}\begin{tabular}[c]{@{}c@{}}\diva{}\end{tabular} & \cellcolor{oursrow}\begin{tabular}[c]{@{}c@{}}\diva{}-DP\end{tabular} \\
\midrule
\multirow{6}{*}{\textbf{Qwen3}} & 0.6B & 15.43 & 18.20 & 16.94 & 38.96 & 15.52 & 16.65 & 15.43 & 15.43 & 15.91 & 15.40 & 40.41 & \cellcolor{oursrow}\underline{5.62} & \cellcolor{oursrow}\textbf{5.46} \\
 & 1.7B & 19.28 & 25.68 & 29.61 & 36.04 & 23.52 & 15.34 & 14.78 & 17.43 & 17.41 & 14.46 & 41.55 & \cellcolor{oursrow}\underline{5.17} & \cellcolor{oursrow}\textbf{5.13} \\
 & 4B & 34.52 & 31.76 & 34.15 & 29.84 & 26.07 & 30.88 & 12.02 & 14.41 & 13.22 & 11.66 & 35.67 & \cellcolor{oursrow}\underline{1.64} & \cellcolor{oursrow}\textbf{1.44} \\
 & 8B & 32.44 & 27.49 & 30.07 & 28.77 & 18.39 & 31.25 & 11.96 & 13.22 & 13.03 & 11.53 & 35.12 & \cellcolor{oursrow}\underline{1.19} & \cellcolor{oursrow}\textbf{0.88} \\
 & 14B & 31.05 & 29.44 & 30.95 & 27.13 & 33.44 & 13.18 & 10.44 & 27.25 & 12.05 & 10.38 & 35.43 & \cellcolor{oursrow}\underline{0.92} & \cellcolor{oursrow}\textbf{0.70} \\
 & 32B & 19.55 & 24.78 & 27.48 & 22.82 & 32.34 & 7.44 & 6.99 & 27.28 & 6.94 & 6.17 & 35.91 & \cellcolor{oursrow}\underline{0.96} & \cellcolor{oursrow}\textbf{0.62} \\
\cmidrule(lr){1-15}
\multirow{3}{*}{\textbf{Qwen2.5}} & 7B & 33.48 & 33.41 & 34.33 & 22.45 & 32.61 & 11.59 & 12.33 & 19.81 & 15.90 & 11.62 & 35.63 & \cellcolor{oursrow}\underline{1.58} & \cellcolor{oursrow}\textbf{1.42} \\
 & 14B & 36.36 & 34.47 & 35.99 & 17.33 & 36.57 & 13.76 & 11.18 & 29.28 & 14.78 & 10.83 & 37.29 & \cellcolor{oursrow}\underline{1.18} & \cellcolor{oursrow}\textbf{0.90} \\
 & 72B & 19.46 & 29.64 & 30.60 & 16.29 & 21.98 & 9.40 & 9.57 & 28.67 & 11.02 & 9.40 & 35.34 & \cellcolor{oursrow}\underline{1.31} & \cellcolor{oursrow}\textbf{0.96} \\
\cmidrule(lr){1-15}
\multirow{2}{*}{\textbf{DeepSeek-R1}} & 7B & 32.53 & 18.01 & 22.29 & 31.50 & 14.63 & 14.66 & 12.78 & 12.45 & 15.95 & 12.54 & 40.91 & \cellcolor{oursrow}\underline{2.18} & \cellcolor{oursrow}\textbf{2.00} \\
 & 14B & 21.79 & 32.97 & 34.73 & 28.72 & 26.28 & 15.80 & 10.52 & 16.00 & 15.10 & 10.00 & 36.31 & \cellcolor{oursrow}\textbf{1.31} & \cellcolor{oursrow}\underline{1.37} \\
\cmidrule(lr){1-15}
\multirow{2}{*}{\textbf{Llama}} & 8B & 31.04 & 35.04 & 34.53 & 27.97 & 29.00 & 13.00 & 12.07 & 12.94 & 16.55 & 11.76 & 34.91 & \cellcolor{oursrow}\underline{3.38} & \cellcolor{oursrow}\textbf{3.31} \\
 & 70B & 34.43 & 29.23 & 30.05 & 13.64 & 33.27 & 18.29 & 8.37 & 21.86 & 11.44 & 8.22 & 35.40 & \cellcolor{oursrow}\underline{1.03} & \cellcolor{oursrow}\textbf{0.73} \\
\cmidrule(lr){1-15}
\multirow{3}{*}{\textbf{Gemma}} & 9B & 34.12 & 29.36 & 33.05 & 19.58 & 34.56 & 30.33 & 9.38 & 24.27 & 19.12 & 9.77 & 34.33 & \cellcolor{oursrow}\underline{1.56} & \cellcolor{oursrow}\textbf{1.23} \\
 & 12B & 33.80 & 34.58 & 35.25 & 25.05 & 35.33 & 14.15 & 10.59 & 19.32 & 12.26 & 10.50 & 36.22 & \cellcolor{oursrow}\underline{1.29} & \cellcolor{oursrow}\textbf{1.03} \\
 & 27B & 36.87 & 35.17 & 35.68 & 15.38 & 37.24 & 22.76 & 11.86 & 20.54 & 14.69 & 11.91 & 37.42 & \cellcolor{oursrow}\underline{0.52} & \cellcolor{oursrow}\textbf{0.22} \\
\cmidrule(lr){1-15}
\multirow{2}{*}{\textbf{Mistral}} & 7B & 26.08 & 15.33 & 23.49 & 32.35 & 25.22 & 33.03 & 10.08 & 12.10 & 11.49 & 10.07 & 36.29 & \cellcolor{oursrow}\underline{2.41} & \cellcolor{oursrow}\textbf{2.40} \\
 & 24B & 34.25 & 31.99 & 32.63 & 20.62 & 33.73 & 11.46 & 10.42 & 20.59 & 11.31 & 10.89 & 35.44 & \cellcolor{oursrow}\underline{0.84} & \cellcolor{oursrow}\textbf{0.59} \\
\cmidrule(lr){1-15}
\multirow{1}{*}{\textbf{Phi}} & 14B & 30.99 & 25.10 & 27.18 & 12.08 & 25.47 & 10.84 & 9.87 & 31.97 & 11.42 & 9.76 & 36.12 & \cellcolor{oursrow}\underline{1.26} & \cellcolor{oursrow}\textbf{0.88} \\
\cmidrule(lr){1-15}
\multirow{1}{*}{\textbf{gpt-oss}} & 20B & 11.70 & 18.35 & 17.77 & 13.06 & 31.53 & 13.07 & 12.50 & 13.11 & 13.70 & 11.72 & 35.77 & \cellcolor{oursrow}\underline{0.65} & \cellcolor{oursrow}\textbf{0.38} \\
\bottomrule
}
\caption{\textbf{Controlled episode regret per LLM}, averaged over domains (Table~\ref{tab:episodes}). As each LLM takes the final action, Never varies by LLM, unlike in the other per-LLM tables. Baselines are cited in Table~\ref{tab:know}.}
\label{tab:full_episodes_llm}
\end{table*}